\documentclass[11pt,a4paper]{article}

\usepackage[utf8]{inputenc}
\usepackage[T1]{fontenc}

\usepackage[a4paper, margin=25mm]{geometry}
\usepackage{setspace}
\usepackage{amsmath}
\usepackage{amssymb}
\usepackage{amsfonts}

\usepackage{graphicx}
\usepackage{subcaption}
\usepackage{pdflscape}
\usepackage{placeins}

\usepackage{booktabs}
\usepackage{longtable}
\usepackage{multirow}
\usepackage{makecell}
\usepackage{array}
\usepackage[para,online,flushleft]{threeparttable}

\usepackage[numbers,sort&compress]{natbib}

\usepackage[table]{xcolor}
\usepackage[hidelinks]{hyperref}

\title{Cross-Dataset Transfer and Reliability of Explainable Artificial
  Intelligence for RhythmFormer Remote Photoplethysmography}

\author{%
  Louis Chen\thanks{%
    Department of Mechanical Engineering, National Cheng Kung University, Tainan, Taiwan.
    ORCID: \href{https://orcid.org/0009-0001-8373-2295}{0009-0001-8373-2295}.
    Email: \texttt{louis.chen@nordlinglab.org}.%
  }%
  \and
  Torbj{\"o}rn E.\,M.\ Nordling\thanks{%
    Department of Mechanical Engineering, National Cheng Kung University, Tainan, Taiwan.
    ORCID: \href{https://orcid.org/0000-0003-4867-6707}{0000-0003-4867-6707}.
    Corresponding author: \texttt{torbjorn.nordling@nordlinglab.org}.%
  }%
}

\date{\today}

\begin{document}
\maketitle

\begin{abstract}
\textbf{Background.}
Remote photoplethysmography estimates the cardiovascular pulse from facial video, and its explanations have rested on inspecting heatmaps rather than on quantitative evidence about where a model reads it.
We quantified the explanations and asked whether such explanations transfer between datasets and track model performance.
\textbf{Method.}
We trained eight condition-specific RhythmFormer models on NCKU-rPPG, recorded under three illumination levels, speaking, rotation, and cycling, estimated one heart rate per 5.12-second clip, and set them beside a UBFC-rPPG reproduction.
Raw attention, rollout, attention flow, and Beyond Intuition were assessed by skin coverage and the Salience-guided Faithfulness Coefficient (SaCo).
\textbf{Results.}
Beyond Intuition ranked highest on both datasets, at median coverage 0.789 and SaCo 0.837 on Static level~3 against 0.826 and 0.917 on UBFC-rPPG; lower ranks differed.
Within one participant of one condition, neither measure was related to a clip's heart-rate error, waveform correlation, or signal-to-noise ratio on either dataset: 186 of the 252 coefficients fell below $|\rho|=0.10$ and 28 reached $p<0.05$ against the 13 expected by chance.
Across the eight scenarios only Beyond Intuition's coverage followed the three performance measures, at $\rho=-0.43$, $+0.57$, and $+0.43$, while the attention-only methods' SaCo ran opposite to each.
It failed at 40~lux alone, its median coverage falling to 0.180 and its median SaCo to $-0.178$, whereas motion degraded the estimates far more without such a drop.
\textbf{Conclusions.}
Skin coverage and SaCo carry information complementary to the performance measures rather than a proxy for them: attributing to the skin does not guarantee an accurate estimate.
What an attribution reveals about a condition is where the model looks rather than how faithfully its map is ordered.

\end{abstract}


\bigskip

\section{Introduction}
\label{sec:intro}

Remote photoplethysmography (rPPG) recovers the cardiac pulse from ordinary facial video, but the pulse-induced skin-colour change is weak and easily confounded by illumination, camera response, and motion.
Public rPPG datasets span stationary recordings, deliberate head motion, speaking, exercise, and changes in illumination, but they differ markedly in camera characteristics, physiological references, sampling rates, and participant composition~\cite{stricker2014NoncontactMeasurementIEEE,heusch2017ReproducibleMeasurementArXiv,niu2018VIPLHRMultimodalAsianVision,bobbia2019UnsupervisedPPGPatternRecognitLett,tang2023MmpdMultidomainEMBC}.
These differences make dataset adaptation a substantive part of model evaluation and a model trained on one dataset may not generalise to another, even if the datasets are similar in some respects~\cite{liu2023RPPGToolboxToolboxNeurIPS,tang2023MmpdMultidomainEMBC}.
The best performing rPPG models are deep neural networks (DNNs)~\cite{debnath2025MeasPPGBioMedEngOnLine}, but their complexity makes it difficult to understand how they work and whether they can be trusted~\cite{hassija2024InterpretingExplainableCognComput}.
In particular in clinical applications, it is important to know when a model's predictions are reliable and whether its explanations are faithful to the model's behaviour.
The present study investigates whether explainable artificial intelligence (XAI) methods can provide insight into rPPG model behaviour and whether their explanations transfer across datasets and track model reliability.

RhythmFormer is a DNN demonstrated to perform well on some datasets.
RhythmFormer uses a hierarchical Temporal Periodic Transformer with Periodic Sparse Attention to recover an rPPG waveform from facial video~\cite{zou2025RhythmFormerrPPGPatternRecognit}.
At the Nordling Lab at National Cheng Kung University (NCKU), we have collected a remote photoplethysmography (rPPG) multi-condition dataset, hereafter NCKU-rPPG, which includes eight scenarios with different illumination, motion, and speaking conditions.
This is to overcome the limitations of existing datasets, like the widely used UBFC-rPPG dataset that we also used, and to provide a more comprehensive evaluation of rPPG models.
We therefore investigated whether RhythmFormer explanations transfer across datasets and whether their faithfulness tracks model reliability.
First, we compare explainable artificial intelligence (XAI) methods on Static level~3 and UBFC-rPPG, the NCKU-rPPG condition and the reference dataset whose recording settings and clip lengths are closest, to test whether attribution rankings and correlation structures recur.
Second, we compare each method's median Salience-guided Faithfulness Coefficient across eight NCKU-rPPG scenarios with scenario-level mean absolute error, waveform Pearson correlation, and signal-to-noise ratio.
The conversion, participant-independent split, and eight independently trained checkpoints provide the experimental basis for these comparisons, but the design supports descriptive transfer and reliability evidence rather than significance, causality, or cross-dataset generalisation claims.

\section{Related Work}
\label{sec:related_work}

Remote photoplethysmography (rPPG) validation datasets pair facial video with synchronised contact cardiac signals, such as electrocardiography (ECG) or photoplethysmography (PPG).
Dataset suitability for this study depends first on whether complete facial videos and synchronised physiological signals can be obtained under workable research terms.
PURE~\citep{stricker2014NoncontactMeasurementIEEE}, COHFACE~\citep{heusch2017ReproducibleMeasurementArXiv}, and UBFC-rPPG~\citep{bobbia2019UnsupervisedPPGPatternRecognitLett} provide direct research-use acquisition routes, although their publications do not state a named standard public licence.
MAHNOB-HCI~\citep{soleymani2012MultimodalTaggingIEEETransAffectComput}, ECG-Fitness~\citep{spetlik2018EstimationConvolutionalMachineVision}, AMIGOS~\citep{correa2021AmigosDatasetIEEETransAffectComput}, VIPL-HR~\citep{niu2018VIPLHRMultimodalAsianVision}, and MMPD~\citep{tang2023MmpdMultidomainEMBC} require a signed academic or non-commercial data-use agreement before access and are therefore usable only after that administrative step.

The remaining datasets have more restrictive practical roles.
LGI Multi-Session~\citep{pilz2018InvarianceEstimationIEEE} has an announced research licence, but a complete working download was not available at the time of this study.
Bias-eval-rPPG~\citep{dasari2021BiasesPhotoplethysmographyNPJDigitMed} releases frame-wise cropped skin-region colour values rather than the original facial videos, so it cannot support the present spatial XAI pipeline.
SCAMPS~\citep{mcduff2022ScampsSyntheticsAdvNeuralInfProcessSyst} is directly obtainable for non-commercial research but contains synthetic rather than recorded participants.
No public acquisition route for the complete OBF dataset~\citep{li2018OBFFibrillationIEEE} was identified.
NCKU-rPPG is available to the present study under its ethics approvals and institutional controls but is not a public dataset.
These access conditions, together with the need for complete facial frames and synchronised waveforms, made UBFC-rPPG the practical public reference and NCKU-rPPG the controlled multi-condition extension.

Recent explainable artificial intelligence (XAI) studies in rPPG have mainly used visual attribution to indicate where a model may obtain its physiological evidence.
Dual-path TokenLearner, PhysKANNet, DD-rPPGNet, and TS-CAN+ respectively report attention maps, gradient saliency, or Gradient-weighted Class Activation Mapping (Grad-CAM) visualisations, while CIN-rPPG examines channel--spatial interaction maps internal to its architecture~\citep{qian2024TokenLearnerPPGIEEETransComputSocSyst,liu2025PhysKANNetKANbasedBiomedSignalProcessControl,huang2025DDrPPGNetrPPGIEEETransInfForensicsSecur,li2025TSCANNonContactIEEETransConsumElectron,li2024InteractEstimIEEETransCircuitsSystVideoTechnol}.
These visualisations can reveal plausible facial, non-facial, and background responses, but visual plausibility alone does not establish that the highlighted regions affect an rPPG prediction.
In particular, the available studies do not quantitatively test whether the ordering imposed by an attribution map agrees with the effect of perturbing its spatial regions.

RhythmFormer was developed to model periodic spatiotemporal patterns with hierarchical Temporal Periodic Transformer stages and Periodic Sparse Attention~\citep{zou2025RhythmFormerrPPGPatternRecognit}.
Chen and Nordling~\citep{chen2026RhythmFormerPhotoplethysmographyEXPLIMED} addressed this limitation for RhythmFormer on UBFC-rPPG by adapting the Salience-guided Faithfulness Coefficient (SaCo) to rPPG regression and pairing perturbation faithfulness with skin-coverage analysis.
That evaluation showed that attention aggregation across layers can restore non-zero multi-hop paths that sparse top-$k$ routing had excluded, which weakens the spatial specificity of attention rollout.
Beyond Intuition obtained the strongest median skin alignment and SaCo result in that controlled evaluation, but the observation was restricted to one dataset, one model family, and a limited recording setting.
It therefore motivates replication rather than a general ranking of XAI methods.

The present study extends that experimental framework from UBFC-rPPG to eight scenario-specific models trained and evaluated on NCKU-rPPG, covering controlled variation in illumination, speaking, head rotation, and cycling intensity.
It examines whether the previous observations transfer to independently trained RhythmFormer models across these recording scenarios.
The analysis also distinguishes SaCo's within-clip perturbation-faithfulness interpretation from its possible association with reliability across independently trained models, which is not implied by its definition.
Finally, comparison across scenarios provides descriptive evidence on whether the multi-hop leakage observed on UBFC-rPPG also appears under differing illumination, speech, head rotation, and cycling conditions.
Published dataset results do not remove the need to document adaptation because input duration, frame rate, face preprocessing, participant grouping, and evaluation aggregation can all change the experiment.

\section{Methods}
\label{sec:methods}

\subsection{Dataset and experimental conditions}
\label{sec:dataset_conditions}

We analysed the Nordling Lab at National Cheng Kung University (NCKU) remote photoplethysmography (rPPG) multi-condition dataset, hereafter NCKU-rPPG.
We followed the experimental protocol established in the Nordling Lab and presented by \citet{wang2020NoncontactMeasurementNCKU}.
The source cohort submitted to conversion contained 618 Camera~1 recording sessions from 78 participants.
The recordings paired facial video with contact photoplethysmography (PPG) and electrocardiography (ECG) under controlled illumination and motion conditions.
Because each participant contributed at most one session per scenario, the numbers of participants with a usable session and usable sessions were identical: 76 for Static level~1, 76 for Static level~3, 75 for Static level~5, 76 for Speak, 76 for Rotate, 77 for Bike level~1, 76 for Bike level~3, and 75 for Bike level~5.

\subsubsection{Ethical statement}
\label{sec:ethics}

The National Cheng Kung University Human Research Ethics Committee approved the relevant experimental procedures under case numbers 106-262, 108-244, and 112-406.
The corresponding approval numbers were NCKU HREC-F-107-008-2, NCKU HREC-E-108-244-2, and NCKU HREC-E-112-406-2, respectively.
Data collection was conducted in accordance with the approved procedures.
\subsubsection{Acquisition system}
\label{sec:experimental_setup}

The acquisition system comprised the following devices.
Camera~1 recorded the frontal facial view with a Panasonic GX85 and an Olympus M.ZUIKO DIGITAL ED 14--150-mm lens at $1920\times1080$ pixels and 50~frames per second as compressed H.264 High Profile video with \texttt{yuv420p} pixel format, and an average bit rate of approximately 26.98~Mb/s (range 25.89--27.08~Mb/s).
A Logitech C170 webcam viewed the fluorescent ceiling light and provided a synchronisation cue from the abrupt brightness change at the beginning of each test.
A Philips IntelliVue MX400 recorded three-lead ECG at 500~Hz and finger PPG from the \texttt{SpO$_2$} channel at 125~Hz.
The alignment files paired the selected Camera~1 frames with the 125~Hz PPG measurements aligned and supplied at the source video's 50~Hz frame cadence used in this analysis.
Two FOTGA LED504 lights and the fluorescent ceiling light established the illumination conditions, which were checked with a TES 1330A lux meter beside the participant's face.

\begin{table}[htbp]
\centering
\caption{Acquisition devices used for the analysed recordings.}
\label{tab:devices}
\small
\setlength{\tabcolsep}{4pt}
\begin{tabular}{>{\raggedright\arraybackslash}p{2.8cm} >{\raggedright\arraybackslash}p{4.2cm} >{\raggedright\arraybackslash}p{7.3cm}}
\toprule
Component & Device & Analytic configuration \\
\midrule
Camera~1 & Panasonic GX85 with Olympus 14--150-mm lens & Frontal portrait recording; $1920\times1080$ pixels; H.264 High Profile, \texttt{yuv420p}; approximately 26.98~Mb/s (25.89--27.08~Mb/s); 50~frames per second; fixed 5500-K white balance. \\
Synchronisation camera & Logitech C170 & Viewed the fluorescent light to identify the alignment cue. \\
Physiological monitor & Philips IntelliVue MX400 & Three-lead ECG at 500~Hz and finger PPG at 125~Hz; the aligned PPG sequence was supplied at 50~Hz. \\
Exercise apparatus & BH Fitness H917 flywheel & Used for the three six-minute cycling conditions. \\
Illumination & Two FOTGA LED504 lights and fluorescent ceiling light & Frontal illuminance was measured with a TES 1330A lux meter. \\
Instruction display & Television & Displayed the text and head-rotation instructions used in the Speak and Rotate scenarios. \\
\bottomrule
\end{tabular}
\end{table}

\subsubsection{Experiment design}
\label{sec:experimental_conditions}

The protocol comprised five two-minute seated recordings and three six-minute cycling recordings.
Static recordings were acquired at illumination levels~1, 3, and~5.
Speaking and head rotation were recorded at level~3.
The three cycling recordings repeated illumination levels~1, 3, and~5 while participants changed cadence and resistance under researcher supervision.
Table~\ref{tab:tests_in_sessions} summarises the eight analysed scenarios.

\begin{table}[htbp]
\centering
\caption{Analysed experimental scenarios.}
\label{tab:tests_in_sessions}
\small
\setlength{\tabcolsep}{3pt}
\begin{tabular}{>{\raggedright\arraybackslash}p{2.7cm} c c >{\raggedright\arraybackslash}p{7.5cm}}
\toprule
Scenario & \makecell{Illumination\\level} & Duration & Procedure \\
\midrule
Static level~1 & 1 & 2~min & Sit still on the flywheel and look towards Camera~1. \\
Static level~3 & 3 & 2~min & Sit still on the flywheel and look towards Camera~1. \\
Static level~5 & 5 & 2~min & Sit still on the flywheel and look towards Camera~1. \\
Speak & 3 & 2~min & Read the displayed text aloud while facing the screen. \\
Rotate & 3 & 2~min & Follow displayed instructions to rotate the head through multiple poses. \\
Bike level~1 & 1 & 6~min & Cycle while following instructions to change cadence and resistance. \\
Bike level~3 & 3 & 6~min & Cycle while following instructions to change cadence and resistance. \\
Bike level~5 & 5 & 6~min & Cycle while following instructions to change cadence and resistance. \\
\bottomrule
\end{tabular}
\end{table}

Illumination levels~1, 3, and~5 corresponded to frontal illuminance targets of $40\pm50$, $200\pm50$, and $700\pm50$~lux, respectively.
The static conditions isolated the planned illumination settings while limiting voluntary movement.
Speaking introduced local facial motion, rotation introduced large pose changes, and cycling combined body motion with a changing heart rate.

\subsection{Face extraction and resizing}
\label{sec:conversion}

\begin{figure}[!htbp]
\centering
\includegraphics[width=\textwidth,height=0.76\textheight,keepaspectratio]{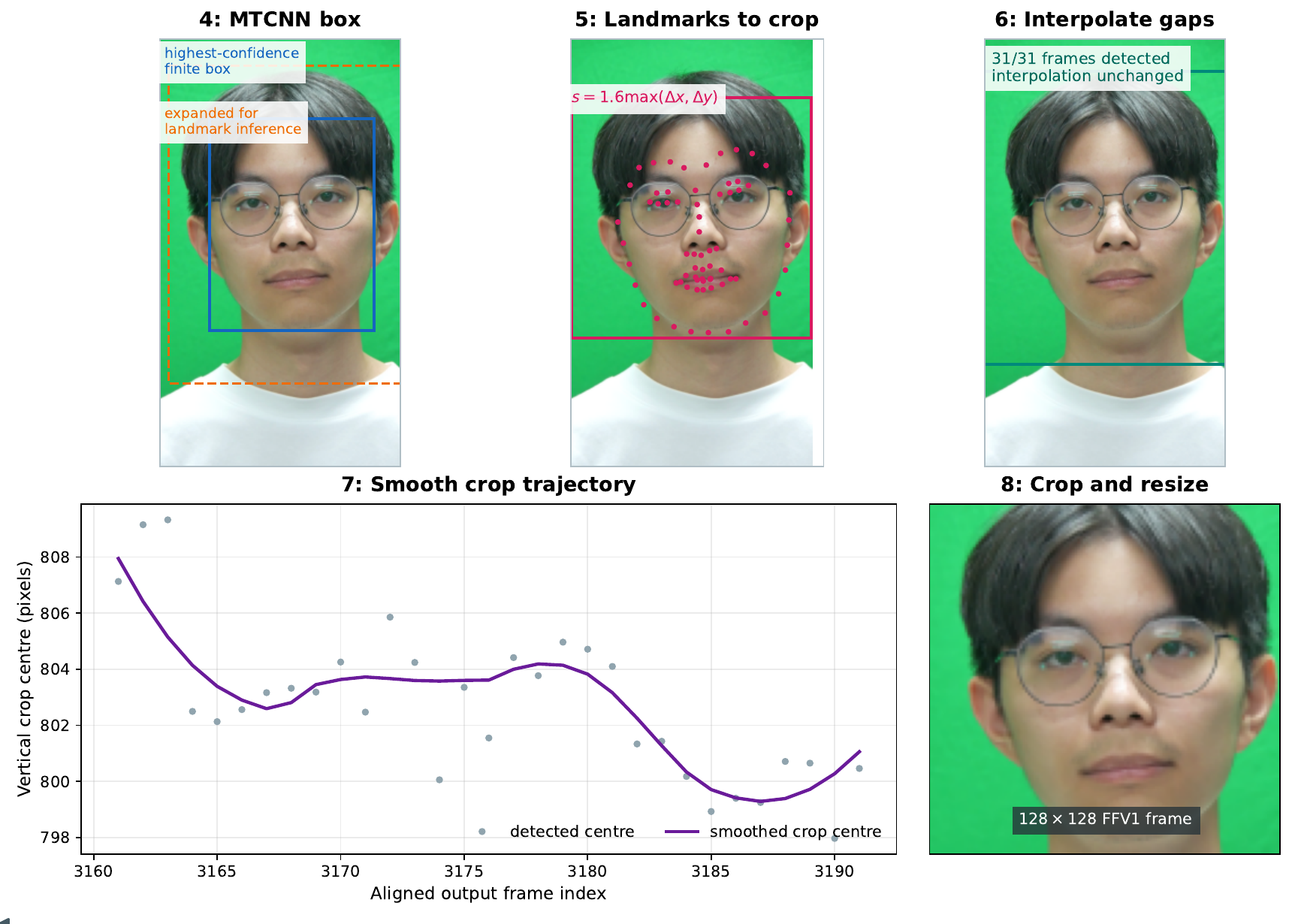}
\caption{Face extraction and resizing for NCKU-rPPG conversion, illustrated with output frame 3176 from participant 798 in the static level 3 condition and the corresponding reproduced preprocessing evidence.}
\label{fig:facex_extraction}
\end{figure}
We converted the source recordings into cropped facial videos paired with aligned PPG sequences for RhythmFormer training and evaluation.
  The face-extraction pipeline first generated face candidates using the Multi-task Cascaded Convolutional Network (MTCNN) implementation provided by \texttt{facenet-pytorch} version
  2.6.0~\citep{zhang2016CascadedConvolutionalIEEESignalProcessLett}.
  It then estimated 68 facial landmarks using FaceXFormer~\citep{narayan2025FaceXFormerTransformerICCV}.
  For reproducibility, FaceXFormer was pinned to commit \path{10fe8291f8a64e2ca1daf938e3e0007bd860303b} of \url{https://github.com/Kartik-3004/facexformer} and to the released checkpoint \path{ckpts/model.pt} from \url{https://
  huggingface.co/kartiknarayan/facexformer}.
  The checkpoint had the SHA-256 digest \path{327a755849ba64d336fb96589ff87b27e84a12be1ecf8bcfaa503d66f803286d}.
  Before processing, the converter verified the source commit and checkpoint digest and terminated if either did not match the specified version.
  The conversion proceeded as follows.
The conversion proceeded as follows.

\begin{enumerate}
  \item We read each Camera~1 video and its unique alignment CSV file, which supplied the aligned start frame and PPG values.
  \item We retained the shorter of the decodable aligned-video sequence and the contiguous finite PPG sequence beginning at the aligned start frame.
  The processing and verification stages required all expected source frames to be decodable.
  We retained the 50~Hz video cadence and did not resample the recordings to the 30~Hz rate used by UBFC-rPPG.
  \item We applied the participant-specific clockwise or counter-clockwise 90-degree correction to obtain a portrait frame.
  \item We ran a Multi-task Cascaded Convolutional Network (MTCNN) face detector~\cite{zhang2016CascadedConvolutionalIEEESignalProcessLett} on every frame, selected the finite candidate with the highest confidence, and expanded that candidate before landmark inference.
  \item FaceXFormer \citep{narayan2025FaceXFormerTransformerICCV} estimated 68 facial landmarks in source-frame coordinates.
  We converted the landmark extrema to a square rPPG crop with side length
  \begin{equation}
    s=1.6\max(x_{\max}-x_{\min},y_{\max}-y_{\min}),
    \label{eq:facex_side}
  \end{equation}
  centred horizontally on the landmarks and shifted upward by eight percent of the square crop side to include more forehead.
  Both the side and the shift were recomputed from each frame's own landmark span, so the displacement varied frame by frame and had no single value in pixels.
  \item We linearly interpolated missing crop centres and side lengths.
  Nearest valid values filled leading and trailing gaps.
  Each session required a raw face-detection rate of at least 0.80 before interpolation.
  \item We smoothed each interpolated centre coordinate and side length with a five-frame median filter followed by an 11-frame, second-order Savitzky--Golay filter.
  If a recording was shorter than a requested filter window, the implementation used the largest odd integer no greater than both the requested window and the number of frames; it skipped the median stage below three frames and the Savitzky--Golay stage unless the window exceeded the polynomial order.
  \item We shifted each square crop inside the frame without changing its side length, extracted it from the rotated source frame, and resized it from approximately $510\times640$ pixels to $128\times128$ pixels.
  \item We wrote a lossless FFV1 \texttt{vid.avi} and a corresponding \texttt{ground\_truth.txt} for each session.
  We required a minimum encode--decode temporal RGB correlation of 0.99 between the source frames and the decoded output, and the converter deleted the output and stopped for any session below it.
  Every one of the 607 accepted sessions therefore met the threshold.
\end{enumerate}

\subsection{Participant split and RhythmFormer preprocessing}
\label{sec:rhythmformer_preprocessing}

Of the 78 participants in the source cohort, one had all eight sessions skipped because the required alignment CSV files were absent.
The converted dataset therefore contained 607 sessions from 77 participants, which we split before selecting a scenario.
The fixed manifest assigned 65 participants and 511 converted sessions to training and 12 participants and 96 converted sessions to testing.
Scenario filtering was applied only after this grouping, preventing different sessions from the same participant from entering both partitions.
Missing sessions reduced the scenario-specific training set to 63--65 participants, whereas every test scenario contained all 12 test participants.
No original participant or converted session appeared in both the training and test partitions in any of the eight scenarios.

RhythmFormer loader's own face cropping was disabled throughout.It read the already cropped $128\times128$ FFV1 videos directly and converted OpenCV's decoded blue--green--red channel order to red--green--blue and did not perform another face detection, dynamic tracking, expanded crop, or change of spatial resolution in the formal experiments.

For each complete session, the loader standardised the video with one scalar mean and one scalar standard deviation computed jointly over the time, height, width, and RGB-channel dimensions of the $T\times H\times W\times C$ tensor.
It independently standardised the PPG vector using its mean and standard deviation over time.
Standardisation preceded temporal segmentation.

We divided each session into non-overlapping 256-frame clips.
At 50~Hz, each clip spanned 5.12~s, which approximated but did not exactly match the 5.33~s duration of the 160-frame, 30~Hz clips used to train RhythmFormer on UBFC-rPPG.
We discarded the terminal remainder when fewer than 256 frames remained and did not resample time.
Video clips and labels were stored as 32-bit floating-point arrays.
The loader converted the video to time--channel--height--width order before batching.

We excluded 285 training clips and 80 test clips whose PPG-labels contained a non-finite value or had a 32-bit floating-point standard deviation no greater than $10^{-6}$, and trained on the remainder.
Supplementary Section~\ref{sec:supp_labelchecks} distinguishes this gate from the training-time augmentation fallback.

\subsection{RhythmFormer architecture}
\label{sec:rhythmformer_architecture}

RhythmFormer maps an RGB facial video to a one-dimensional rPPG waveform~\cite{zou2025RhythmFormerrPPGPatternRecognit}.
The Fusion Stem combines raw frames with temporal frame differences, and a three-dimensional convolution embeds non-overlapping spatial patches.
The features pass through three hierarchical Temporal Periodic Transformer (TPT) stages.
Each stage contains two Periodic Sparse Attention (PSA) blocks, and stage-specific temporal downsampling and upsampling expose periodic structure at multiple temporal scales.
After the six PSA blocks, spatial averaging and a one-dimensional predictor produce one waveform value per input frame.

\ifdefined\thesisourdatasetarchitecturexref
\thesisourdatasetarchitecturexref
\else
\begin{figure}[htbp]
\centering
\includegraphics[width=0.78\textwidth]{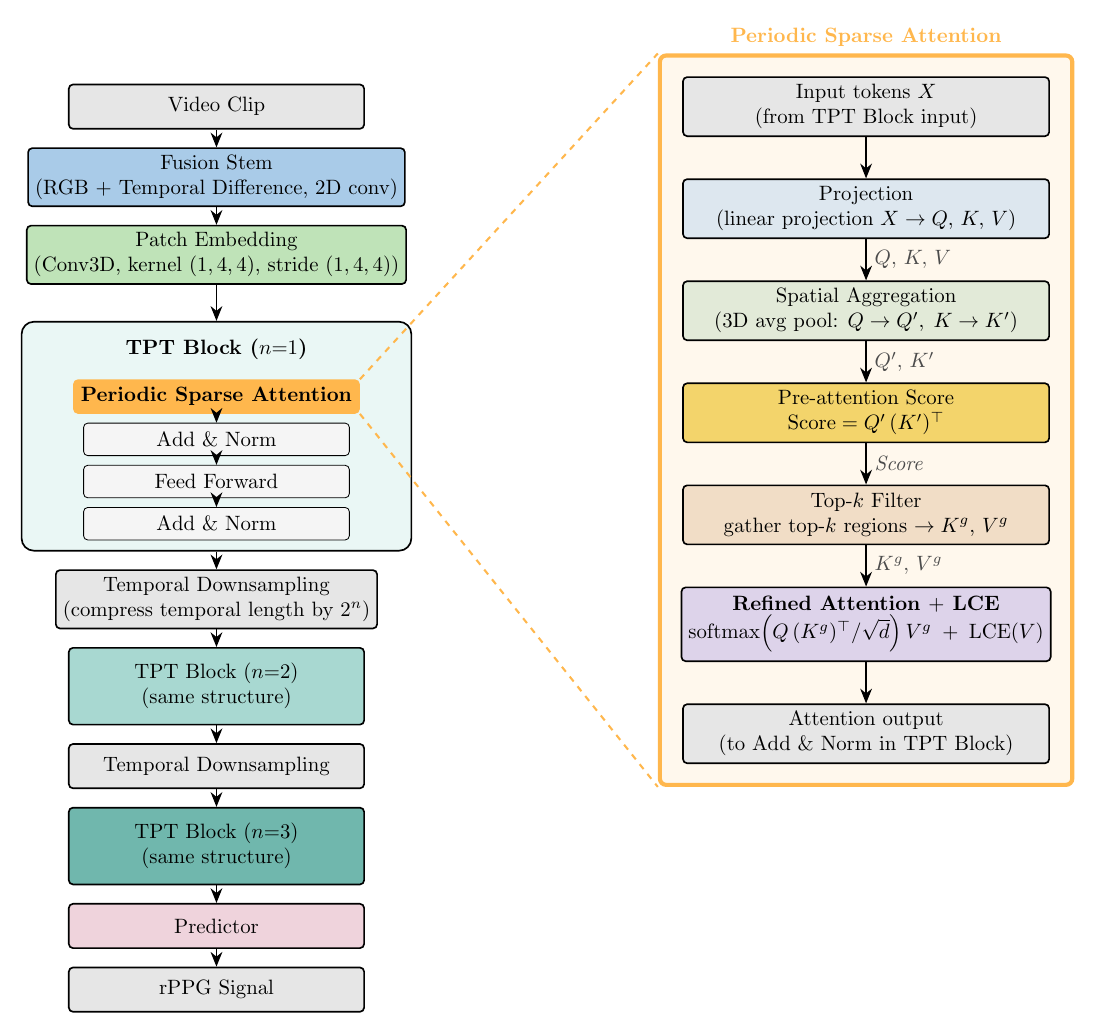}
\caption{RhythmFormer architecture and Periodic Sparse Attention (PSA).
The Fusion Stem and patch embedding precede three Temporal Periodic Transformer (TPT) stages, each containing two PSA blocks.
PSA uses coarse scores to route selected keys and values before refined attention, with a parallel local-context branch.}
\label{fig:rhythmformer_architecture}
\end{figure}
\fi
For these 256-frame, 50~Hz experiments, the trainer instantiated the three stages with top-$k=40$ routing.
The model applied no stochastic depth: its drop path rate was 0.

\subsection{Scenario-specific training}
\label{sec:training_protocol}

We trained an independent RhythmFormer model for each of the eight scenarios in Table~\ref{tab:tests_in_sessions}.
Each model used only the training clips from its scenario and was tested only on the corresponding held-out scenario clips.
Table~\ref{tab:training_settings} gives the shared formal configuration.

\begin{table}[htbp]
\centering
\caption[Training and inference configuration]{Shared training and inference configuration for the eight scenario-specific models.}
\label{tab:training_settings}
\small
\begin{tabular}{>{\raggedright\arraybackslash}p{5.2cm} >{\raggedright\arraybackslash}p{9.0cm}}
\toprule
Setting & Value \\
\midrule
Random seed and precision & 100; FP32 \\
Hardware & Four NVIDIA RTX A5000 graphics processing units \\
Software & PyTorch 2.0.1 with CUDA 11.8 \\
Epochs and checkpoint & 30; final checkpoint from epoch~30 \\
Training and test batch sizes & 4 and 2 \\
Learning rate and scheduler & 0.009; OneCycleLR \\
Optimiser & AdamW; weight decay 0; effective $\epsilon=10^{-8}$ \\
Recorded model drop rate & 0.2 \\
Data-loader workers & 16 \\
\bottomrule
\end{tabular}
\end{table}

For predicted waveform $\hat{y}$ and reference waveform $y$, the training objective was
\begin{equation}
  \mathcal{L}
  =0.2\mathcal{L}_{\mathrm{NP}}(\hat{y},y)
  +\mathcal{L}_{\mathrm{CE}}(\hat{y},y)
  +\mathcal{L}_{\mathrm{HR-KL}}(\hat{y},y),
  \label{eq:training_loss}
\end{equation}
where $\mathcal{L}_{\mathrm{NP}}$ is the negative-Pearson time-domain loss, $\mathcal{L}_{\mathrm{CE}}$ is cross-entropy over the frequency-domain heart-rate target, and $\mathcal{L}_{\mathrm{HR-KL}}$ is the Kullback--Leibler loss between predicted and reference heart-rate distributions.
These three components follow the published RhythmFormer training objective~\citep{zou2025RhythmFormerrPPGPatternRecognit}.

Heart-rate-aware temporal augmentation was sampled during training.
Each clip was selected for augmentation independently with probability 0.5.
The routine estimated each clip's reference heart rate from the clip's own label, by the detrending, 0.75--2.5-Hz bandpass, and Welch procedure described in Section~\ref{sec:performance_evaluation} applied at the 50-Hz training sampling rate.
A clip counted as low-heart-rate below 75 beats per minute and as high-heart-rate above 90 beats per minute.
For a selected low-heart-rate clip, the routine retained every second frame and repeated the accelerated sequence; for a selected high-heart-rate clip, it expanded a randomly selected half-length interval by alternating source samples with their interpolated neighbours.
Other selected clips, those from 75 to 90 beats per minute inclusive, remained temporally unchanged.
A batch-level horizontal flip was sampled independently.

Some NCKU-rPPG labels contained a constant terminal suffix.
Temporal augmentation could select only that suffix and thereby create a constant target with an undefined Pearson denominator even when the original full clip was valid.
After all temporal-augmentation random draws, the trainer therefore checked each augmented label.
If any value was non-finite or the standard deviation was no greater than $10^{-8}$, it restored the original unaugmented video--label pair.
The check did not redraw the augmentation or horizontal-flip decision and therefore did not alter the random sampling sequence.

The optimiser was AdamW with $\epsilon=10^{-8}$ and zero weight decay.
The formal runs did not construct a validation loader or select a model by validation performance.
Testing instead used the model from the final training epoch.

\subsection{Performance evaluation}
\label{sec:performance_evaluation}

\paragraph{Analysis unit.}
Two conventions for heart-rate metrics dominate the 2022--2026 remote photoplethysmography literature.
The rPPG-Toolbox convention estimates one heart rate per test sample, computes mean absolute error (MAE), root mean square error (RMSE), mean absolute percentage error (MAPE), Pearson correlation, and signal-to-noise ratio (SNR) across those samples, and reports every estimate with a standard error~\citep{liu2023RPPGToolboxToolboxNeurIPS}.
The RhythmNet and PhysFormer convention divides a 30-second recording into three non-overlapping 10-second clips, estimates a heart rate in each, averages the three into one recording-level value, and reports standard deviation, MAE, RMSE, and Pearson correlation across recordings~\citep{niu2020RhythmNetEstimationIEEETransImageProcess,yu2022PhysFormerTransformerCVF}.

We follow the rPPG-Toolbox definitions for all five metrics.
It is the code this experiment ran, it is the convention behind the published RhythmFormer results we compare with~\citep{zou2025RhythmFormerrPPGPatternRecognit}, and of the two it is the only one that defines every metric we report, the other reporting neither MAPE nor SNR.

The toolbox takes a whole test video as one sample, which in the datasets it was built around lasts about a minute.
Our held-out recordings are far longer: 123~s per participant in the static, speaking, and rotation conditions, and 364~s under cycling.
A single spectral peak over 364~s is a modal value rather than a heart rate, and the cycling protocol exists precisely to move the heart rate within that interval.
The reference finger photoplethysmogram's median within-participant interquartile range is 4.1--5.1~beats per minute in the five stationary conditions and 13.4--17.5 under cycling, and the change between consecutive clips exceeds 10~beats per minute in 2.5--9.8~\% of stationary pairs against 18.7--23.8~\% of cycling pairs.
We therefore estimate one heart rate per 256-frame clip, which is 5.12~s, as the second convention does on its 10-second clips, but we do not average the clips into a recording-level value.
This also makes the comparison with the UBFC-rPPG reproduction possible, because it uses a 160-frame clip, approximately 5.33~s at 30~Hz~\citep{chen2026RhythmFormerPhotoplethysmographyEXPLIMED}.

The shorter window reduces the native frequency resolution, so we quantified its practical effect. For a 256-sample window at 50 Hz, adjacent discrete Fourier transform bins are separated by 11.7 beats per minute. Zero-padding and peak interpolation permit estimates between these sampled frequencies, although they do not increase the intrinsic spectral resolution. We estimated heart rate from each reference photoplethysmogram clip using two methods: the dominant spectral peak and the median interval between detected systolic peaks. Across the five stationary conditions, the median absolute difference between the two estimates ranged from 0.65 to 1.09 beats per minute. Across the three cycling conditions, the median absolute difference ranged from 1.81 to 2.47 beats per minute, whereas the mean absolute difference ranged from 5.49 to 8.54 beats per minute. The substantially larger means indicate that a minority of cycling clips produced large discrepancies between the two estimators. In addition, 30 of the 2,559 cycling clips produced spectral heart-rate estimates at the upper boundary of the 150-beats-per-minute search band and were therefore censored; no stationary clip reached this boundary. Together, these results show that short-clip reference heart-rate estimation was generally consistent under stationary conditions but less stable during cycling, particularly for outlying or high-heart-rate clips.

\paragraph{Heart-rate estimation.}
Each clip's predicted and reference waveforms were detrended with smoothness parameter 100 and filtered with a first-order, zero-phase Butterworth bandpass from 0.75 to 2.5~Hz.
We estimated the dominant heart rate over the corresponding 45--150~beats-per-minute range with Welch spectra using a discrete Fourier transform length of $\mathrm{nfft}=200{,}000$.

\paragraph{Definitions.}
Let $P=12$ be the held-out participants of a condition, $\mathcal{C}_p$ the clips of participant $p$, $N_p=\lvert\mathcal{C}_p\rvert$, and $N=\sum_{p=1}^{P}N_p$.
Write $\hat{h}_i$ and $h_i$ for the predicted and reference heart rate of clip $i$ in beats per minute, and $\hat{s}_i[t]$ and $s_i[t]$ for its predicted and reference waveform samples, $t=1,\dots,T$ with $T=256$.

Every estimate below is a participant-weighted mean: clip values are averaged within a participant first, and the participant means are then averaged.
For a clip-level quantity $x_i$,
\begin{equation}
\label{eq:clustered_mean}
\bar{x}_p = \frac{1}{N_p}\sum_{i\in\mathcal{C}_p} x_i,
\qquad
\langle x\rangle = \frac{1}{P}\sum_{p=1}^{P}\bar{x}_p,
\qquad
\mathrm{SE}\left(\langle x\rangle\right) = \frac{1}{\sqrt{P}}\sqrt{\frac{1}{P-1}\sum_{p=1}^{P}\left(\bar{x}_p-\langle x\rangle\right)^2}.
\end{equation}
The three error measures are then
\begin{align}
\label{eq:mae}
\mathrm{MAE} &= \Bigl\langle\, \lvert \hat{h}_i - h_i \rvert \,\Bigr\rangle, \\
\label{eq:rmse}
\mathrm{RMSE} &= \sqrt{\Bigl\langle\, \bigl(\hat{h}_i - h_i\bigr)^2 \,\Bigr\rangle}, \\
\label{eq:mape}
\mathrm{MAPE} &= \Biggl\langle\, \frac{\lvert \hat{h}_i - h_i \rvert}{h_i} \,\Biggr\rangle \times 100\,\%,
\end{align}
with the standard error of the RMSE obtained from that of the mean squared error inside Equation~\eqref{eq:rmse} by the delta method, that is by dividing it by $2\,\mathrm{RMSE}$.

We report two Pearson correlations, because the manuscript uses the word for two different quantities and they must not be confused.
The heart-rate correlation compares the $N$ estimated heart rates of a condition with their references,
\begin{equation}
\label{eq:pearson_hr}
r_{\mathrm{HR}} = \frac{\sum_{i=1}^{N}\bigl(\hat{h}_i-\overline{\hat{h}}\bigr)\bigl(h_i-\overline{h}\bigr)}
{\sqrt{\sum_{i=1}^{N}\bigl(\hat{h}_i-\overline{\hat{h}}\bigr)^{2}}\;\sqrt{\sum_{i=1}^{N}\bigl(h_i-\overline{h}\bigr)^{2}}},
\qquad
\mathrm{SE}\left(r_{\mathrm{HR}}\right)=\sqrt{\frac{1-r_{\mathrm{HR}}^{2}}{P-2}},
\end{equation}
where $\overline{\hat{h}}$ and $\overline{h}$ are the means over the $N$ clips.
The waveform correlation is taken within a clip, between the predicted and the reference photoplethysmogram, and then averaged,
\begin{equation}
\label{eq:pearson_wave}
r_i = \frac{\sum_{t=1}^{T}\bigl(\hat{s}_i[t]-\overline{\hat{s}_i}\bigr)\bigl(s_i[t]-\overline{s_i}\bigr)}
{\sqrt{\sum_{t=1}^{T}\bigl(\hat{s}_i[t]-\overline{\hat{s}_i}\bigr)^{2}}\;\sqrt{\sum_{t=1}^{T}\bigl(s_i[t]-\overline{s_i}\bigr)^{2}}},
\qquad
r_{\mathrm{wave}} = \bigl\langle r_i \bigr\rangle .
\end{equation}
It is $r_{\mathrm{wave}}$, not $r_{\mathrm{HR}}$, that every clip-level relationship in Sections~\ref{sec:static3_xai_results} and~\ref{sec:crosscondition_xai_results} calls the waveform Pearson correlation.

\emph{Equation~\eqref{eq:pearson_wave} is evaluated at a fitted lag rather than at zero.}
Each recording is synchronised on its own light-off trigger, so a video-to-photoplethysmogram synchronisation error is one constant per participant and recording.
We fit that constant by maximising the participant's mean clip correlation over $\pm1$~second and apply it to every clip of that recording, so that $r_{\mathrm{wave}}$ measures waveform agreement rather than synchronisation error.
Supplementary Table~\ref{tab:supp_waveform_lag} reports the correlation before and after the fit for every participant, the fitted lag, and a null in which the same fit is applied to mismatched participant pairs.
The fit is worth two to seven times its null on NCKU-rPPG and less than its null on UBFC-rPPG, whose fitted lags are zero for eight of twelve subjects and within two samples for eleven of them.
No other quantity in Table~\ref{tab:performance_results} changes under it, because a time shift leaves a spectrum's magnitude, and therefore every heart-rate and signal-to-noise measure, unaltered.

\citet{chen2026RhythmFormerPhotoplethysmographyEXPLIMED} evaluates the same correlation at zero lag, which is the convention that study states, so its archived per-clip value and ours are not the same quantity.
The difference is small in aggregate and concentrated in two subjects: the fit raises the participant-averaged UBFC-rPPG correlation from $0.785$ to $0.828$, of which those two supply nine tenths, gaining $0.22$ and $0.25$ against $0.04$ or less for the other ten.
Recomputed at the fitted lag, that study's clip-pooled rank correlations between skin coverage and waveform quality move from $+0.31$ to $+0.29$ for attention rollout and from $+0.24$ to $+0.19$ for Beyond Intuition, without changing which of its four attribution methods resolve.

The signal-to-noise ratio follows \citet{dehaan2014RobustnessRemotePPGPhysiolMeas} in the form the toolbox implements,
\begin{equation}
\label{eq:snr}
\mathrm{SNR}_i = 10\log_{10}\frac{\sum_{f}\bigl[U_i(f)\,S_i(f)\bigr]^{2}}{\sum_{f}\bigl[\bigl(1-U_i(f)\bigr)S_i(f)\bigr]^{2}},
\qquad
\mathrm{SNR} = \bigl\langle \mathrm{SNR}_i \bigr\rangle,
\end{equation}
where $S_i(f)$ is the power spectrum of the predicted waveform of clip $i$ and $U_i(f)$ is a binary template equal to one within $\pm6$~beats per minute of the reference fundamental and of its second harmonic and zero elsewhere.
The two sums do not run over the same support.
The denominator is confined to 0.75--2.5~Hz, that is 45--150~beats per minute, whereas the numerator is the template's own neighbourhoods wherever they fall.
Above a reference of 75~beats per minute the second harmonic lies beyond 2.5~Hz, outside that range and inside the stop band of the bandpass already applied, so for those clips the second-harmonic term contributes attenuated power rather than signal.
This is the common case rather than the exception: 75.2~\% of the 4001 NCKU-rPPG clips have a reference above 75~beats per minute, from 59.9~\% under rotation to 84.4~\% on Bike level~3.
The measure is therefore closer to a first-harmonic one than Equation~\eqref{eq:snr} suggests, in the same conservative direction as the resolution argument below.
\emph{$S_i(f)$ is not the spectrum in which the heart rate was located, and the difference matters enough to state.}
Equation~\eqref{eq:pearson_hr} reads the heart rate from a Welch spectrum whose segment is capped at 256 samples and which is zero-padded to $\mathrm{nfft}=200{,}000$;
Equation~\eqref{eq:snr} takes its own periodogram of the clip, zero-padded to the next power of two above the clip length and averaged over nothing.
The split is not an oversight.
Locating a peak rewards segment averaging, which lowers the variance of the location, whereas partitioning power against a fixed $\pm6$~beats-per-minute template rewards resolution, because a spectral lobe wider than the template spills the signal's own power into the noise band.
At a 256-frame clip the periodogram resolves 11.7~beats per minute, which is wider than the template, so the reported SNR is a conservative measure of waveform quality at this clip length rather than a neutral one.

Equation~\eqref{eq:snr} squares a quantity it calls a power spectrum, so it returns twice the value that a ratio of powers in decibels would give, which is the amplitude convention rather than the $10\log_{10}$ conventionally used for a ratio of powers.
The underlying ratio is therefore recovered as $10^{\mathrm{SNR}/20}$ and not as $10^{\mathrm{SNR}/10}$.
We retained this convention because it is the one implemented in the rPPG-Toolbox code the experiment used and the one behind the published SNR values we compare with~\citep{liu2023RPPGToolboxToolboxNeurIPS,zou2025RhythmFormerrPPGPatternRecognit}, so that changing it would make our values incomparable with theirs.

\paragraph{Uncertainty.}
\emph{Our standard errors depart from the toolbox in one respect.}
It computes $\sigma/\sqrt{n}$ over $n$ test samples, and $\sqrt{(1-r^2)/(n-2)}$ for the Pearson correlation, both of which treat every test sample as independent~\citep{liu2023RPPGToolboxToolboxNeurIPS}.
With one estimate per clip the samples are not independent: each participant contributes 12 to 71 clips, and heart rate, skin tone, and facial geometry are properties of a person.
A standard error taken over clips would understate the uncertainty by roughly the square root of the number of clips per participant.
Equation~\eqref{eq:clustered_mean} therefore averages within participant first and takes the standard error over the resulting participant means, and Equation~\eqref{eq:pearson_hr} substitutes $P$ for $n$ in the toolbox formula, so that the effective sample size is the number of participants in every column of Table~\ref{tab:performance_results}.
The same treatment applies to the skin-coverage summaries, which are clip-level for the same reason.
A pre-attention-to-refined-attention difference is taken between two means of the same clips, so we formed the difference clip by clip before averaging, which gives it a smaller standard error than either mean it is built from.
This uncertainty describes between-participant variation for one trained seed; it is neither numerical calculation error nor retraining variability.

We report every estimate by the convention of letting the uncertainty set the precision: round the standard error to two significant figures, then round the estimate to that same decimal place.
We applied the same procedure to the UBFC-rPPG reproduction of \citet{chen2026RhythmFormerPhotoplethysmographyEXPLIMED}, so that row of Table~\ref{tab:performance_results} is comparable with the scenario rows.
Its archived clip evidence records one absolute heart-rate error and one waveform correlation per clip and neither a reference heart rate nor a signal-to-noise ratio, so we did not read that row off the evidence.
We cut the archived per-subject prediction and reference waveforms into the same 141 clips of 160 frames at 30~Hz and passed each through the identical estimation used for the NCKU-rPPG scenarios, which yields every column from one code path.
Recomputed and archived per-clip errors agree to within 0.03~beats per minute, two orders of magnitude below the precision reported, and the recomputation is checked against the archive whenever the row is generated.

\emph{The two datasets are matched in clip duration but not in every transform parameter, and the residue falls on one column only.}
A UBFC-rPPG clip is 160 frames at 30~Hz and an NCKU-rPPG clip 256 frames at 50~Hz, that is 5.33 against 5.12~s, so the resolving power of the two spectra differs by 4~\%, at 11.25 against 11.72~beats per minute.
Two parameters are nonetheless set by the clip rather than chosen, and differ.
The Welch segment is $\min(N-1,256)$, which the UBFC-rPPG clip does not reach, so it is 159 there against 256 here.
The SNR periodogram is padded to the next power of two above the clip, which pads UBFC-rPPG by a factor 1.6 and NCKU-rPPG not at all.
Neither can be equalised by a choice of window: matching the Welch segment would need 265, above the cap, and matching the padded length would need 427, which is not a power of two.
Measured, the padding difference is worth about 1~decibel and runs against UBFC-rPPG, so the SNR advantage reported for it in Table~\ref{tab:performance_results} is if anything understated;
the heart-rate columns are unaffected, because they depend on the peak's location rather than on how the power around it is partitioned.

\subsection{Explainable artificial intelligence attribution methods}
\label{sec:xai_attribution_methods}

We extracted attribution from all six Periodic Sparse Attention (PSA) blocks in the three Temporal Periodic Transformer (TPT) levels, with two blocks per level.
Pre-attention produced a $4\times4$ spatial-region map, whereas refined attention produced an $8\times8$ spatial-token map.
For raw attention, attention rollout, and attention flow, we completely collapsed the temporal query and key axes within every block before any aggregation across blocks or TPT levels.
The resulting spatial matrices retained $4\times4$ pre-attention or $8\times8$ refined spatial coordinates, but no temporal coordinate.
These three methods therefore did not use a $T=20$ temporal rollout.
Beyond Intuition retained the temporal coordinate during cross-layer aggregation, mapped all six refined-attention blocks to a common $T=20$ representation, and pooled time only after producing the integrated attribution.
Its reasoning-feedback term used 20 integrated-gradient steps from a standardised-space zero baseline to the standardised input clip.
Zero in this space corresponded to the per-clip mean-intensity state.

\paragraph{Raw attention.}
Raw attention provides a layer-local explanation by measuring how strongly each source position is used as a key, without propagating attribution across layers.
Because RhythmFormer has no classification token, we used the received attention obtained by summing the head-averaged attention matrix $A^{(l)}$ over query positions:
\begin{equation}
  r_k^{(l)}=\sum_q A_{qk}^{(l)}.
  \label{eq:xai_raw_attention}
\end{equation}
Here, $k$ indexes the source position and $l$ indexes a PSA block; the temporal query and key axes were collapsed before the resulting spatial map was aggregated across blocks.

\paragraph{Attention rollout.}
Attention rollout~\cite{abnar2020QuantifyingTransformersCOLING} estimates cross-layer information propagation by augmenting each attention matrix with the residual connection and multiplying the matrices in forward order.
With residual weight $1/2$, the augmented matrix and input-side salience were defined as
\begin{equation}
  \widehat{A}^{(l)}=0.5I+0.5A^{(l)},
  \qquad
  \widetilde{A}=\widehat{A}^{(L)}\cdots\widehat{A}^{(1)},
  \qquad
  \psi_k=\sum_q\widetilde{A}_{qk}.
  \label{eq:xai_attention_rollout}
\end{equation}
The rollout was applied to the temporally collapsed spatial matrices, so no cross-stage temporal downsampling was used.

\paragraph{Attention flow.}
Attention flow~\cite{abnar2020QuantifyingTransformersCOLING} treats the stacked attention matrices and residual connections as a layered capacity graph.
Each source position $i$ was assigned the total maximum flow that reached every output position $j$ across the graph:
\begin{equation}
  \phi_i=\sum_j\operatorname{MaxFlow}\!\left(v_i^{(0)},v_j^{(L)}\right).
  \label{eq:xai_attention_flow}
\end{equation}
The graph used the same temporally collapsed $4\times4$ pre-attention or $8\times8$ refined-attention spatial nodes as raw attention and rollout, with the residual connection represented as diagonal capacity.

\paragraph{Beyond Intuition.}
Beyond Intuition~\cite{chen2023RethinkingAttributionsTransMachLearnRes} combines structural propagation with a gradient-based reasoning-feedback mask.
Its final attribution was the Hadamard product of a value-projection-corrected rollout $P^{(L)}$ and an integrated-gradient mask $F^c$:
\begin{equation}
  T=P^{(L)}\odot F^c,
  \qquad
  F^c=\operatorname{ReLU}\!\left(-\frac{1}{N_{\mathrm{steps}}}
  \sum_{s=1}^{N_{\mathrm{steps}}}
  \nabla_{A^{(L)}}\mathcal{L}\!\left(\frac{s}{N_{\mathrm{steps}}}X\right)\right),
  \label{eq:xai_beyond_intuition}
\end{equation}
where the negative gradient is used because the regression loss is minimised.
The correction scales the rollout contribution of source token $j$ by
$\alpha_j=\lVert z_jW\rVert_2/\lVert z_j\rVert_2$ before forming $P^{(L)}$.
Unlike the other three methods, Beyond Intuition retained the temporal coordinate, unified the refined-attention blocks at $T=20$, and pooled time only after computing the integrated attribution.

Mixed-precision attribution for the rotation scenario produced non-finite values.
We therefore generated the formal rotation attributions in FP32.
This attribution-only change did not alter the rotation checkpoint, model parameters, or training procedure.

\subsection{Anatomical plausibility and perturbation faithfulness}
\label{sec:xai_evaluation}

We quantified anatomical plausibility with skin coverage.
A pretrained Bilateral Segmentation Network (BiSeNet) face parser~\cite{yu2018BiSeNetSegmentationECCV} supplied a binary mask containing only the face-skin and nose classes.
Each attribution map was min--max normalised before bicubic upsampling to $128\times128$ pixels.
For normalised attribution $h_i$ and binary skin mask $m_i$ at pixel $i$, skin coverage was
\begin{equation}
  C_{\mathrm{skin}}
  =\frac{\sum_i h_i m_i}{\sum_i h_i}.
  \label{eq:xai_skin_coverage}
\end{equation}
Skin coverage measured anatomical plausibility and was not treated as a faithfulness measure.

We measured perturbation faithfulness with the Salience-guided Faithfulness Coefficient (SaCo)~\cite{wu2024FaithfulnessTransformerCVPR}.
For each clip, we sorted the spatial positions by salience and divided them into $K=8$ equally sized groups $G_1,\ldots,G_8$.
Group salience was the summed salience, $s(G_i)=\sum_{p\in G_i}h_p$.
Each group was masked separately in every video frame by replacing its pixels with zero in standardised space, equivalent to the per-clip per-channel mean intensity before standardisation.
The perturbation impact was the waveform mean absolute error
\begin{equation}
  d(G_i)=\operatorname{MAE}\!\left(\Phi(x),\Phi(x_{G_i})\right),
  \label{eq:xai_saco_impact}
\end{equation}
where $\Phi(x)$ and $\Phi(x_{G_i})$ were the waveforms predicted from the original and independently masked clips.
For every pair $i<j$, $w_{ij}=s(G_i)-s(G_j)$ when $d(G_i)\geq d(G_j)$ and $w_{ij}=-[s(G_i)-s(G_j)]$ otherwise, and
\begin{equation}
  F=\frac{\sum_{i<j}w_{ij}}{\sum_{i<j}|w_{ij}|}\in[-1,1].
  \label{eq:xai_saco}
\end{equation}
Beyond Intuition supplied its refined $8\times8$ attribution map to SaCo, whereas raw attention, attention rollout, and attention flow supplied $4\times4$ source maps.

The attribution visualisations in Figure~\ref{fig:static3_xai_visualization} and Supplementary Section~\ref{sec:supp_xai} averaged all test clips within their scenario.
The violin and scatter plots used clip-level observations (Figures~\ref{fig:combined_xai_distributions} and~\ref{fig:combined_xai_scatter}).
For the skin-coverage violins, each clip was first averaged across TPT$_1$--TPT$_3$.
For the scatter plots, the attention-only methods used refined TPT$_3$ skin coverage and Beyond Intuition used its single coverage value.
The combined split violins in Figure~\ref{fig:combined_xai_distributions} placed the 141 UBFC-rPPG clips on each left half and the eight NCKU-rPPG scenarios on each right half.
We normalised each scenario's kernel density estimate independently rather than pooling the eight NCKU-rPPG scenarios into one density.
We plotted every available clip with deterministic half-constrained jitter and omitted the central boxplots because one box could not represent the nine data groups.
Position and explicit sublabels distinguished pre-attention from refined attention; colour represented only the dataset or scenario.
The combined scatter plots in Figure~\ref{fig:combined_xai_scatter} used circles for NCKU-rPPG clips and crosses for UBFC-rPPG clips and retained the complete linear data ranges.
Within each attribution-method panel, we fitted one ordinary least-squares line to all displayed clips and reported the corresponding all-clips Pearson correlation, nominal two-sided $p$ value, and sample size.
The cross-method summaries pooled all test clips and all three TPT levels before calculating each mean.

\subsection{Cross-dataset transfer and scenario-level reliability comparisons}
\label{sec:xai_comparisons}

We defined two complementary levels of comparison.
At the cross-dataset level, we compared Static level~3 with the published UBFC-rPPG evaluation of \citet{chen2026RhythmFormerPhotoplethysmographyEXPLIMED} because their heart-rate accuracy was similar, while retaining their different participants, cameras, preprocessing, frame rates, and checkpoints.
We compared the median skin coverage and SaCo rankings, pre-attention-to-refined-attention changes, and clip-level correlation structures without treating the two evaluations as a controlled cross-dataset experiment.
We related three clip-level measures of model performance, the heart-rate mean absolute error, the waveform correlation of Equation~\eqref{eq:pearson_wave}, and the signal-to-noise ratio of Equation~\eqref{eq:snr}, to two clip-level attribution summaries, skin coverage and SaCo.
That yielded six relationships, and a seventh between the two attribution summaries themselves, SaCo against skin coverage.
For each attribution method we calculated all seven separately within each of the eight NCKU-rPPG scenarios and within UBFC-rPPG, which yielded 252 coefficients in total.
The signal-to-noise ratio was recorded per clip in the NCKU-rPPG evidence and was recomputed from the archived waveforms for UBFC-rPPG, whose clip evidence did not carry it, by the same estimation used for every other clip.

\emph{Each correlation was taken inside a participant and averaged over participants, not pooled over clips.}
Pooling every clip of every participant would have measured mostly the differences between participants, because heart rate, skin tone and facial geometry are properties of a person rather than of a clip, and a coefficient built from them would not have shown whether a clip with more attributed skin carried a better waveform.
We therefore computed the Spearman coefficient within each participant, averaged the coefficients through the Fisher transform, and tested the mean against zero with a one-sample $t$ test over participants.
Participants were independent, so this $p$ value was inferential rather than descriptive, which no test pooled over nested clips could have been.
The difference was not cosmetic: when clips were pooled, the waveform Pearson--skin coverage relationship of Static level~3 was $-0.41$ with $p<10^{-12}$ for all four methods alike, whereas the same data taken within participants yielded coefficients between $-0.16$ and $+0.13$, none of them resolved.
A participant contributed a coefficient only when it had at least four clips in the relationship, and a relationship was reported only when at least three participants contributed.
The same estimator annotated every panel of Supplementary Figure~\ref{fig:combined_xai_scatter}, so every correlation reported as a finding was taken within participants;
a panel spanning all eight conditions and UBFC summarised the display rather than reproducing one table row and, when restricted to a single condition, returned that row exactly.
The two clip-pooled coefficients quoted anywhere in the manuscript were the $-0.41$ above and the recomputation of the clip-pooled values of \citet{chen2026RhythmFormerPhotoplethysmographyEXPLIMED} in Section~\ref{sec:performance_evaluation}; both were labelled as pooled where they appeared, and neither was reported as a finding.
The correlation calculations used the same aligned arrays as the corresponding scatter panels and never pooled clips across the nine data groups.
We reported nominal $p$ values descriptively because clips were nested within participants.

At the scenario level, each of the eight NCKU-rPPG scenarios was one observation.
For each attribution method, we calculated the median SaCo across its evaluated clips within each scenario and related those eight medians to scenario-level heart-rate mean absolute error (MAE), waveform Pearson correlation, and signal-to-noise ratio (SNR) using Spearman's rank correlation coefficient.
If SaCo tracked model reliability, its expected directions were negative with MAE and positive with waveform Pearson correlation and SNR.
Because $n=8$ scenarios were independently trained conditions rather than repeated samples from a common intervention, we report the coefficients descriptively without significance tests or causal interpretation.

\newcommand{\xaisuppspearmanreference}{Supplementary Table~\ref{tab:supp_xai_spearman_full}}

\section{Results}
\label{sec:results}

\subsection{Performance across illumination and motion}
\label{sec:performance_results}

Table~\ref{tab:performance_results} reports clip-level performance for the eight final-epoch checkpoints.
Every NCKU-rPPG estimate and standard error was calculated from the clips of the 12 held-out participants in the corresponding scenario, averaged within participant before averaging across participants.
The UBFC-rPPG row comes from the study of \citet{chen2026RhythmFormerPhotoplethysmographyEXPLIMED}, which trained and tested RhythmFormer on UBFC-rPPG alone.
We did not repeat that experiment.
We calculated its estimates and standard errors from the 141 clips of its archived waveforms by the same procedure.

\begin{table}[htbp]
\centering
\caption[RhythmFormer performance across NCKU-rPPG conditions]{RhythmFormer performance across the eight NCKU-rPPG conditions, and on UBFC-rPPG from the archived waveforms of \protect\citet{chen2026RhythmFormerPhotoplethysmographyEXPLIMED}.
One heart rate is estimated per clip, 5.12~s for the NCKU-rPPG rows and approximately 5.33~s for UBFC-rPPG, and each row reports the participant-weighted estimate $\pm$ its standard error across the 12 held-out participants, as defined in Section~\ref{sec:performance_evaluation}.
$r_{\mathrm{HR}}$ is the correlation between the estimated and reference heart rates of the condition's clips, Equation~\eqref{eq:pearson_hr}, and $r_{\mathrm{wave}}$ the mean within-clip correlation between the predicted and reference waveforms, Equation~\eqref{eq:pearson_wave}.
The UBFC-rPPG row is recomputed from that study's archived waveforms rather than read from its published summary, so that its analysis unit matches the scenario rows.}
\label{tab:performance_results}
\footnotesize
\setlength{\tabcolsep}{2.6pt}
\begin{tabular}{lcccccc}
\toprule
Scenario & MAE (bpm) & RMSE (bpm) & MAPE (\%) & $r_{\mathrm{HR}}$ & $r_{\mathrm{wave}}$ & SNR (dB) \\
\midrule
UBFC-rPPG~\cite{chen2026RhythmFormerPhotoplethysmographyEXPLIMED} & $0.96\pm0.16$ & $1.37\pm0.28$ & $0.99\pm0.18$ & $0.991\pm0.041$ & $0.828\pm0.022$ & $8.54\pm0.97$ \\
\midrule

Static level~1 & $6.2\pm1.9$ & $13.0\pm3.0$ & $7.3\pm2.2$ & $0.57\pm0.26$ & $0.516\pm0.076$ & $-1.7\pm2.5$ \\
Static level~3 & $\mathbf{1.96\pm0.34}$ & $\mathbf{3.74\pm0.75}$ & $\mathbf{2.36\pm0.42}$ & $\mathbf{0.968\pm0.079}$ & $\mathbf{0.626\pm0.037}$ & $\mathbf{5.7\pm1.5}$ \\
Static level~5 & $3.9\pm1.1$ & $9.0\pm2.1$ & $4.5\pm1.3$ & $0.80\pm0.19$ & $0.588\pm0.055$ & $3.6\pm2.3$ \\
Speak & $8.9\pm2.6$ & $15.4\pm4.1$ & $9.4\pm2.1$ & $0.43\pm0.29$ & $0.324\pm0.045$ & $-4.7\pm2.3$ \\
Rotate & $12.9\pm2.6$ & $18.3\pm2.8$ & $15.6\pm2.8$ & $0.22\pm0.31$ & $0.300\pm0.043$ & $-11.0\pm2.5$ \\
Bike level~1 & $18.6\pm5.5$ & $30.3\pm7.6$ & $18.8\pm4.0$ & $-0.15\pm0.31$ & $0.235\pm0.044$ & $-10.2\pm3.3$ \\
Bike level~3 & $16.1\pm2.3$ & $24.3\pm2.6$ & $17.9\pm2.3$ & $0.12\pm0.31$ & $0.232\pm0.047$ & $-11.9\pm2.0$ \\
Bike level~5 & $15.1\pm3.5$ & $23.9\pm3.9$ & $17.5\pm3.6$ & $0.25\pm0.31$ & $0.319\pm0.048$ & $-8.9\pm3.2$ \\
\bottomrule

\end{tabular}
\end{table}

Among the static conditions, level~3 had the lowest error for MAE, RMSE, and MAPE, the highest heart-rate correlation, and the highest SNR.
Static level~5 ranked second on those measures and static level~1 last, and level~1 was the only static condition with a negative mean SNR.
Level~3 also had the highest waveform correlation, but the three are unevenly spaced: level~1 falls 7.4~dB below level~3 in SNR and 0.11 below it in waveform correlation, whereas level~5 falls only 2.1~dB and 0.04 below it. %

Speaking and head rotation both degraded performance relative to the static level~3 model trained and tested at the same nominal illumination level.
Rotation degraded both measures further than speaking, but not evenly: its heart-rate error was about 1.4 times as large, while its SNR fell a further 6.3~dB, so the two conditions are separated more by waveform quality than by heart-rate error.

The three cycling conditions were the weakest of the eight on every column, with mean absolute errors of 15 to 19~beats per minute and mean SNR from $-8.9$ to $-11.9$~dB.
Bike level~5 was nominally the best of them on all six measures, but which of the other two was worst depends on the measure: bike level~1 had the largest errors and the lowest heart-rate correlation, and bike level~3 the lowest waveform correlation and the lowest SNR.
Every pairwise difference among the three is within twice the standard errors of the two values it separates, so the three are not separated by these data.

The measures order the eight scenarios closely: across scenario-level values, MAE ranks against the heart-rate correlation at $\rho=-0.98$ ($p<.001$), against the waveform correlation at $\rho=-0.95$ ($p<.001$), and against SNR at $\rho=-0.88$ ($p=.004$), while the waveform correlation ranks against SNR at $\rho=+0.98$ ($p<.001$).
The two correlations are still not interchangeable.
Static level~3 pairs a heart-rate correlation of $0.968\pm0.079$ with a waveform correlation of $0.626\pm0.037$, because a dominant frequency can be recovered from a waveform that matches the reference only moderately well point by point.
Supplementary Figure~\ref{fig:supp_waveform_examples} shows a representative clip from static level~3 and one from bike level~1.

The heart rate is easier to predict on UBFC-rPPG than on our data: it stands ahead of Static level~3 on all six measures, at less than half the mean absolute error and 2.8~dB more SNR.
The differing clip sample counts do not create that SNR gap, because Equation~\eqref{eq:snr} partitions the power of a periodogram padded to the next power of two above the clip, which pads the 160-sample UBFC-rPPG clip by a factor 1.6 and leaves the 256-sample NCKU-rPPG clip unpadded, and that costs UBFC-rPPG about 1~dB (Section~\ref{sec:performance_evaluation}).

\subsection{Attribution summaries for Static level 3 and UBFC-rPPG}
\label{sec:static3_xai_results}

\begin{figure}[p]
\centering
\includegraphics[width=0.72\textwidth]{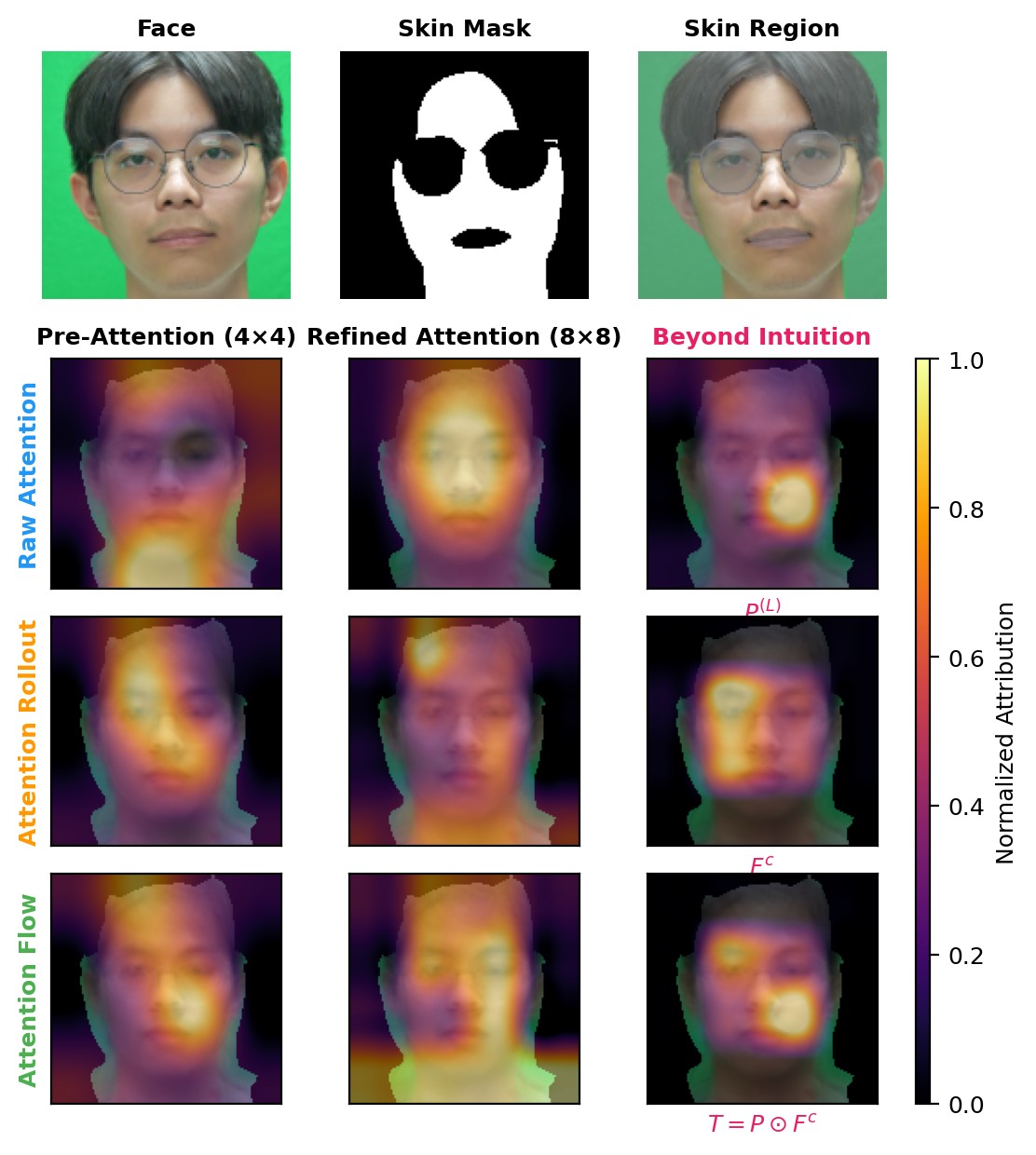}
\caption{Static level~3 attribution visualisation.
The published representative face appears only in the top panels labelled \emph{Face}, \emph{Skin Mask}, and \emph{Skin Region}.
The attribution maps below those panels, the average face, and the displayed summary statistics use the complete Static level~3 test set rather than that representative alone.}
\label{fig:static3_xai_visualization}
\end{figure}

Static level~3 is the NCKU-rPPG condition closest to the UBFC-rPPG recording setting, so we examine the two together before turning to the whole set of conditions.
Figure~\ref{fig:static3_xai_visualization} shows the Static level~3 attributions, and Table~\ref{tab:static3_xai_coverage} the pre-attention and refined-attention skin-coverage means of both evaluations.
Most of the attribution falls on the mid-face, but the four methods place their maxima differently.
Raw attention peaks on the neck below the chin before refinement and on the forehead and the nose after it;
attention rollout and attention flow hold a broad ridge over one cheek, on opposite sides of the face, that after refinement spreads onto the hairline and the surrounding background;
and Beyond Intuition contracts to a single spot on one cheek.

\begin{figure}[p]
\centering
\begin{subfigure}[t]{\textwidth}
\centering
\includegraphics[width=0.78\textwidth]{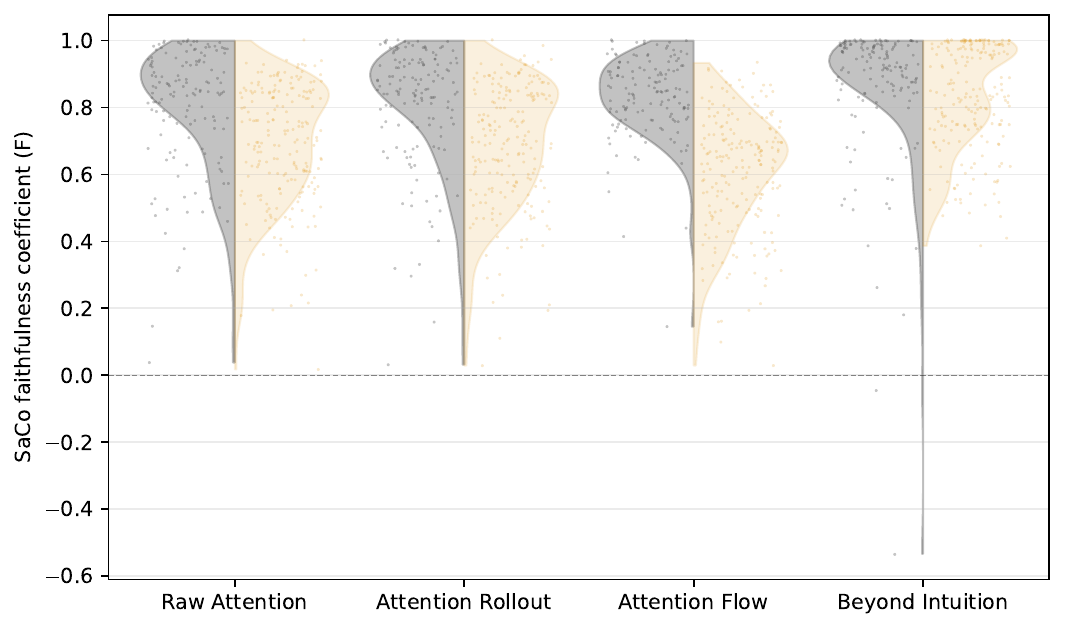}
\caption{SaCo faithfulness coefficient $F$.}
\end{subfigure}
\par\medskip
\begin{subfigure}[t]{\textwidth}
\centering
\includegraphics[width=\textwidth]{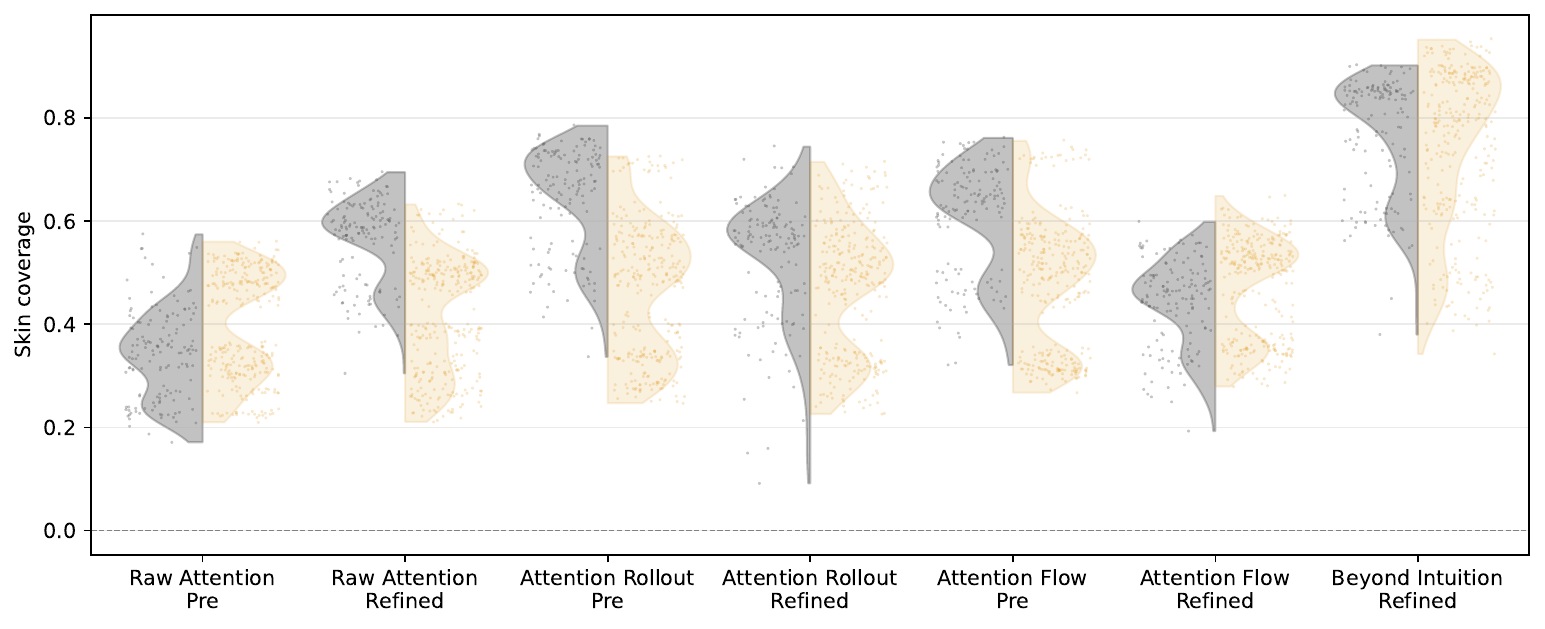}
\caption{Skin coverage, before and after refinement.}
\end{subfigure}
\caption[Clip-level SaCo and skin coverage for Static level~3 and UBFC-rPPG]{Clip-level split-violin summaries for Static level~3 and UBFC-rPPG.
Each left half is UBFC-rPPG (dark grey) and each right half Static level~3 (orange), the two normalised independently rather than pooled.
All available clips are plotted as points on their own half: UBFC-rPPG contributes 141 SaCo and 141 skin-coverage observations per method, and Static level~3 contributes 200 and 289.
In~(b) each clip is averaged across TPT$_1$--TPT$_3$ for the attention-only methods, and Beyond Intuition has a refined value only.}
\label{fig:static3_ubfc_violins}
\end{figure}

Figure~\ref{fig:static3_ubfc_violins} shows the clip-level SaCo and skin-coverage distributions of the two evaluations, the coverage means of which Table~\ref{tab:static3_xai_coverage} summarises.
Beyond Intuition has the highest median SaCo and skin coverage in both evaluations, but the other three methods differ between them.
For Static level~3, refinement changed skin coverage by less than its own standard error for every attention-only method, so no change is resolved by these data.
The methods differed from each other more than refinement changed any of them: raw attention covered the least skin and attention flow the most, a spread of about 0.08 before refinement and 0.05 after.
UBFC-rPPG behaved differently in three respects.
Refinement moved coverage by 0.12 to 0.22 there, seven to fourteen times the corresponding standard error, against nothing resolved on Static level~3.
It moved the methods in opposite directions, raw attention gaining coverage while attention rollout and attention flow lost it, whereas on Static level~3 all three changes were indistinguishable from zero.
And it reordered them: raw attention has the lowest refined coverage on Static level~3 and the highest on UBFC-rPPG, while attention flow moves from highest to lowest.
The same implementation of the three methods therefore give a refinement step that is inert on one evaluation and decisive on the other.
The whole set of conditions locates that difference in magnitude rather than in direction, however.
Across all nine conditions refinement raises the coverage of raw attention and lowers that of attention flow without exception, and lowers attention rollout in eight of the nine, so the directions on UBFC-rPPG are the directions everywhere;
what is peculiar to Static levels~1 and~3 is that the changes are too small to resolve (Supplementary Table~\ref{tab:supp_xai_cross_method}).

\begin{table}[htbp]
\centering
\caption[Skin coverage for Static level~3 and UBFC-rPPG]{Mean skin coverage for Static level~3 and UBFC-rPPG, pooled over all held-out clips and TPT$_1$--TPT$_3$, as estimate $\pm$ its standard error across the 12 test participants.
The difference is between the two means of the same clips and therefore carries its own, smaller standard error.
Beyond Intuition has no pre-attention/refined-attention pair and is therefore not included.
The UBFC-rPPG rows are calculated from the 141 clips of \protect\citet{chen2026RhythmFormerPhotoplethysmographyEXPLIMED}.}
\label{tab:static3_xai_coverage}
\small
\begin{tabular}{lrrr}
\toprule
Method & Pre & Refined & Difference \\
\midrule
\multicolumn{4}{l}{UBFC-rPPG} \\
Raw Attention & $0.342\pm0.025$ & $0.566\pm0.023$ & $0.224\pm0.016$ \\
Attention Rollout & $0.644\pm0.028$ & $0.527\pm0.031$ & $-0.117\pm0.017$ \\
Attention Flow & $0.609\pm0.030$ & $0.442\pm0.023$ & $-0.167\pm0.016$ \\
\midrule
\multicolumn{4}{l}{Static level~3} \\
Raw Attention & $0.398\pm0.031$ & $0.416\pm0.033$ & $0.018\pm0.018$ \\
Attention Rollout & $0.465\pm0.039$ & $0.465\pm0.038$ & $0.0004\pm0.0068$ \\
Attention Flow & $0.474\pm0.038$ & $0.470\pm0.029$ & $-0.005\pm0.014$ \\
\bottomrule

\end{tabular}
\end{table}

\begin{table}[htbp]
\centering
\caption{Two-sided Spearman summaries for Static level~3 and UBFC-rPPG, taken within each participant and averaged over participants.
Each cell gives $\rho; p$, where $p$ comes from a one-sample $t$ test over participants and is therefore inferential rather than descriptive.
Cells in bold satisfy both $p<0.05$ and $|\rho|\ge0.10$, the condition Table~\ref{tab:xai_direction_counts} counts as directional.
The first sample-size pair in each dataset label gives the number of skin-only clips followed by the number of clips in relationships involving SaCo;
the second gives the participants contributing to each, which differ because the SaCo run does not cover every held-out participant.}
\label{tab:static3_ubfc_spearman}
\scriptsize
\setlength{\tabcolsep}{1.8pt}
\begin{tabular}{llccccccc}
\toprule
Dataset & Method & MAE--skin & MAE--SaCo & Pearson--skin & Pearson--SaCo & SNR--skin & SNR--SaCo & SaCo--skin \\
\midrule
\multirow{4}{*}{\makecell[l]{Static level 3\\($n=289/200$)\\($P=12/9$)}} & raw & $\mathbf{-0.14; .020}$ & $0.01; .953$ & $0.13; .051$ & $0.11; .293$ & $-0.06; .390$ & $0.02; .794$ & $0.09; .427$ \\
 & rollout & $0.04; .508$ & $-0.02; .821$ & $0.02; .750$ & $0.12; .284$ & $0.00; .970$ & $0.01; .874$ & $0.03; .853$ \\
 & flow & $0.09; .111$ & $-0.00; .944$ & $0.05; .422$ & $0.02; .864$ & $-0.05; .612$ & $0.04; .764$ & $-0.13; .125$ \\
 & BI & $0.06; .513$ & $-0.19; .344$ & $-0.16; .086$ & $0.02; .809$ & $-0.08; .218$ & $\mathbf{0.19; .044}$ & $-0.13; .264$ \\
\midrule
\multirow{4}{*}{\makecell[l]{UBFC-rPPG\\($n=141/141$)\\($P=12/12$)}} & raw & $-0.05; .714$ & $-0.01; .853$ & $-0.12; .372$ & $0.01; .914$ & $-0.08; .471$ & $-0.03; .821$ & $-0.08; .656$ \\
 & rollout & $-0.09; .405$ & $-0.05; .490$ & $\mathbf{0.32; .047}$ & $0.06; .495$ & $0.15; .113$ & $0.01; .931$ & $0.01; .869$ \\
 & flow & $-0.07; .326$ & $0.02; .848$ & $\mathbf{0.21; .041}$ & $0.06; .533$ & $-0.01; .932$ & $-0.03; .796$ & $0.14; .066$ \\
 & BI & $-0.20; .073$ & $-0.02; .800$ & $0.25; .062$ & $0.05; .659$ & $0.13; .145$ & $0.04; .773$ & $\mathbf{0.28; .007}$ \\
\bottomrule

\end{tabular}
\end{table}

Table~\ref{tab:static3_ubfc_spearman} summarises all five clip-level relationships for Static level~3 and UBFC-rPPG.
The complete nine-group results appear in \xaisuppspearmanreference.
Once the participant is held fixed, almost nothing survives in either evaluation.
Of these twenty-eight relationships, two are directional on Static level~3, the MAE--skin coverage of raw attention at $\rho=-0.14$ ($p=.020$) and the SNR--SaCo of Beyond Intuition at $\rho=+0.19$ ($p=.044$), and three on UBFC-rPPG: the waveform Pearson--skin coverage of attention rollout at $\rho=+0.32$ ($p=.047$) and of attention flow at $\rho=+0.21$ ($p=.041$), and the SaCo--skin coverage of Beyond Intuition at $\rho=+0.28$ ($p=.007$).
All significant but at modest significance levels.
The remaining twenty-three are unresolved.
The two evaluations therefore do not disagree so much as fail to establish much in either.
What the near-absence of resolved relationships establishes is weaker but firmer than a disagreement between the two: within a participant, neither skin coverage nor SaCo predicts how well the model does on a given clip.

\subsection{Attribution summaries across the nine conditions}
\label{sec:crosscondition_xai_results}

\begin{figure}[p]
\centering
\begin{subfigure}[t]{\textwidth}
\centering
\includegraphics[width=0.78\textwidth]{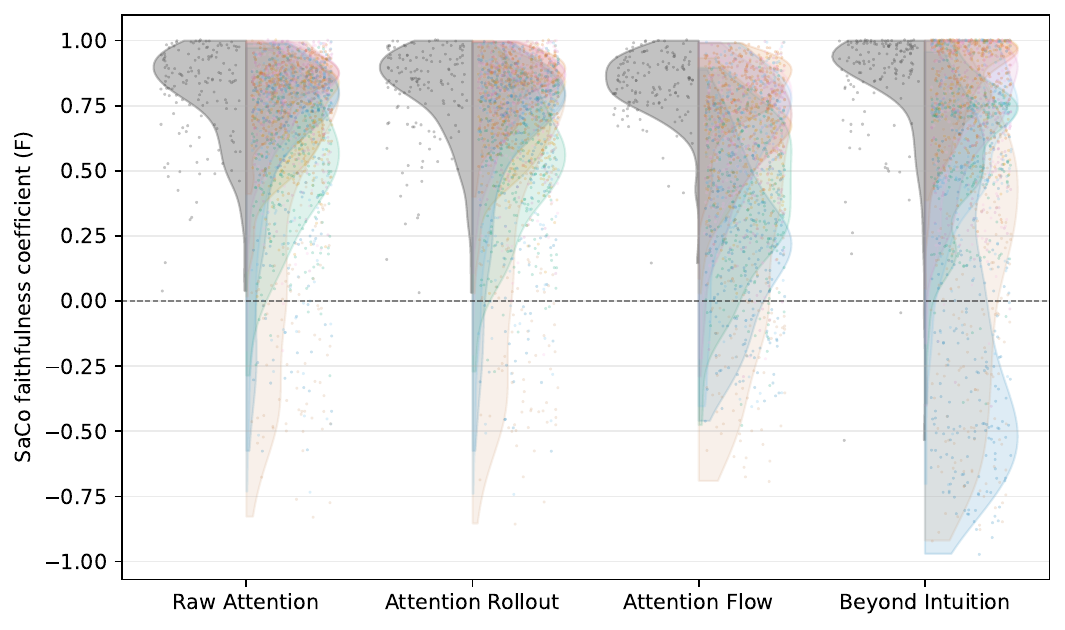}
\caption{SaCo, with all eight NCKU-rPPG scenarios on the right.}
\end{subfigure}
\par\medskip
\begin{subfigure}[t]{\textwidth}
\centering
\includegraphics[width=\textwidth]{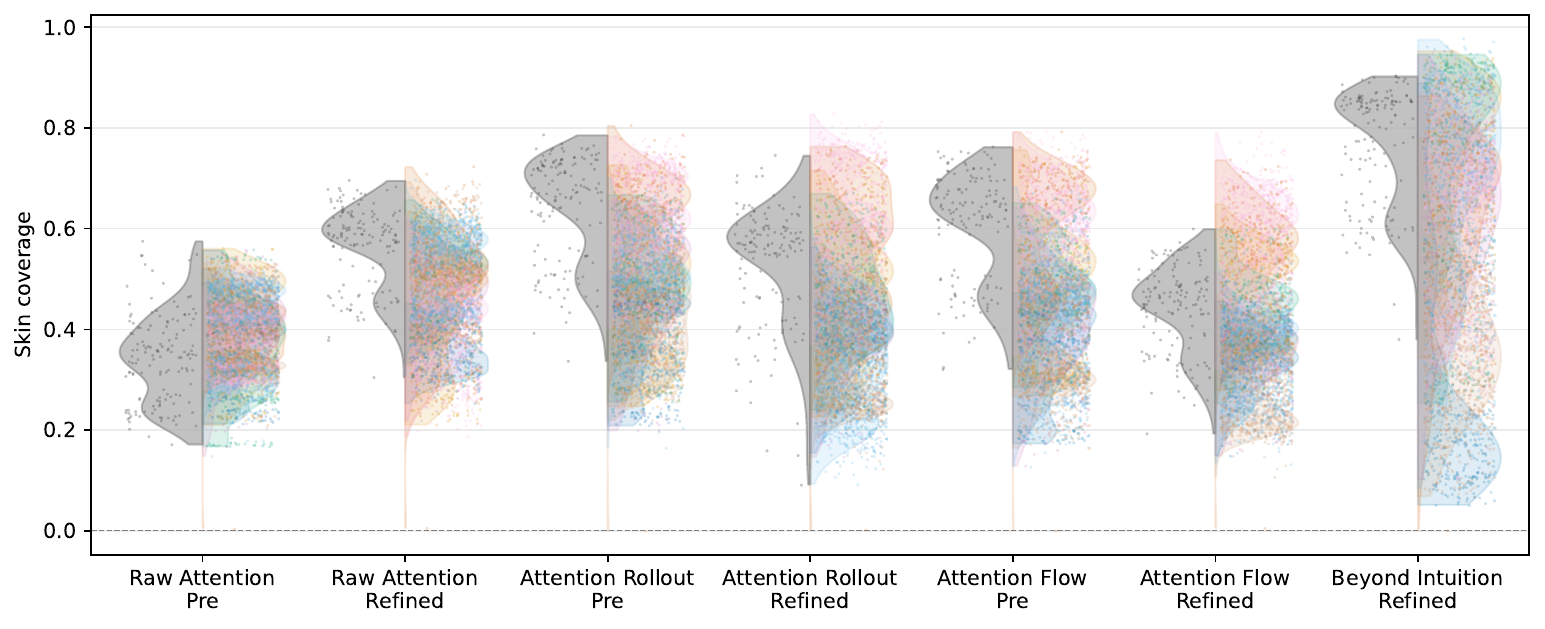}
\caption{Skin coverage, with all eight NCKU-rPPG scenarios on the right.}
\end{subfigure}
\caption{Combined clip-level split-violin summaries for raw attention, attention rollout, attention flow, and Beyond Intuition.
Each left half shows UBFC-rPPG, and each right half overlays eight independently normalised NCKU-rPPG scenario kernel densities without pooling them.
All available clips are plotted on their corresponding half: UBFC-rPPG contains 141 skin-coverage and 141 SaCo observations per method; Static levels~1, 3, and~5 contain 289 skin-coverage observations each; Speak contains 288; Rotate contains 287; Bike levels~1, 3, and~5 contain 853 each; and each NCKU-rPPG scenario contains 200 SaCo observations per method.
For attention-only skin coverage, each clip is averaged across TPT$_1$--TPT$_3$ before plotting; Beyond Intuition has only a refined value.
Colours denote Static level~1 (blue, \texttt{\#0173B2}), Static level~3 (orange, \texttt{\#DE8F05}), Static level~5 (green, \texttt{\#029E73}), Speak (red-orange, \texttt{\#D55E00}), Rotate (purple, \texttt{\#CC78BC}), Bike level~1 (brown, \texttt{\#CA9161}), Bike level~3 (pink, \texttt{\#FBAFE4}), Bike level~5 (light blue, \texttt{\#56B4E9}), and UBFC-rPPG (dark grey, \texttt{\#4D4D4D}).}
\label{fig:combined_xai_distributions}
\end{figure}

Figure~\ref{fig:combined_xai_distributions} compares the clip-level skin-coverage and SaCo distributions of the eight NCKU-rPPG scenarios with UBFC-rPPG.
Table~\ref{tab:xai_median_iqr} reports the medians of Figure~\ref{fig:combined_xai_distributions} for every condition, with the interquartile range as the measure of spread.
The skin-coverage rows use every held-out clip of the condition and the SaCo rows the clips the SaCo run evaluated, so the two differ in sample size where the run did not cover the whole condition.
Beyond Intuition holds the highest median refined skin coverage in seven of the nine conditions and the highest median SaCo in Static level~3, Static level~5, Speak and UBFC-rPPG, so where the recording is clean it both attributes more inside the skin region and orders the clips more faithfully than any attention-only method.
Negative SaCo is confined to a minority of outlier clips in 34 of the 36 condition-method pairs;
the exception is Beyond Intuition on Static level~1, where 117 of the 200 clips fall below zero and the median itself is $-0.178$, and Bike level~1, where 41\% fall below zero.
Both measures are lowest under the dimmest illumination.
Within the static triple and the cycling triple, which differ only in illuminance, the 40-lux member has the lowest median refined coverage for seven of the eight method--triple pairs and the lowest median SaCo for six of them, the exceptions being raw attention for coverage in the cycling triple and raw attention and attention rollout for SaCo in the static triple.
The 40-lux conditions also carry almost every negative clip: 28\% of their SaCo values fall below zero, against 0.6\% at 200~lux and 5.6\% at 700~lux.

The four methods agree with one another more closely in what they attribute to than in how faithful their orderings are, and not in the same pairs (Table~\ref{tab:method_agreement}).
Raw attention and attention rollout produce almost the same SaCo, at $\rho=0.97$, while their refined skin coverage agrees at only $0.18$;
attention rollout and attention flow are the other way round, agreeing at $0.96$ in coverage and $0.56$ in SaCo.
Beyond Intuition is the least like the others in SaCo, at $0.36$ to $0.52$, and closer to them in coverage, at $0.48$ to $0.65$.
Two methods can therefore place their attribution almost identically and still be ordered differently enough to separate their faithfulness, and the reverse.

\begin{table}[htbp]
\centering
\caption[Agreement between the four attribution methods]{Spearman correlations between the four attribution methods, taken over every clip of every participant of all nine conditions.
The lower triangle gives the refined skin coverage, over 4142 clips, and the upper triangle SaCo, over the 1741 clips the SaCo run evaluated.
Coverage is the mean over TPT$_1$--TPT$_3$ for the attention-only methods and the single refined value for Beyond Intuition.
Each cell gives $\rho; p$, and cells in bold satisfy both $p<0.05$ and $|\rho|\ge0.10$, the condition counted as directional;
every cell meets it at these sample sizes, and a $p$ that underflows double precision is written at the bound $10^{-300}$.
Both triangles pool the conditions and the participants, so differences between them contribute here, unlike the coefficients of Table~\ref{tab:correlation_levels}.}
\label{tab:method_agreement}
\small
\setlength{\tabcolsep}{4pt}
\begin{tabular}{lcccc}
\toprule
 & Raw Attention & Attention Rollout & Attention Flow & Beyond Intuition \\
\midrule
Raw Attention & --- & $\mathbf{0.97; {<}10^{-300}}$ & $\mathbf{0.55; {<}10^{-138}}$ & $\mathbf{0.36; {<}10^{-52}}$ \\
Attention Rollout & $\mathbf{0.18; {<}10^{-32}}$ & --- & $\mathbf{0.56; {<}10^{-145}}$ & $\mathbf{0.36; {<}10^{-53}}$ \\
Attention Flow & $\mathbf{0.22; {<}10^{-45}}$ & $\mathbf{0.96; {<}10^{-300}}$ & --- & $\mathbf{0.52; {<}10^{-122}}$ \\
Beyond Intuition & $\mathbf{0.48; {<}10^{-234}}$ & $\mathbf{0.57; {<}10^{-300}}$ & $\mathbf{0.65; {<}10^{-300}}$ & --- \\
\bottomrule

\end{tabular}
\end{table}

\begin{table}[htbp]
\centering
\caption[Skin coverage and SaCo across the nine conditions]{Median skin coverage and SaCo faithfulness ($F$, $K{=}8$) for the eight NCKU-rPPG scenarios and UBFC-rPPG, each given as median with the interquartile range in brackets.
For attention-based skin coverage, each clip was first averaged across TPT$_1$--TPT$_3$; Beyond Intuition operates only at the refined level.}
\label{tab:xai_median_iqr}
\scriptsize
\setlength{\tabcolsep}{3pt}
\begin{tabular}{llccc}
\toprule
Condition & Method & Skin Cov.\ (pre)$\uparrow$ & Skin Cov.\ (refined)$\uparrow$ & SaCo $F$$\uparrow$ \\
\midrule
\multirow{4}{*}{Static level~1} & Raw Attention & 0.369 [0.276, 0.403] & 0.345 [0.316, 0.378] & 0.739 [0.445, 0.833] \\
 & Attention Rollout & 0.387 [0.300, 0.451] & 0.376 [0.279, 0.423] & 0.732 [0.443, 0.827] \\
 & Attention Flow & 0.365 [0.280, 0.428] & 0.337 [0.286, 0.376] & 0.202 [-0.012, 0.409] \\
 & Beyond Intuition & --- & 0.180 [0.114, 0.428] & -0.178 [-0.523, 0.420] \\
\midrule
\multirow{4}{*}{Static level~3} & Raw Attention & 0.433 [0.311, 0.495] & 0.457 [0.308, 0.504] & 0.701 [0.563, 0.843] \\
 & Attention Rollout & 0.491 [0.338, 0.566] & 0.492 [0.333, 0.557] & 0.703 [0.563, 0.839] \\
 & Attention Flow & 0.500 [0.331, 0.563] & 0.510 [0.362, 0.545] & 0.641 [0.494, 0.722] \\
 & Beyond Intuition & --- & 0.789 [0.627, 0.875] & 0.837 [0.740, 0.974] \\
\midrule
\multirow{4}{*}{Static level~5} & Raw Attention & 0.365 [0.289, 0.420] & 0.511 [0.436, 0.554] & 0.548 [0.355, 0.727] \\
 & Attention Rollout & 0.492 [0.425, 0.568] & 0.458 [0.383, 0.548] & 0.543 [0.369, 0.713] \\
 & Attention Flow & 0.475 [0.419, 0.551] & 0.435 [0.374, 0.480] & 0.416 [0.227, 0.620] \\
 & Beyond Intuition & --- & 0.779 [0.544, 0.897] & 0.708 [0.457, 0.755] \\
\midrule
\multirow{4}{*}{Speak} & Raw Attention & 0.394 [0.353, 0.440] & 0.507 [0.405, 0.568] & 0.841 [0.730, 0.905] \\
 & Attention Rollout & 0.630 [0.457, 0.681] & 0.575 [0.455, 0.663] & 0.847 [0.718, 0.917] \\
 & Attention Flow & 0.639 [0.480, 0.687] & 0.538 [0.431, 0.601] & 0.824 [0.735, 0.899] \\
 & Beyond Intuition & --- & 0.632 [0.495, 0.731] & 0.863 [0.714, 0.974] \\
\midrule
\multirow{4}{*}{Rotate} & Raw Attention & 0.410 [0.342, 0.461] & 0.480 [0.434, 0.523] & 0.819 [0.675, 0.901] \\
 & Attention Rollout & 0.478 [0.396, 0.563] & 0.392 [0.340, 0.452] & 0.816 [0.677, 0.902] \\
 & Attention Flow & 0.419 [0.346, 0.484] & 0.353 [0.312, 0.395] & 0.652 [0.466, 0.793] \\
 & Beyond Intuition & --- & 0.620 [0.491, 0.736] & 0.751 [0.581, 0.922] \\
\midrule
\multirow{4}{*}{Bike level~1} & Raw Attention & 0.397 [0.336, 0.435] & 0.505 [0.424, 0.544] & 0.537 [0.003, 0.816] \\
 & Attention Rollout & 0.446 [0.357, 0.527] & 0.321 [0.255, 0.385] & 0.521 [0.032, 0.817] \\
 & Attention Flow & 0.325 [0.290, 0.444] & 0.298 [0.226, 0.360] & 0.345 [-0.083, 0.697] \\
 & Beyond Intuition & --- & 0.369 [0.273, 0.487] & 0.201 [-0.289, 0.546] \\
\midrule
\multirow{4}{*}{Bike level~3} & Raw Attention & 0.383 [0.326, 0.421] & 0.418 [0.346, 0.472] & 0.830 [0.659, 0.900] \\
 & Attention Rollout & 0.629 [0.569, 0.696] & 0.612 [0.560, 0.703] & 0.821 [0.687, 0.898] \\
 & Attention Flow & 0.620 [0.548, 0.670] & 0.572 [0.494, 0.639] & 0.665 [0.540, 0.822] \\
 & Beyond Intuition & --- & 0.695 [0.532, 0.779] & 0.822 [0.671, 0.976] \\
\midrule
\multirow{4}{*}{Bike level~5} & Raw Attention & 0.432 [0.335, 0.478] & 0.554 [0.454, 0.588] & 0.767 [0.555, 0.845] \\
 & Attention Rollout & 0.493 [0.379, 0.545] & 0.344 [0.248, 0.427] & 0.756 [0.527, 0.832] \\
 & Attention Flow & 0.447 [0.332, 0.499] & 0.366 [0.292, 0.411] & 0.680 [0.431, 0.821] \\
 & Beyond Intuition & --- & 0.696 [0.461, 0.815] & 0.724 [0.462, 0.883] \\
\midrule
\multirow{4}{*}{UBFC-rPPG}
 & Raw Attention & 0.348 [0.261, 0.404] & 0.587 [0.521, 0.624] & 0.845 [0.735, 0.936] \\
 & Attention Rollout & 0.671 [0.556, 0.725] & 0.567 [0.461, 0.600] & 0.850 [0.740, 0.932] \\
 & Attention Flow & 0.638 [0.506, 0.693] & 0.457 [0.388, 0.495] & 0.845 [0.763, 0.913] \\
 & Beyond Intuition & --- & 0.826 [0.661, 0.852] & 0.917 [0.843, 0.979] \\
\bottomrule

\end{tabular}
\end{table}

Table~\ref{tab:xai_direction_counts} summarises the direction of all 252 correlations across the nine conditions.
A correlation counts as directional only if its $p$ is below 0.05 and $|\rho|$ is at least 0.10, the second limit being Cohen's benchmark for a small effect.
Twenty-five of the 252 are directional, 16 positive and 9 negative, and 28 reach $p<0.05$ against the 13 expected by chance.

\begin{table}[htbp]
\centering
\caption[Direction of the clip-level correlations]{Direction of the clip-level Spearman correlations across the eight NCKU-rPPG scenarios and UBFC-rPPG.
Each row counts the nine conditions, so the three columns sum to nine.
A correlation is counted as negative or positive only when its $p$ is below 0.05 and $|\rho|$ is at least 0.10; everything else is counted as no relationship.
Each coefficient is the participant-averaged Spearman correlation of Section~\ref{sec:xai_comparisons}, so its $p$ value is a one-sample $t$ test over participants.}
\label{tab:xai_direction_counts}
\small
\setlength{\tabcolsep}{4pt}
\begin{tabular}{llccc}
\toprule
Relationship & Method & Negative & None & Positive \\
\midrule
HR MAE vs. skin coverage & Raw Attention & 1 & 8 & 0 \\
 & Attention Rollout & 1 & 7 & 1 \\
 & Attention Flow & 0 & 9 & 0 \\
 & Beyond Intuition & 0 & 6 & 3 \\
\midrule
HR MAE vs. SaCo & Raw Attention & 1 & 8 & 0 \\
 & Attention Rollout & 0 & 9 & 0 \\
 & Attention Flow & 0 & 8 & 1 \\
 & Beyond Intuition & 0 & 9 & 0 \\
\midrule
Waveform Pearson vs. skin coverage & Raw Attention & 0 & 8 & 1 \\
 & Attention Rollout & 0 & 7 & 2 \\
 & Attention Flow & 0 & 7 & 2 \\
 & Beyond Intuition & 1 & 8 & 0 \\
\midrule
Waveform Pearson vs. SaCo & Raw Attention & 0 & 9 & 0 \\
 & Attention Rollout & 0 & 9 & 0 \\
 & Attention Flow & 0 & 8 & 1 \\
 & Beyond Intuition & 2 & 7 & 0 \\
\midrule
SNR vs. skin coverage & Raw Attention & 0 & 9 & 0 \\
 & Attention Rollout & 0 & 9 & 0 \\
 & Attention Flow & 0 & 9 & 0 \\
 & Beyond Intuition & 2 & 7 & 0 \\
\midrule
SNR vs. SaCo & Raw Attention & 0 & 9 & 0 \\
 & Attention Rollout & 0 & 9 & 0 \\
 & Attention Flow & 0 & 9 & 0 \\
 & Beyond Intuition & 1 & 6 & 2 \\
\midrule
SaCo vs. skin coverage & Raw Attention & 0 & 9 & 0 \\
 & Attention Rollout & 0 & 9 & 0 \\
 & Attention Flow & 0 & 9 & 0 \\
 & Beyond Intuition & 0 & 6 & 3 \\
\bottomrule

\end{tabular}
\end{table}

No relationship is directional in more than six of its 36 method--condition pairs, and four of the seven are directional in fewer than four.
MAE--skin coverage has 4 positive and 2 negative pairs, the largest being Beyond Intuition on Bike level~1 and on Speak, both at $\rho=+0.24$ ($p<10^{-3}$), so where it resolves at all more attributed skin accompanies larger error.
Waveform Pearson--skin coverage has 5 positive and 1 negative, and its three largest are the two UBFC-rPPG coefficients quoted above together with attention flow on Speak at $\rho=+0.19$ ($p=.050$).
The two SNR relationships behave like the rest and add nothing that the others did not already show: SNR--skin coverage resolves twice in 36, both negative and both Beyond Intuition, on Speak at $\rho=-0.25$ ($p=.001$) and Bike level~1 at $\rho=-0.12$ ($p=.011$), and SNR--SaCo three times, at $|\rho|\le0.19$.
MAE--SaCo resolves twice in 36, Waveform Pearson--SaCo three times, and SaCo--skin coverage three times, the last being Beyond Intuition on Static level~5, Rotate and UBFC-rPPG, all positive.
Beyond Intuition accounts for 14 of the 25 directional pairs, more than the three attention-only methods together, and every one of the five SNR pairs.

Every coefficient reported above is taken inside one condition for one attribution method, over the clips of that condition alone, so the 36 coefficients of a relationship never mix two conditions and the differences between conditions cannot enter any of them.
Pooling those clips without regard to who they came from yields 146 nominally significant coefficients of the 252 and reaches $|\rho|=0.71$;
taking the coefficient inside each participant first yields 28 and reaches $0.32$.
The pooled coefficients are therefore dominated by differences between the participants of a condition rather than by anything an attribution map reveals about a clip, and the largest of them reverse sign under the correction: the waveform Pearson--skin coverage of Static level~3 reads $-0.41$ pooled and between $-0.16$ and $+0.13$ within participants.

Table~\ref{tab:correlation_levels} splits that pooled coefficient into the two levels it mixes.
Its three blocks summarise the same 36 coefficients of each relationship, one per condition and attribution method, in three ways: pooled over every clip of the condition, taken inside each participant of the condition and averaged through the Fisher transform, and taken between the participants of the condition by correlating their means.
The between-participant block carries its own denominator because the SaCo run reaches three of the twelve participants in each cycling condition, where a coefficient over three points carries no information and is left out.
Between the participants of one condition the medians are the largest of the three levels in six of the seven relationships, and they carry the pooled sign for every relationship pairing skin coverage with a performance measure: $+0.20$ against a pooled $+0.11$ for MAE--skin coverage, $-0.24$ against $-0.11$ for the waveform correlation, and $-0.28$ against $-0.14$ for SNR.
Within participants the same three medians sit within $0.03$ of zero, so what the pooled coefficient measures is the participant and not the clip.
The largest coefficients behave the same way: between participants they reach $0.66$ to $0.77$ in every relationship, against $0.19$ to $0.32$ within them, and on twelve points rather than the two hundred and more a pooled coefficient uses.
The level above the condition, where illumination and movement differ, is taken up in Section~\ref{sec:scenario_reliability_results}.

\begin{table}[htbp]
\centering
\caption[Correlations at each level of the nesting]{Each relationship summarised at three levels--pooled over every clip of the condition, taken inside each participant of the condition and averaged through the Fisher transform, and taken between the participants of the condition by correlating their means--over the same 36 condition-and-method coefficients.
\emph{Sig.}\ counts those reaching $p<0.05$, \emph{Median} is their median and \emph{Largest} the largest of them by magnitude.}
\label{tab:correlation_levels}
\small
\setlength{\tabcolsep}{4pt}
\begin{tabular}{lccccccccc}
\toprule
 & \multicolumn{3}{c}{Pooled over clips} & \multicolumn{3}{c}{Within participants} & \multicolumn{3}{c}{Between participants} \\
\cmidrule(lr){2-4}\cmidrule(lr){5-7}\cmidrule(lr){8-10}
Relationship & Sig. & Median & Largest & Sig. & Median & Largest & Sig. & Median & Largest \\
\midrule
MAE--skin & 24 & $+0.11$ & $0.35$ & 6 & $-0.01$ & $0.24$ & 2/36 & $+0.20$ & $0.66$ \\
MAE--SaCo & 16 & $+0.01$ & $0.53$ & 3 & $-0.01$ & $0.19$ & 1/24 & $+0.13$ & $0.67$ \\
Pearson--skin & 27 & $-0.11$ & $0.51$ & 7 & $+0.02$ & $0.32$ & 7/36 & $-0.24$ & $0.69$ \\
Pearson--SaCo & 14 & $+0.06$ & $0.55$ & 3 & $+0.01$ & $0.22$ & 2/24 & $+0.08$ & $0.75$ \\
SNR--skin & 25 & $-0.14$ & $0.38$ & 3 & $+0.01$ & $0.25$ & 4/36 & $-0.28$ & $0.66$ \\
SNR--SaCo & 17 & $+0.08$ & $0.43$ & 3 & $+0.02$ & $0.19$ & 3/24 & $-0.07$ & $0.77$ \\
SaCo--skin & 23 & $+0.02$ & $0.71$ & 3 & $+0.03$ & $0.29$ & 1/24 & $-0.06$ & $0.77$ \\
\bottomrule

\end{tabular}
\end{table}

\subsection{Scenario-level SaCo and model reliability}
\label{sec:scenario_reliability_results}

Table~\ref{tab:scenario_saco_reliability} takes every relationship to the level above the condition, which is the level at which illumination and movement differ.
Each coefficient correlates the scenario medians of the relationship's two measures over the eight NCKU-rPPG scenarios, one coefficient per attribution method.
UBFC-rPPG is left out because it is a different dataset evaluated with a separately trained checkpoint rather than a ninth point on the same axis.
The medians are those of Table~\ref{tab:xai_median_iqr} for the attribution measures and the medians of the same clips for MAE, the waveform Pearson correlation $r_{\mathrm{wave}}$ of Equation~\eqref{eq:pearson_wave}, and SNR, so they are medians throughout rather than the means of Table~\ref{tab:performance_results}.
The performance side of a relationship does not depend on which attribution method produced the map, so within a row the four coefficients differ only through the attribution measure.
Eight points resolve nothing at this level: $p$ falls below $0.05$ only at $|\rho|=0.74$ and below $0.10$ only at $0.64$, and the largest coefficient of the table is $+0.57$.
Bold therefore marks $|\rho|$ above $0.40$, which says that a pair moves together across the scenarios strongly enough to be worth reading and never that a relationship is established.

\begin{table}[htbp]
\centering
\caption[Scenario-level correlations between the measures]{Spearman correlations between the scenario medians of each pair of measures, over the eight NCKU-rPPG scenarios.
None is significant at $p<0.05$ due to the small number of scenarios.
Each cell gives $\rho; p$ for one attribution method, and a coefficient in bold has $|\rho|$ above $0.40$ before rounding.}
\label{tab:scenario_saco_reliability}
\small
\setlength{\tabcolsep}{5pt}
\begin{tabular}{lcccc}
\toprule
Relationship & Raw Attention & Attention Rollout & Attention Flow & Beyond Intuition \\
\midrule
MAE--skin & $-0.21; .610$ & $+0.02; .955$ & $-0.02; .955$ & $\mathbf{-0.43}; .289$ \\
MAE--SaCo & $+0.31; .456$ & $+0.31; .456$ & $+0.33; .420$ & $-0.10; .823$ \\
Pearson--skin & $+0.05; .911$ & $+0.12; .779$ & $+0.14; .736$ & $\mathbf{+0.57}; .139$ \\
Pearson--SaCo & $-0.24; .570$ & $-0.24; .570$ & $-0.17; .693$ & $+0.05; .911$ \\
SNR--skin & $+0.21; .610$ & $-0.02; .955$ & $+0.02; .955$ & $\mathbf{+0.43}; .289$ \\
SNR--SaCo & $\mathbf{-0.50}; .207$ & $\mathbf{-0.50}; .207$ & $\mathbf{-0.40}; .320$ & $-0.05; .911$ \\
SaCo--skin & $-0.33; .420$ & $\mathbf{+0.50}; .207$ & $\mathbf{+0.48}; .233$ & $\mathbf{+0.50}; .207$ \\
\bottomrule

\end{tabular}
\end{table}

The reliability hypothesis predicts a negative coefficient with MAE and positive ones with the waveform correlation and SNR.
Only the skin coverage of Beyond Intuition points the hypothesised way on all three performance measures, at $-0.43$ with MAE, $+0.57$ with the waveform correlation and $+0.43$ with SNR, while the three attention-only methods stay within $|\rho|=0.21$ of zero on those same relationships.
The SaCo relationships point the other way: for the three attention-only methods SaCo rises with MAE, at $+0.31$ to $+0.33$, and falls with the waveform correlation, $-0.17$ to $-0.24$, and with SNR, $-0.40$ to $-0.50$, each opposite to the direction a reliability indicator would take, while the three coefficients of Beyond Intuition sit within $0.10$ of zero.
SaCo and skin coverage move together for every method but raw attention, at $+0.48$ to $+0.50$ against $-0.33$.
The eight-scenario comparison therefore did not support median SaCo as a consistent tracker of model reliability.

\section{Discussion}
\label{sec:discussion}

\subsection{Principal findings}

Within a single participant of a single condition, neither the Salience-guided Faithfulness Coefficient nor the skin coverage of any of the four attribution methods is related to the heart-rate error, the waveform correlation, or the signal-to-noise ratio of a clip (Table~\ref{tab:correlation_levels}).
The explainable artificial intelligence measures therefore carry information that is complementary to the three performance measures rather than a proxy for them, and attributing to the skin does not by itself guarantee that the heart rate or the waveform is well estimated.

The evidence is the near-absence of resolved relationships rather than a small effect.
Of the 252 clip-level coefficients, 186 have $|\rho|$ below $0.10$, Cohen's benchmark for a small effect, and only 28 reach $p<0.05$ at all, against the 13 that a nominal 0.05 level would produce by chance in a family of that size (Supplementary Table~\ref{tab:supp_xai_spearman_full}).
Both positive and negative relationships occur for the same method pair across the four attentions (Table~\ref{tab:xai_direction_counts}).
The largest of the 252 reaches $\rho=0.32$, no condition resolves more than five, and no relationship resolves in more than six of its 36 method-and-condition pairs (Table~\ref{tab:xai_direction_counts}).
The same clips pooled without regard to who they came from yield 146 coefficients at $p<0.05$ and reach $0.71$, so what a pooled analysis of this kind measures is the participants of a condition and not the clips (Table~\ref{tab:correlation_levels}).

Beyond Intuition is the most useful of the four attribution methods on this evidence.
It supplies 14 of the 25 relationships that are directional, meaning $p<0.05$ with $|\rho|\ge0.10$ together, which is more than the three attention-only methods supply between them (Table~\ref{tab:xai_direction_counts});
it holds the highest median refined skin coverage in seven of the nine conditions and the highest median SaCo in four (Table~\ref{tab:xai_median_iqr});
and across the eight scenarios it is the only method whose skin coverage follows the reliability measures, falling with the heart-rate error at $\rho=-0.43$ and rising with the waveform correlation and SNR at $+0.57$ and $+0.43$ (Table~\ref{tab:scenario_saco_reliability}).
The three attention-only methods invite caution for the opposite reason: at the same scenario level their SaCo rises with the heart-rate error, at $+0.31$ to $+0.33$, and falls with the waveform correlation and SNR, at $-0.17$ to $-0.24$ and $-0.40$ to $-0.50$, each the reverse of what a reliability indicator would do (Table~\ref{tab:scenario_saco_reliability}).
They also cannot be treated as interchangeable readings of the same thing, nor as four independent ones.
Raw attention and attention rollout order the clips almost identically by SaCo, at $\rho=0.97$, while agreeing on where they attribute at only $0.18$, and attention rollout and attention flow do the reverse, at $0.96$ in coverage and $0.56$ in SaCo (Table~\ref{tab:method_agreement}).
Two methods can therefore attribute to almost the same place and still be ordered differently enough for their faithfulness to separate, and two that are ordered alike can attribute to different places, so a choice between them is a choice of both properties and not of one.

The one condition in which Beyond Intuition fails is the dimmest.
At 40~lux its median refined skin coverage falls to $0.180$ on Static level~1 against $0.789$ and $0.779$ at 200 and 700~lux, and its median SaCo falls to $-0.178$, with 117 of its 200 clips below zero (Table~\ref{tab:xai_median_iqr}).
Coverage and faithfulness dropping together points at where the attribution lands rather than at how it is ordered, so in dim light the model does not concentrate on the skin.
Motion produces no comparable drop even though it degrades the estimates far more: speaking, rotation, and cycling at 200 and 700~lux raise the heart-rate error from $1.96\pm0.34$ to between $8.9\pm2.6$ and $16.1\pm2.3$~beats per minute while the median refined coverage of Beyond Intuition stays between $0.620$ and $0.696$ and its median SaCo between $0.724$ and $0.863$ (Tables~\ref{tab:performance_results} and~\ref{tab:xai_median_iqr}).
What an attribution reveals about a condition is therefore where the model looks rather than how faithfully its map is ordered or the quality of the heart rate estimates.
Just because a model looks in the right place does not mean the estimate will be accurate.
For some clips even a little attention on the skin can be enough to produce a good heart-rate estimate, and for others even a lot of attention on the skin can be insufficient.

\subsection{Model performance across NCKU-rPPG}

The static results were non-monotonic: level~3 performed best, level~5 ranked second, and level~1 was weaker across the reported performance measures (Table~\ref{tab:performance_results}).
This ordering does not show that performance improved or deteriorated monotonically with illumination.
The three rows came from independently trained models and scenario-specific samples, so participant availability, recording quality, and training variation could also contribute to their differences.
A causal illumination analysis would require paired observations that hold the model and participants constant while changing illumination, together with repeated training seeds.

The cycling conditions were ordered by illumination where the static ones were not.
Their heart-rate error fell from $18.6\pm5.5$ to $16.1\pm2.3$ to $15.1\pm3.5$~beats per minute from 40 to 200 to 700~lux, and their root mean square error, mean absolute percentage error, and heart-rate correlation ordered the three the same way;
the waveform correlation and SNR did not, placing Bike level~3 below Bike level~1 on both (Table~\ref{tab:performance_results}).
The ordering is not a separation, however: the three cycling conditions were not separated from one another at the clip level, every pairwise difference being within twice the standard errors of the values it separated.
A monotonic trend in four of the six measures across three conditions that no measure resolves is a direction worth testing with paired recordings, not a demonstration that illumination helps under motion.

Speaking and head rotation at illumination level~3 had markedly lower correlations and SNRs than Static level~3 (Table~\ref{tab:performance_results}).
Speaking introduced local deformation around the cheeks and mouth, whereas rotation changed head pose and the visible facial surface.
The face-region conversion stabilised a square crop before training, but it could not remove appearance changes or motion-related colour variation within that crop.
All three cycling conditions likewise had lower correlations and negative SNRs relative to static levels~3 and~5, indicating that motion robustness remained limited.

Static level~3 makes the more durable point, pairing a heart-rate correlation of $0.968\pm0.079$ with a waveform correlation of $0.626\pm0.037$ (Table~\ref{tab:performance_results}).
This shows why dominant-frequency heart-rate error and spectral waveform quality should be interpreted together: a frequency peak can remain identifiable even when much of the predicted waveform power lies outside the reference fundamental and harmonic bands.

These comparisons describe eight independently trained models, each produced with one random seed and evaluated at the final epoch, the thirtieth (Supplementary Table~\ref{tab:supp_training_settings}).
No validation loader or validation-based checkpoint selection was used.
The design therefore supports a descriptive baseline but not statistical superiority among scenarios, attribution of between-row differences to illumination or motion alone, or generalisation to another dataset.

These results further indicate that RhythmFormer is not robust to variations in illumination.
Underexposed conditions substantially degrade its performance, regardless of whether the subject remains stationary or moves.

\subsection{Static level 3 versus UBFC-rPPG}

The Static level~3 and UBFC-rPPG evaluations used the same RhythmFormer architecture, similar observation windows, stationary subjects in controlled settings, and 12 held-out participants in each test set.
Static level~3 used 256 frames at 50~Hz, corresponding to 5.12~s, whereas the UBFC-rPPG reproduction used 160 frames at 30~Hz, corresponding to approximately 5.33~s~\citep{chen2026RhythmFormerPhotoplethysmographyEXPLIMED}.

Static level~3 yielded mean absolute error (MAE) $1.96\pm0.34$~beats per minute, root mean square error (RMSE) $3.74\pm0.75$, mean absolute percentage error (MAPE) $2.36\pm0.42$, heart-rate correlation $0.968\pm0.079$, waveform correlation $0.626\pm0.037$, and signal-to-noise ratio (SNR) $5.7\pm1.5$~dB (Table~\ref{tab:performance_results}).
The UBFC-rPPG reproduction yielded $0.96\pm0.16$, $1.37\pm0.28$, $0.99\pm0.18$, $0.991\pm0.041$, $0.828\pm0.022$, and $8.54\pm0.97$~dB on the same six measures~\citep{chen2026RhythmFormerPhotoplethysmographyEXPLIMED}.
UBFC-rPPG is therefore ahead on every column, but by very different margins.
The two heart-rate correlations are close, 0.991 against 0.968, because both models order the clips of their evaluation correctly.
The waveform correlations are closer than they first appeared, 0.828 against 0.626, but only once a per-recording synchronisation lag is fitted;
at zero lag Static level~3 reads 0.28, and the difference between those two figures is a property of the NCKU-rPPG recordings rather than of the model, since the same fit is worth nothing on UBFC-rPPG (Supplementary Table~\ref{tab:supp_waveform_lag}).
This indicates an error in the synchronisation of the NCKU-rPPG recordings, which is not a property of the model and does not affect the heart-rate estimates and was removed in the waveform correlation reported here by fitting a per-recording lag.
What still separates the two evaluations is the heart-rate error, half as large on UBFC-rPPG, and a higher signal-to-noise ratio.
The reported difference of 2.8~decibels is a lower bound rather than a like-for-like figure, because the two SNR periodograms are zero-padded differently, and matching their resolution widens it to about 4~decibels.
They remain separate within-dataset evaluations of different checkpoints and do not test cross-dataset generalisation.

Several differences remain uncontrolled: frame rate and clip length, data splitting, participant composition, cameras, face preprocessing, recorded activities, and the trained checkpoint.
These differences prevent a causal explanation of the small metric differences and limit the comparison to a descriptive reference.

\subsection{Static level 3 XAI versus UBFC-rPPG}

Beyond Intuition achieved the highest median skin coverage in both evaluations, with 0.789 on Static level~3 and 0.826 on UBFC-rPPG (Table~\ref{tab:xai_median_iqr})~\citep{chen2026RhythmFormerPhotoplethysmographyEXPLIMED}.
It also achieved the highest median SaCo in both evaluations, with 0.837 on Static level~3 and 0.917 on UBFC-rPPG.
The secondary rankings nevertheless differed: attention rollout, raw attention, and attention flow yielded median SaCo values of 0.703, 0.701, and 0.641 on Static level~3, compared with 0.851, 0.846, and 0.846 on UBFC-rPPG (Table~\ref{tab:xai_median_iqr}).
The repeated top rank for Beyond Intuition is descriptive evidence from these two evaluations rather than proof that the full XAI ranking transfers across datasets.

The split violins also show substantial within-dataset heterogeneity that the medians conceal (Figure~\ref{fig:combined_xai_distributions}).
For Static level~3, the SaCo interquartile ranges were 0.228--0.280 across methods, and individual values extended from 0.017 to 1.000.
The corresponding UBFC-rPPG interquartile ranges were 0.135--0.201, with a Beyond Intuition outlier at $-0.534$ and upper values of 1.000.
\citet{chen2026RhythmFormerPhotoplethysmographyEXPLIMED} traced that outlier to the participant's hand entering the field of view in six of that clip's 160 frames, 3.75~\% of it.
Replacing those six frames with stable neighbouring frames from the same clip lifts the Beyond Intuition coefficient from $-0.53$ to $0.76$ and moves all four attribution methods towards $F\approx0.8$, so the clip rather than the method produced the value.
Refined skin coverage also varied widely among clips, particularly for Beyond Intuition and rollout.
Part of this spread is consistent with how SaCo itself is computed: partitioning each $4\times4$ salience map into $K=8$ rank-ordered groups of two patches each leaves only 28 pairwise comparisons, so a small change in the salience map that moves one patch across a group boundary can reverse several comparisons at once and push $F$ towards the extremes reported above.
The groups are also large, and they are formed by salience rank alone rather than by anatomy: $K=8$ divides the map into eight groups whatever its resolution, so each group covers an eighth of the $128\times128$ crop, as two $32\times32$ patches for the attention-only methods and eight $16\times16$ patches for the $8\times8$ map of Beyond Intuition.
A region that size cannot follow the boundary of the skin mask, so several groups hold skin and non-skin together in one perturbation, and a low coefficient need not mean that any part of the map is anatomically implausible.
The comparison also rests on one trained checkpoint per condition from a single random seed with no validation-based selection, as noted above; the resulting attribution behaviour reflects that run's particular idiosyncrasies as well as genuine clip-level variation, a source of noise that the more controlled Static level~3 recording protocol does not by itself reduce.
Consequently, a higher median does not imply uniformly better explanations, and pooled correlations may be influenced by scenario imbalance and extreme clips.

The relation between pre-attention and refined attention also differed.
On UBFC-rPPG, refinement produced a large increase in raw-attention skin coverage and decreases in rollout and attention-flow coverage~\citep{chen2026RhythmFormerPhotoplethysmographyEXPLIMED}.
On Static level~3, the corresponding pooled mean differences were only $+0.0179$, $+0.0004$, and $-0.0048$ (Table~\ref{tab:static3_xai_coverage}).
The pronounced UBFC-rPPG stage effect therefore did not recur at a comparable magnitude.
The Chen and Nordling UBFC-rPPG study directly quantified multi-hop leakage, whereas the Static level~3 analysis did not quantify leakage mass.
The small stage differences on Static level~3 consequently neither reproduce nor refute that mechanism.

The clip-level correlation structures provide a further contrast.
On UBFC-rPPG, the skin coverage of attention rollout and of attention flow was positively related to waveform Pearson correlation within participants, at Spearman $\rho=0.32$ ($p=.047$) and $0.21$ ($p=.041$) respectively (Table~\ref{tab:static3_ubfc_spearman}), and these are two of only twenty-five relationships resolved out of 252.
On Static level~3, the corresponding coefficients were close to zero at $\rho=-0.00$ and $-0.09$; the raw-attention and attention-flow coefficients were likewise small at $0.09$ and $0.01$.
Heart-rate MAE and skin coverage were positively related for all four Static level~3 methods ($\rho=0.19$--$0.26$), whereas the UBFC-rPPG coefficients were negative or close to zero ($\rho=-0.22$--$-0.06$), both in Table~\ref{tab:static3_ubfc_spearman}.
The MAE--SaCo coefficients remained close to zero on both datasets: $\rho=-0.06$--$0.06$ on Static level~3 and $0.00$--$0.04$ on UBFC-rPPG.
For SaCo--skin coverage, Beyond Intuition was resolved on UBFC-rPPG at $\rho=0.28$ ($p=.007$) and unresolved on Static level~3.
The wider lesson is not that the two evaluations disagree but that almost nothing survives in either: holding the participant fixed, neither skin coverage nor SaCo predicts how well the model does on a given clip.
That the coefficients vanish once the participant is held fixed says the measures tell different stories at the clip level: among the clips of one person in one condition, skin coverage, SaCo, heart-rate error, waveform correlation and the signal-to-noise ratio move independently of one another.
They stop being independent at the levels above the clip.
Between the participants of a condition the three relationships pairing skin coverage with a performance measure take a consistent sign (Table~\ref{tab:correlation_levels}), and across the eight scenarios the performance measures and the skin coverage of Beyond Intuition move together (Table~\ref{tab:scenario_saco_reliability}), which is what illumination and movement acting on all of them at once would produce.
SaCo is the exception at the scenario level, moving opposite to the reliability measures rather than with them, so the common driver reaches where an attribution lands but not how faithful its ordering is.

The correlation $p$ values are one-sample $t$ tests over participants, which are independent, so they are inferential rather than descriptive.
The clips are nested within subjects and therefore are not independent samples for confirmatory inference.

\subsection{Do skin coverage and SaCo track reliability across scenarios?}

The two measures answer this differently, and the difference is the clearest separation between them that this work produces.
Neither is resolved at the scenario level, where eight points put $p$ below 0.05 only at $|\rho|=0.74$, so what follows is a reading of directions and magnitudes rather than of established relationships (Table~\ref{tab:scenario_saco_reliability}).

Skin coverage tracks reliability for the method that attributes most tightly, and for that method only.
The coverage of Beyond Intuition falls as the heart-rate error rises, at $\rho=-0.43$, and rises with the waveform correlation and SNR, at $+0.57$ and $+0.43$: the three directions the reliability hypothesis predicts, and the three largest coverage coefficients in the table.
The three attention-only methods give nothing in either direction, staying within $|\rho|=0.21$ of zero on all three measures, so the property belongs to Beyond Intuition's maps rather than to skin coverage as a measure.

SaCo does the opposite of what a reliability indicator should do.
For raw attention, attention rollout, and attention flow, median SaCo rises with the heart-rate error, at $+0.31$ to $+0.33$, and falls with the waveform correlation and SNR, at $-0.17$ to $-0.24$ and $-0.40$ to $-0.50$, each the reverse of the predicted direction.
Beyond Intuition alone points the predicted way on the error and the waveform correlation, and even there its three coefficients sit within $0.10$ of zero, which is weaker than any of its coverage coefficients.
A method can therefore order its own perturbation responses consistently in a condition on which the model performs badly, which is what the attention-only rows say.

This does not invalidate SaCo as a within-clip perturbation-faithfulness measure.
SaCo asks whether the salience ordering agrees with the relative effects of perturbing spatial groups for a particular model output;
it is not defined as a surrogate for prediction accuracy or waveform quality across independently trained models, and the scenario-level analysis shows only that its median does not serve as one.
The contrast with coverage is what makes the negative result informative: where an attribution lands carries scenario-level information about how well the model does, at least for the method whose maps are concentrated enough for the location to mean something, and how faithfully that attribution is ordered does not.

The two measures nevertheless fail together in one place.
At 40~lux the median skin coverage and the median SaCo are both at their lowest, and 28~\% of the SaCo values there fall below zero against 0.6~\% at 200~lux.
Coverage and faithfulness falling together points at the attribution's location rather than at its ordering alone: in dim light the model does not concentrate on the skin, so the regions a map ranks highest are no longer the ones the prediction depends on.
The two also agree with each other more than either agrees with performance, at $+0.48$ to $+0.50$ across the scenarios for every method but raw attention.
With eight heterogeneous scenarios and one checkpoint each, all of this is a directional diagnostic and not significance, causal, or generalisation evidence.

\subsection{The evaluation window, and what the literature compares}
\label{sec:discussion_window}

The reported error depends on the evaluation window more strongly than on any other choice in Section~\ref{sec:performance_evaluation}, and lengthening the window always lowers it (Supplementary Table~\ref{tab:supp_window_length}).
Stability cannot decide the window on its own, because the steadiest window is the whole recording, which returns one number for a quantity that moved.
Judged instead against a target that needs no window, the beat-to-beat heart rate taken from the systolic-peak intervals of the reference photoplethysmogram, the error splits into a part that falls with length, the spectral estimate against the mean reference over the same window, and a part that grows with it, that window mean against the instantaneous reference at the window centre.
On Static level~3 the first falls from 2.62 to 0.84~beats per minute between 5.12 and 61.44~s while the second rises from 3.42 to 4.27, and the sum is least at 7.68~s (Supplementary Table~\ref{tab:supp_window_bias_variance}).
The curve is shallow, 3.34~beats per minute at that minimum against 3.60 at 5.12~s, so the clip we use costs 7~\% of a quantity that already carries a standard error of 0.34.
We keep 5.12~s because it is what makes the UBFC-rPPG row of Table~\ref{tab:performance_results} comparable at all, its clip being 5.33~s.
Under cycling the minimum moves to 8.96 and 10.24~s and sits three to five times higher, on curves that vary by no more than 17~\% of their own minimum across a twelvefold range of length: no window is good there, and the limitation is the reference rather than the window.

Whether a published error is comparable with ours depends on the same choice, and the field does not make it consistently.
Across the 58 articles published from 2020 onwards in our collection that apply a remote photoplethysmography method, the window is the least standardised of the four settings we recorded (Supplementary Section~\ref{sec:supp_hr_settings}).
Those that state one use between 5 and 60~s, one computes a single heart rate for a whole video, and the two commonest choices differ by a factor of three;
in our own data, moving between those two commonest choices changes the mean absolute error by 0.23 to 1.63~beats per minute across the eight conditions, which is the order of the differences that separate the methods being compared.
Only nine of the 58 state the transform length or the window function behind the spectrum they take a peak from, although at 30~frames per second an $N$-point transform resolves $1800/N$~beats per minute, so an unpadded 160-frame clip resolves no better than 11.
The frame rate is the one setting the field has settled, with 50 of the 58 at 30~Hz, which is why a window given in frames is shorter in seconds here, at 50~Hz, than in most of that literature and why we give both.
A heart-rate error is therefore not a number that can be compared across articles without the window, the padding, and the frame rate that produced it.

\subsection{Reproducibility, limitations, and future work}

The participant-independent split, the face-region conversion, the retention of the source video's 50~Hz frame cadence, and the quality gates applied during conversion together define the adaptation.
No participant and no session appeared in both the training and the test set of any scenario, and Supplementary Sections~\ref{sec:supp_labelchecks} and~\ref{sec:supp_training} state the inclusion thresholds and the training settings that were in force.

These results rest on two datasets, and each was chosen for what the other is not.
UBFC-rPPG is a public benchmark on which the heart rate is recognised as easy to recover: its subjects are stationary and evenly lit, and the median clip error here is 0.67~beats per minute against 1.14 on the most favourable of our own conditions.
NCKU-rPPG was recorded in our laboratory with the illumination set to 40, 200, and 700~lux, with speaking, head rotation, and cycling, and with an exercise bike that widens the reference heart rate from 54.5--127.2~beats per minute when the participant is stationary to 47.5--150.0 when cycling.
Two datasets, one easy and one deliberately varied, are still two.

The training protocol is a deliberate limitation rather than an oversight.
RhythmFormer was trained for 30 epochs, once, from a single random seed, as a separate model per condition so that each is specialised to its own data distribution and the nine differ in how well they perform;
the final-epoch checkpoint was evaluated with no validation loader and no checkpoint selection (Supplementary Table~\ref{tab:supp_training_settings}).
Nothing about the test data therefore reached the model at any point, so every figure in Table~\ref{tab:performance_results} is an estimate of generalisation error rather than a value selected against the set it is reported on.
A better model for each condition could almost certainly be obtained by training longer, selecting a checkpoint on a validation split, and searching over seeds.
We did not, because an observation made only on highly optimised models may hold only for highly optimised models.
The span was the point: conditions in which the error approaches the reference's own beat-to-beat noise, conditions in which it reaches 15 to 19~beats per minute, and models at both ends of that range.

The main limit on how far these conclusions generalise is that one model structure produced all of them.
Every statement here is about RhythmFormer, and whether the same attribution behaviour appears in another architecture is untested.
Training the nine-condition protocol for further architectures is beyond what we can afford at present, and we judge the observations worth reporting now rather than holding until it is.

The reference bounds what any of these numbers can mean.
One heart rate per 5.12~s clip, taken from the reference photoplethysmogram by the same spectral estimator, changes by more than 10~beats per minute between consecutive clips of one recording in 601 of 3905 such pairs, from 2.5~\% on Static level~5 to 23.8~\% on Bike level~5, so a share of the reported error is reference movement inside the window rather than model error.
Before any training or testing, 365 of the 25,848 complete raw clips were excluded, 80 of them from test sets, because the reference trace was non-finite or flat to within $10^{-6}$ of its own standard deviation and therefore carried no measurable pulse (Supplementary Table~\ref{tab:supp_clip_counts});
no model output entered that decision.
A further 30 of the 2559 cycling test clips are censored at the top of the 45--150~beats-per-minute search band.

The 5.12~s window is at the short end of the literature, so the errors reported here are larger than a 30~s or whole-recording evaluation of the same predictions would give: lengthening our own window to 61.44~s lowers the mean absolute error in every condition, by 1.4~beats per minute on Static level~3 and 6.6 on Bike level~1 (Supplementary Table~\ref{tab:supp_window_length}).
Section~\ref{sec:discussion_window} sets out what that costs against the beat-to-beat reference and why the window was kept nonetheless.

A condition's mean absolute error also conceals a spread of two orders of magnitude (Supplementary Figure~\ref{fig:supp_metric_distributions}).
On Static level~3 the middle half of clips lies between 0.51 and 2.18~beats per minute, the ninetieth percentile is 3.49, and the worst clip reaches 25.0;
on Bike level~1 the middle half spans 1.97 to 29.86, the ninetieth percentile is 53.90, and the worst reaches 98.8, with 42.9~\% of clips above 10~beats per minute against 2.8~\% on Static level~3.
Every condition is right-skewed and none is bimodal, so each mean sits above the bulk of its own clips and a comparison of means understates how often a clip fails outright.

The model comparison remains limited by one seed, independently trained condition-specific models, final-epoch checkpoints, no validation-based selection, and 12 test participants per scenario.
The XAI comparison adds differences in attribution resolution and layer aggregation: Beyond Intuition supplied a refined $8\times8$ map to SaCo, while the attention-only methods supplied $4\times4$ maps, and the methods handled temporal and cross-layer aggregation differently.
Skin coverage is also sensitive to the face parser and mask definition.
Finally, the UBFC-rPPG and Static level~3 summaries used different clip counts and different checkpoints, so neither method rankings nor correlation magnitudes should be treated as controlled dataset effects.

Future work should repeat training across seeds, compare joint-condition and condition-specific models, and recover matched participant identities for paired condition analyses.
It should also carry the same protocol to model structures beyond RhythmFormer, which is the single change that would most extend how far these observations reach.
XAI evaluation should compare SaCo at a common spatial resolution, directly quantify leakage on Static level~3, and test the stability of skin masks and attribution rankings across preprocessing choices.
Training and evaluation across NCKU-rPPG and UBFC-rPPG under matched preprocessing would then permit a direct test of cross-dataset transfer.
The multi-hop path mechanism should be evaluated in additional experiments so that its consistency with the mechanism observed on UBFC-rPPG can be tested directly.
The skin-mask and face-tracking pipeline should be strengthened by adopting FaceXFormer for face localisation and region estimation.
Analyses of additional scenarios should establish whether apparently unreliable XAI results arise from differences in model training quality rather than from a lack of interpretive value in the XAI methods themselves.
Finally, integrating results across multiple scenarios should help identify clear relationships among model-performance, waveform-quality, and XAI metrics.

\section{Conclusion}
\label{sec:conclusion}

Held within one participant of one condition, neither skin coverage nor SaCo is related to the heart-rate error, the waveform correlation, or the signal-to-noise ratio of a clip: 186 of the 252 coefficients fall below $|\rho|=0.10$ and 28 reach $p<0.05$ against the 13 chance would give.
The attribution measures therefore carry information complementary to the performance measures rather than a proxy for them, and attributing to the skin does not by itself guarantee a good estimate.
Pooling the same clips yields 146 coefficients at $p<0.05$ and reaches $0.71$, which measures the participants of a condition and not the clips, so a clip-pooled analysis of this kind answers a different question from the one it appears to answer.

Of the four attribution methods, Beyond Intuition is the most useful here: it supplies 14 of the 25 directional relationships, holds the highest median refined skin coverage in seven of the nine conditions, and is the only method whose coverage follows the heart-rate error, the waveform correlation, and the signal-to-noise ratio across the eight scenarios.
The three attention-only methods warrant caution, their SaCo running opposite to each of those three measures at that level, and the four are neither interchangeable nor independent: two of them can order the clips almost identically while agreeing on where they attribute at $\rho=0.18$.
Beyond Intuition fails in one condition only, the dimmest, where its median coverage falls to $0.180$ and its median SaCo to $-0.178$;
motion degrades the estimates far more and produces no such drop.
What an attribution reveals about a condition is therefore where the model looks, not how faithfully its map is ordered, and not how well the model performs.

On NCKU-rPPG the static ordering is non-monotonic in illumination while the cycling one is monotonic on four of the six measures, and no measure separates the three cycling conditions at the clip level.
The coexistence on Static level~3 of a heart-rate correlation of $0.968\pm0.079$ with a waveform correlation of $0.626\pm0.037$ shows that neither illumination labels nor heart-rate error alone characterise robustness, and every condition's mean absolute error hides a clip-level spread of two orders of magnitude.
Measured on clips of comparable length, the UBFC-rPPG reproduction predicts about twice as accurately as Static level~3, and the magnitudes and secondary rankings of the attribution measures differ between the two.

These findings are descriptive.
Each condition used one independently trained final-epoch checkpoint from a single seed, which is what makes the reported errors estimates of generalisation error but also leaves each model unoptimised;
one architecture produced every result;
the 5.12~s window sits at the short end of a literature whose windows range from 5 to 60~s, so no heart-rate error here is comparable with a published one without the window, padding, and frame rate that produced it.
Repeated seeds, further architectures, common-resolution attribution evaluation, and controlled cross-dataset experiments are needed before any causal claim about illumination, generalisation, or clinical validity.

\section*{Declarations}

\paragraph{Ethical approval.}
The National Cheng Kung University Human Research Ethics Committee approved the experimental procedures, as stated in Section~\ref{sec:ethics}, and every participant gave written informed consent.

\paragraph{Generative artificial intelligence.}
During the preparation of this work, the authors used Claude (Anthropic) for grammar and spelling checking, code generation for the analysis pipeline and figure-plotting scripts, and \LaTeX{} formatting assistance.
After using these tools and services, the authors reviewed and edited the content as needed and take full responsibility for the content of this work.

\paragraph{Data availability.}
UBFC-rPPG is a public dataset and is described by its authors~\citep{bobbia2019UnsupervisedPPGPatternRecognitLett}.
NCKU-rPPG is not publicly available at the time of writing.
The recordings of the participants who consented to their release are planned for publication in a later article, and until then the data can be made available by the corresponding author on reasonable request and subject to the approvals in Section~\ref{sec:ethics}.

\paragraph{Code availability.}
The conversion, training, evaluation, and attribution code is not publicly released.
It can be made available by the corresponding author on reasonable request.

\paragraph{Funding.}
This work was supported by the Taiwan National Science and Technology Council (NSTC) under Grant 111-2221-E-006-186 and Grant 114-2221-E-006-089.

\paragraph{Competing interests.}
The authors declare that they have no known competing financial interests or personal relationships that could have appeared to influence the work reported in this paper.

\paragraph{Author contributions.}
\textbf{Louis Chen:} Software, Data curation, Investigation, Formal analysis, Methodology, Validation, Visualization, Writing -- original draft, Writing -- review \& editing.
\textbf{Torbj{\"o}rn E.\,M.\ Nordling:} Conceptualization, Methodology, Formal analysis, Validation, Visualization, Writing -- original draft, Writing -- review \& editing, Supervision, Project administration, Resources, Funding acquisition.

\bibliographystyle{unsrtnat}
\bibliography{references}

\clearpage
\section*{Supplementary Material}
\label{sec:supplementary}

\makeatletter
\providecommand{\phantomsection}{}
\providecommand{\suppsection}[3]{%
  \phantomsection
  \subsection*{#2. #3}%
  \def\@currentlabel{#2}%
  \label{#1}%
}
\makeatother

\suppsection{sec:supp_contract}{S1}{Conversion contract and quality control}

The production conversion contract fixed the converter source revision, the preprocessing implementation, the FaceXFormer version (commit), model weight checkpoint digest, output codec, quality thresholds, expected session/subject identifiers, and expected skips before we ran the conversion.
Table~\ref{tab:supp_conversion_contract} summarises the contract.

\begin{table}[htbp]
\centering
\caption{Production conversion contract.}
\label{tab:supp_conversion_contract}
\small
\begin{tabular}{>{\raggedright\arraybackslash}p{6.2cm} >{\raggedright\arraybackslash}p{7.7cm}}
\toprule
Contract item & Recorded requirement or result \\
\midrule
Source sessions & 618 \\
Successful or resumed outputs & 607 \\
Skipped sessions & 11 \\
Conversion failures & 0 \\
Output video & $128\times128$ pixels; source frame rate; FFV1 \\
Per-session detection rate & At least 0.80 before interpolation \\
Per-session temporal RGB correlation & Prespecified minimum 0.99 after encode--decode; measured accepted-session values not retained \\
Aggregate detection rate & At least 0.90 \\
Aggregate interpolation rate & At most 0.10 \\
Session-level count checks & Reported video frames = decoded video frames = PPG samples \\
\bottomrule
\end{tabular}
\end{table}

Nine sessions were skipped because their alignment CSV files were absent.
One session failed the prespecified raw detection-rate gate, and one source video was corrupt.
Table~\ref{tab:supp_skipped_sessions} lists the complete skip set.

\begin{table}[htbp]
\centering
\caption{Sessions skipped by the production conversion contract.}
\label{tab:supp_skipped_sessions}
\small
\begin{tabular}{>{\raggedright\arraybackslash}p{5.2cm} >{\raggedright\arraybackslash}p{9.0cm}}
\toprule
Session identifier & Recorded reason \\
\midrule
\texttt{113}--\texttt{120}, \texttt{284} & Alignment CSV missing (nine sessions). \\
\texttt{381} & Raw face-detection rate 0.5995, below the 0.80 gate. \\
\texttt{402} & Corrupt H.264 source; only 7,552 of 7,560 source frames were decodable. \\
\bottomrule
\end{tabular}
\end{table}

\suppsection{sec:supp_split}{S2}{Participant and session split}

The fixed manifest represented 77 original participants and 607 converted sessions.
One of the 78 participants of the 618 source sessions had every session skipped, which is why the converted cohort has 77.
The global split assigned 65 participants and 511 sessions to training and 12 participants and 96 sessions to testing.
Scenario filtering followed the global participant split.
Because each scenario contributed at most one converted session per participant, the participant and session counts are equal within a scenario.

\begin{table}[htbp]
\centering
\caption[Participants and sessions by scenario]{Participants and sessions per scenario.
The overlap columns compare training with testing.}
\label{tab:supp_split}
\small
\setlength{\tabcolsep}{4.5pt}
\begin{tabular}{lrrrrrr}
\toprule
Scenario & \makecell{Train\\participants} & \makecell{Test\\participants} & \makecell{Train\\sessions} & \makecell{Test\\sessions} & \makecell{Participant\\overlap} & \makecell{Session\\overlap} \\
\midrule
Static level~1 & 64 & 12 & 64 & 12 & 0 & 0 \\
Static level~3 & 64 & 12 & 64 & 12 & 0 & 0 \\
Static level~5 & 63 & 12 & 63 & 12 & 0 & 0 \\
Speak & 64 & 12 & 64 & 12 & 0 & 0 \\
Rotate & 64 & 12 & 64 & 12 & 0 & 0 \\
Bike level~1 & 65 & 12 & 65 & 12 & 0 & 0 \\
Bike level~3 & 64 & 12 & 64 & 12 & 0 & 0 \\
Bike level~5 & 63 & 12 & 63 & 12 & 0 & 0 \\
\bottomrule
\end{tabular}
\end{table}

\FloatBarrier
\suppsection{sec:supp_clips}{S3}{Clip retention}

Table~\ref{tab:supp_clip_counts} gives the training and test clip counts per scenario together with the clips the low-variance gate excluded.
``Complete raw clips'' denotes the non-overlapping 256-frame clips available after discarding each session's incomplete terminal remainder but before we applied the label gate.
The eight scenario rows sum to 25,848 complete raw clips, of which 25,483 were retained and 365 were excluded.

\begin{table}[htbp]
\centering
\caption{Clip counts by scenario.}
\label{tab:supp_clip_counts}
\small
\setlength{\tabcolsep}{5pt}
\begin{tabular}{lrrrr}
\toprule
Scenario & Train clips & Test clips & \makecell{Excluded\\train / test} & Complete raw clips \\
\midrule
Static level~1 & 1,553 & 289 & 43 / 13 & 1,898 \\
Static level~3 & 1,562 & 289 & 28 / 10 & 1,889 \\
Static level~5 & 1,529 & 289 & 35 / 7 & 1,860 \\
Speak & 1,554 & 288 & 35 / 23 & 1,900 \\
Rotate & 1,552 & 287 & 27 / 6 & 1,872 \\
Bike level~1 & 4,651 & 853 & 40 / 5 & 5,549 \\
Bike level~3 & 4,577 & 853 & 30 / 6 & 5,466 \\
Bike level~5 & 4,504 & 853 & 47 / 10 & 5,414 \\
\midrule
Total & 21,482 & 4,001 & 285 / 80 & 25,848 \\
\bottomrule
\end{tabular}
\end{table}

A clip was excluded when its ground-truth PPG signal failed one of two checks that the preprocessing applied before any training or testing ran.
The first check rejected a label containing a non-finite 32-bit floating-point value.
The second rejected a label whose 32-bit floating-point standard deviation was no greater than $10^{-6}$, that is, a flat or near-constant trace carrying no measurable pulse.
Either condition alone was sufficient, and no other property of the clip entered the decision.
No model error, predicted heart rate, Pearson correlation, or SNR value determined clip inclusion.
Table~\ref{tab:supp_excluded_clip_sessions} identifies every session in which the gate excluded at least one clip, with the number excluded.
These counts were obtained as $\lfloor N_s/256\rfloor-r_s$, where $N_s$ is the aligned frame count in the session mapping and $r_s$ is the retained file-list count for session $s$.
The preprocessing implementation numbered retained outputs consecutively after skipping an invalid temporal chunk, so the filenames preserve the affected session and exclusion count but not the original within-session temporal index.

\begin{longtable}{>{\raggedright\arraybackslash}p{2.2cm} >{\raggedright\arraybackslash}p{6.0cm} >{\raggedright\arraybackslash}p{6.0cm}}
\caption[Sessions with excluded clips]{Sessions containing clips that the low-variance gate excluded.
Parentheses give the number of excluded 256-frame clips in that session.}
\label{tab:supp_excluded_clip_sessions} \\
\toprule
Scenario & Training sessions (excluded clips) & Test sessions (excluded clips) \\
\midrule
\endfirsthead
\multicolumn{3}{c}{\tablename\ \thetable\ continued} \\
\toprule
Scenario & Training sessions (excluded clips) & Test sessions (excluded clips) \\
\midrule
\endhead
\midrule
\multicolumn{3}{r}{Continued on next page} \\
\endfoot
\bottomrule
\endlastfoot
Static level~1 & \texttt{006} (1), \texttt{054} (1), \texttt{102} (1), \texttt{126} (4), \texttt{134} (1), \texttt{200} (3), \texttt{216} (2), \texttt{232} (1), \texttt{272} (5), \texttt{280} (1), \texttt{352} (2), \texttt{392} (3), \texttt{424} (5), \texttt{432} (6), \texttt{480} (1), \texttt{520} (1), \texttt{536} (1), \texttt{544} (1), \texttt{592} (3) & \texttt{168} (1), \texttt{192} (2), \texttt{224} (1), \texttt{472} (4), \texttt{512} (3), \texttt{576} (2) \\
Static level~3 & \texttt{055} (1), \texttt{095} (1), \texttt{111} (2), \texttt{127} (1), \texttt{135} (1), \texttt{201} (5), \texttt{217} (1), \texttt{281} (1), \texttt{353} (1), \texttt{393} (4), \texttt{425} (1), \texttt{433} (1), \texttt{481} (1), \texttt{489} (1), \texttt{521} (1), \texttt{537} (1), \texttt{593} (3), \texttt{601} (1) & \texttt{225} (1), \texttt{409} (1), \texttt{449} (1), \texttt{473} (4), \texttt{577} (1), \texttt{609} (2) \\
Static level~5 & \texttt{136} (1), \texttt{202} (3), \texttt{218} (1), \texttt{242} (2), \texttt{354} (5), \texttt{394} (4), \texttt{426} (4), \texttt{434} (3), \texttt{482} (1), \texttt{490} (1), \texttt{498} (1), \texttt{522} (3), \texttt{538} (1), \texttt{546} (1), \texttt{594} (3), \texttt{618} (1) & \texttt{144} (1), \texttt{450} (2), \texttt{474} (3), \texttt{578} (1) \\
Speak & \texttt{037} (1), \texttt{053} (1), \texttt{101} (1), \texttt{109} (1), \texttt{125} (1), \texttt{133} (1), \texttt{199} (4), \texttt{215} (1), \texttt{239} (1), \texttt{255} (1), \texttt{271} (1), \texttt{279} (1), \texttt{327} (1), \texttt{351} (2), \texttt{391} (7), \texttt{423} (1), \texttt{431} (1), \texttt{455} (1), \texttt{519} (1), \texttt{535} (2), \texttt{591} (4) & \texttt{167} (1), \texttt{175} (1), \texttt{191} (5), \texttt{407} (1), \texttt{447} (5), \texttt{471} (3), \texttt{511} (3), \texttt{575} (1), \texttt{607} (3) \\
Rotate & \texttt{068} (1), \texttt{084} (1), \texttt{124} (1), \texttt{198} (3), \texttt{214} (1), \texttt{270} (1), \texttt{278} (2), \texttt{350} (1), \texttt{358} (2), \texttt{390} (4), \texttt{422} (1), \texttt{430} (1), \texttt{518} (1), \texttt{534} (2), \texttt{542} (2), \texttt{590} (3) & \texttt{446} (1), \texttt{470} (4), \texttt{574} (1) \\
Bike level~1 & \texttt{081} (1), \texttt{121} (2), \texttt{195} (12), \texttt{203} (3), \texttt{211} (2), \texttt{227} (4), \texttt{235} (1), \texttt{267} (1), \texttt{387} (3), \texttt{419} (1), \texttt{427} (2), \texttt{475} (1), \texttt{491} (1), \texttt{515} (2), \texttt{531} (2), \texttt{587} (2) & \texttt{467} (5) \\
Bike level~3 & \texttt{002} (1), \texttt{098} (1), \texttt{122} (5), \texttt{196} (4), \texttt{212} (2), \texttt{388} (5), \texttt{420} (2), \texttt{428} (2), \texttt{492} (1), \texttt{516} (3), \texttt{532} (2), \texttt{588} (2) & \texttt{468} (5), \texttt{572} (1) \\
Bike level~5 & \texttt{051} (1), \texttt{083} (14), \texttt{123} (2), \texttt{131} (1), \texttt{197} (5), \texttt{213} (1), \texttt{269} (1), \texttt{277} (1), \texttt{389} (5), \texttt{421} (4), \texttt{429} (2), \texttt{477} (3), \texttt{517} (3), \texttt{533} (2), \texttt{589} (2) & \texttt{469} (5), \texttt{573} (5) \\
\end{longtable}

\paragraph{Heart rates at the upper edge of the search band.}
A second kind of loss affects retained clips rather than excluded ones, and we report it here because a reader counting what the analysis could not see should find both in the same place.
The heart rate is searched over 45--150~beats per minute, that is 0.75--2.5~Hz, which is the rPPG-Toolbox default and the band behind the published values we compare with~\citep{liu2023RPPGToolboxToolboxNeurIPS,dehaan2014RobustnessRemotePPGPhysiolMeas}.
Thirty of the 2559 cycling test clips, 1.2~\%, have a reference heart rate at the upper edge of that band and are therefore censored: 17 clips in Bike level~1, 12 in Bike level~5, and 1 in Bike level~3.
No clip of any stationary condition reaches the edge, and no clip of any condition reaches the 45~beats-per-minute lower edge.

Every censored clip belongs to one held-out participant.
The session manifest resolves the three cycling sessions \texttt{467}, \texttt{468} and \texttt{469} of Table~\ref{tab:supp_excluded_clip_sessions} to a single person, and each is the eighth held-out session of its condition.
That participant has the highest median reference heart rate of the twelve in all eight conditions, from 110.0~beats per minute in Static level~1 to 135.9 in Bike level~5, against condition medians of 77.2 to 90.9.
The censored clips fall between 250.9 and 353.3~s, which is the third heart-rate level of the protocol and the rest minute that follows it, so they are the clips at the highest cycling load.

\emph{We kept the 150~beats-per-minute limit rather than widening it, and applied it to the reference and to the prediction alike.}
Widening it is not a post-processing choice.
The archived waveforms are not band-limited, so the reference heart rate can be re-estimated over a wider band without retraining, and doing so changes no clip of any stationary condition.
The prediction cannot follow it.
RhythmFormer's frequency-domain training target lies in the same 45--150 band, so the model does not predict above 150 whatever band the estimator searches, and raising the ceiling on the reference alone enlarges the error by an amount the estimator has no way to avoid, which measures the band rather than the model.
Keeping the band flatters nothing: at these 30 clips the model predicts between 51.2 and 116.4~beats per minute against a reference at the ceiling, so they are already among the largest errors in the evaluation.
Recovering these heart rates requires a model trained with a wider target band.

\suppsection{sec:supp_labelchecks}{S4}{Low-variance label checks}

Two low-variance checks acted on the reported experiment, at different stages and on different objects.
Table~\ref{tab:supp_thresholds} states which threshold defines the reported sample.

\begin{table}[htbp]
\centering
\caption{Low-variance label checks applied in the experiment.}
\label{tab:supp_thresholds}
\small
\begin{tabular}{>{\raggedright\arraybackslash}p{3.4cm} c >{\raggedright\arraybackslash}p{8.9cm}}
\toprule
Stage & Threshold & Interpretation \\
\midrule
Preprocessing & $10^{-6}$ & Clips with a non-finite FP32 label or FP32 standard deviation no greater than the threshold were excluded. \\
Training augmentation fallback & $10^{-8}$ & An invalid augmented label restored the original paired video and label after random draws; it did not exclude any clip. \\
\bottomrule
\end{tabular}
\end{table}

The $10^{-6}$ gate excluded 285 training clips and 80 test clips before training began.
The augmentation fallback acted only on a label produced by a random temporal draw and never removed a clip from the sample.

\suppsection{sec:supp_training}{S5}{Training configuration}

Every scenario used the same training code revision and the same participant manifest, and differed only in the scenario, the run identifier, and the cache, file-list, and output paths.
Table~\ref{tab:supp_training_settings} gives the settings they shared.
During training, a non-finite or low-variance augmented label restored the original video--label pair at a threshold of $10^{-8}$.

\begin{table}[htbp]
\centering
\caption{Settings shared by the eight training runs.}
\label{tab:supp_training_settings}
\small
\begin{tabular}{>{\raggedright\arraybackslash}p{5.2cm} >{\raggedright\arraybackslash}p{9.0cm}}
\toprule
Setting & Value \\
\midrule
Training code revision & \path{3b1b3dbe4c9ace25637ce669495c07ed7a59b521}, identical for all eight runs \\
Model and precision & RhythmFormer; FP32 \\
Random seed & 100 \\
Runtime & PyTorch 2.0.1+cu118; CUDA 11.8 \\
Hardware & Four NVIDIA RTX A5000 graphics processing units \\
Epochs & 30 \\
Training batch size & 4 \\
Test batch size & 2 \\
Initial maximum learning rate & 0.009 \\
Scheduler & OneCycleLR with 30 epochs and the training-loader length as steps per epoch \\
Optimiser & AdamW with weight decay 0 and $\epsilon=10^{-8}$ \\
Drop path rate & 0 \\
Data-loader workers & 16 \\
Temporal input & 256 frames at 50~Hz \\
Validation and checkpoint selection & No validation loader; final epoch used for testing \\
\bottomrule
\end{tabular}
\end{table}

\FloatBarrier
\suppsection{sec:supp_xai}{S6}{Explainable artificial intelligence results}

\begin{figure}[htbp]
\centering
\includegraphics[width=0.92\textwidth]{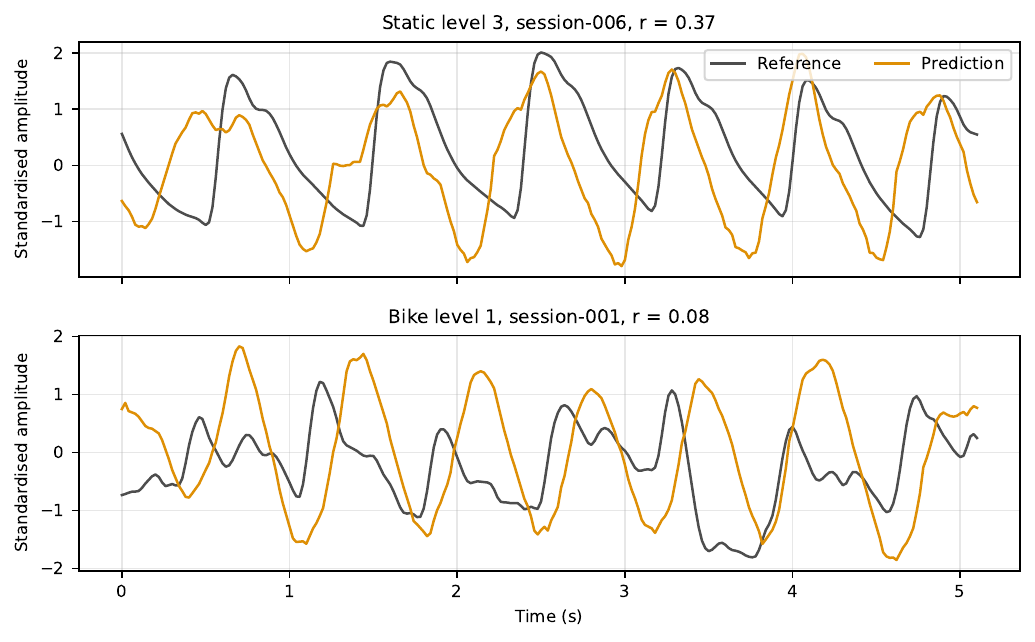}
\caption[Predicted and reference waveforms for two scenarios]{Predicted and reference photoplethysmography waveforms for one held-out clip of Static level~3 and one of Bike level~1.
Each clip is the one whose predicted-to-reference correlation is the median over that scenario's held-out clips, so neither panel is selected for its appearance.
Both signals are standardised, and both clips are 256 frames at 50~Hz, that is 5.12~s.}
\label{fig:supp_waveform_examples}
\end{figure}

Figure~\ref{fig:supp_waveform_examples} contrasts the two ends of the range in Table~\ref{tab:performance_results}.
Static level~3 recovers the heart rate almost exactly, at $r_{\mathrm{HR}}=0.968\pm0.079$, yet its median clip still follows the reference waveform only loosely, while the Bike level~1 clip tracks it hardly at all.
Because each panel is the clip at its condition's median correlation rather than a chosen example, the pair shows directly why the two correlation columns of Table~\ref{tab:performance_results} must be read together.

\begin{figure}[htbp]
\centering
\includegraphics[width=\textwidth]{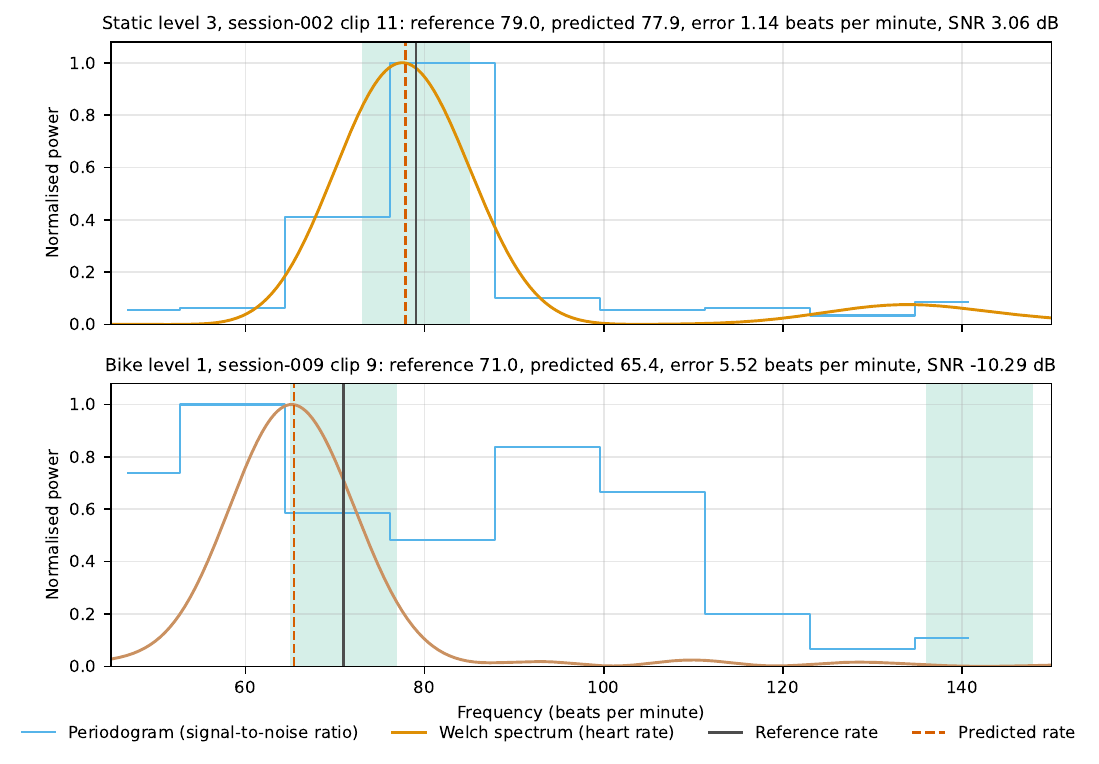}
\caption[Spectra of one median-error clip of two conditions]{The two spectra of one held-out clip of Static level~3 and one of Bike level~1, each the clip whose absolute heart-rate error is the median over that condition's clips, so neither panel is chosen for its appearance.
The Welch spectrum of Equation~\eqref{eq:pearson_hr}, in which the heart rate is located, is drawn against the periodogram of Equation~\eqref{eq:snr}, which the signal-to-noise ratio partitions and which is zero-padded only to the next power of two above the clip.
Both are normalised to their own maximum, since only the location of the power and its distribution matter here.
The shaded bands are that clip's template, $\pm6$~beats per minute of the reference rate and of its second harmonic;
the second harmonic of Static level~3 lies above 150~beats per minute and is therefore outside the plotted range and outside the band the denominator runs over.
}
\label{fig:supp_clip_spectra}
\end{figure}

Figure~\ref{fig:supp_clip_spectra} shows the same contrast in the frequency domain, where both the heart rate and the signal-to-noise ratio are actually read, for the clip of each condition whose absolute heart-rate error is the median of that condition.
On Static level~3 the Welch spectrum carries a single lobe, the rate is estimated at 77.9 against a reference of 79.0, an error of 1.14~beats per minute, and the largest bin of the periodogram, at 82~beats per minute, is the one the six-beat template covers, no other bin holding more than two fifths of its power;
the clip's signal-to-noise ratio is $+3.06$~dB.
On Bike level~1 the rate is estimated at 65.4 against 71.0, an error of 5.52~beats per minute and still inside the template, but the largest periodogram bin lies at 59~beats per minute, below the template altogether, and the bins at 94 and 106 hold four fifths and two thirds of that power, which is what $-10.29$~dB records.
A rate recovered to within six beats per minute therefore coexists with a spectrum that is mostly noise, which is the separation between the two measures that Section~\ref{sec:performance_evaluation} sets out.
The two spectra also show why the two measures are read from different ones: at 256 frames the periodogram resolves 11.7~beats per minute and its bins are wider than the template, while the Welch spectrum, zero-padded to 200,000 points, locates the peak to a fraction of a beat.

\begin{figure}[p]
\centering
\includegraphics[width=0.95\textwidth]{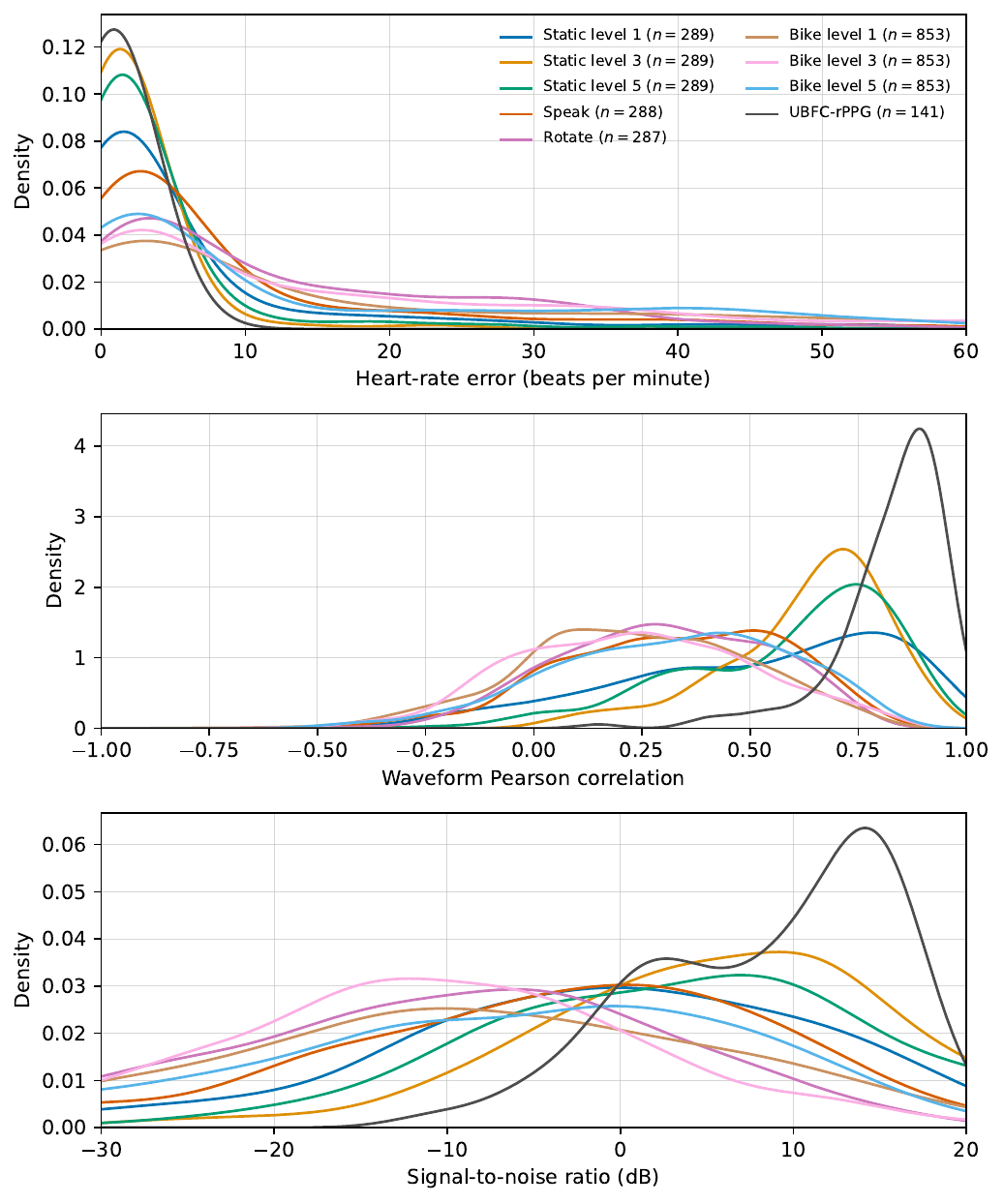}
\caption[Clip-level distributions of the three performance measures]{Kernel densities of the heart-rate error, the waveform Pearson correlation, and the signal-to-noise ratio, with one value per held-out clip and one curve per condition.
The colours are those of Figure~\ref{fig:combined_xai_distributions}, and the sample size of each condition is given in the legend, 4142 clips in total.
The densities are Gaussian with the Silverman bandwidth used for the reference heart-rate distributions, and each is estimated from all of its clips;
the axes are clipped at 60~beats per minute, at $\pm1$, and at $-30$ to $20$~dB, which leaves out the 3.1~\% of clips whose error exceeds 60~beats per minute and the 10.0~\% whose signal-to-noise ratio falls outside the plotted range.
A density is bounded by neither axis limit, so the curves carry the usual edge bias where a measure is bounded, as the waveform correlation is at $+1$.}
\label{fig:supp_metric_distributions}
\end{figure}

Figure~\ref{fig:supp_metric_distributions} gives the clip-level distributions behind the same table, one kernel density per condition for each of the three measures.
Every condition is right-skewed in the heart-rate error and none is bimodal, so the mean of Table~\ref{tab:performance_results} sits above the bulk of its own clips: the median error of Bike level~3 is 8.20~beats per minute against a mean of $16.1\pm2.3$, and that of Static level~3 is 1.14 against $1.96\pm0.34$.
The conditions separate far less cleanly than those means suggest, with the motion densities overlapping the static ones over most of their range and differing mainly in how much weight sits in the tail.
The waveform correlation separates them better than the heart-rate error does, and it is the only one of the three on which UBFC-rPPG stands clear of every NCKU-rPPG condition.

Table~\ref{tab:supp_window_length} varies the evaluation window, which is the one methodological choice in Section~\ref{sec:performance_evaluation} that the reported errors depend on strongly.
Lengthening the window from 5.12 to 61.44~s lowers the mean absolute error in every condition, by 1.4~beats per minute on Static level~3 and 6.6 on Bike level~1, so part of the error even at 5.12~s is an averaging gain rather than pure model error.
The reduction is nonetheless far smaller than the spread between conditions, and it never brings a motion condition near a static one: at 61.44~s the weakest static condition still reads 4.06~beats per minute against 6.63 for the strongest motion condition.
The separation of the two groups therefore holds at every window, and the ordering of Table~\ref{tab:performance_results} holds within each of them.
The order among the motion conditions does not: between 5.12 and 61.44~s Bike level~3 overtakes Rotate, and Bike levels~1 and~5 exchange places.
The last column bounds what any window can achieve: on a single clip the reference photoplethysmogram already disagrees with itself by a median of 0.65--1.09~beats per minute when the participant is stationary and by a mean of 5.49--8.54 when cycling.

\begin{table}[htbp]
\centering
\caption[Evaluation window length and heart-rate error]{Mean absolute error and root mean square error in beats per minute, written as MAE/RMSE, against the number of consecutive clips concatenated into one evaluation window.
One clip is 256 frames at 50~Hz, that is 5.12~s, and the manuscript reports the one-clip column.
Windows do not overlap and never cross a participant boundary, so the number of windows falls as the width grows.
The last column gives the reference photoplethysmogram's agreement with itself on a single clip, as median/mean absolute difference between its spectral heart rate and the heart rate from the median interval between its detected systolic peaks.
The sweep uses the same spectral estimator as the evaluation, whose Welch segment is capped at 256 samples, so a longer window buys segment averaging rather than frequency resolution.
It recomputes every heart rate from the archived waveforms, whereas Table~\ref{tab:performance_results} uses the per-clip values of the explainability run, so the one-clip column differs from that table's mean absolute error by up to 0.7~beats per minute.}
\label{tab:supp_window_length}
\footnotesize
\setlength{\tabcolsep}{4pt}
\begin{tabular}{lccccc c}
\toprule
& \multicolumn{5}{c}{Window (clips, seconds)} & Reference \\
\cmidrule(lr){2-6}\cmidrule(lr){7-7}
Scenario & 1, 5.12 & 2, 10.24 & 4, 20.48 & 6, 30.72 & 12, 61.44 & self-agreement \\
\midrule
Static level~1 & $5.69/12.0$ & $5.18/9.8$ & $4.80/8.7$ & $4.95/8.7$ & $4.06/7.0$ & $0.77/1.81$ \\
Static level~3 & $2.00/3.7$ & $0.97/1.4$ & $0.68/1.0$ & $0.62/0.9$ & $0.63/1.0$ & $0.78/1.69$ \\
Static level~5 & $3.71/8.8$ & $1.98/5.2$ & $1.43/4.0$ & $1.03/1.9$ & $1.11/2.1$ & $0.78/1.11$ \\
Speak & $9.58/16.3$ & $8.24/15.3$ & $6.58/13.4$ & $7.04/14.6$ & $6.63/14.1$ & $1.09/1.92$ \\
Rotate & $12.77/18.2$ & $11.30/16.3$ & $11.48/16.2$ & $10.85/15.4$ & $11.59/16.2$ & $0.65/1.74$ \\
Bike level~1 & $18.68/30.7$ & $15.73/27.1$ & $14.83/25.8$ & $14.10/25.3$ & $12.08/23.3$ & $1.81/7.61$ \\
Bike level~3 & $15.36/23.2$ & $13.17/20.2$ & $11.41/18.0$ & $12.26/19.5$ & $10.55/16.7$ & $1.81/5.49$ \\
Bike level~5 & $15.10/23.9$ & $14.19/22.9$ & $13.58/22.5$ & $13.83/22.9$ & $13.21/22.4$ & $2.47/8.54$ \\
\bottomrule

\end{tabular}
\end{table}

\paragraph{Video-to-photoplethysmogram synchronisation.}
Each recording is synchronised on its own light-off trigger \citep{wang2020NoncontactMeasurementNCKU}, so a synchronisation error is a property of one participant in one recording and is constant within it.
Table~\ref{tab:supp_waveform_lag} fits that constant by maximising the participant's mean clip waveform correlation and reports the correlation before and after.
The correction is large on NCKU-rPPG and absent on UBFC-rPPG.
Its median magnitude is 290--700~milliseconds across the eight NCKU-rPPG conditions and 0~milliseconds on UBFC-rPPG, where no fitted lag exceeds two samples on eleven of the twelve subjects.

\emph{Fitting one lag per participant and recording spends 108 free parameters on the data the correlation is then read from, so we report the gain against a null.}
The null repeats the identical fit with each participant's predictions paired against another participant's reference, where no true lag exists and any gain measures what the fitting itself buys.
On NCKU-rPPG the gain exceeds that floor by two to seven times, from $+0.343$ against $+0.052$ on Static level~3 to $+0.171$ against $+0.083$ on Rotate.
On UBFC-rPPG the gain of $+0.043$ is \emph{below} its own null of $+0.110$, which is what a correctly synchronised recording looks like.
That the same architecture and the same code need a large correction on one dataset and none on the other places the cause in the recording rather than in the model.
Every waveform correlation reported in this manuscript uses the fitted lag of Table~\ref{tab:supp_waveform_lag}.

The waveform column of Table~\ref{tab:performance_results} is not computed from the same predicted signal as the five columns beside it.
The mean absolute error, the root mean square error, the mean absolute percentage error, the heart-rate correlation, and the signal-to-noise ratio are the per-clip values of the explainability run, while the waveform correlation is computed from the predicted waveforms the evaluation run wrote, which are the ones a per-recording lag can be fitted to.
The two runs used the same checkpoint on the same clips, and their reference traces agree exactly, clip for clip, in every condition;
their predictions do not.
Recomputing the heart rate from the evaluation run's waveforms with the estimator of Section~\ref{sec:performance_evaluation} moves it from the value Table~\ref{tab:performance_results} reports by a median of 0.23~beats per minute on Static level~3 and 0.44 on Bike level~1, with single clips moving by up to 17 and 67, and moves the signal-to-noise ratio by a median of 1.2 and 1.7~decibels.
Anyone recomputing one column from one of the two sets of predictions will therefore not reproduce the other's value exactly, and the discrepancy is larger under cycling than under the stationary conditions.

\begingroup
\renewcommand{\multirow}[3]{#3}
\scriptsize
\setlength{\tabcolsep}{5pt}
\begin{longtable}{llccl}
\caption[Fitted video-to-photoplethysmogram lag]{Waveform correlation before and after fitting one lag per participant and recording, with the fitted lag in milliseconds.
A positive lag means the prediction is delayed relative to the reference.
Participants are numbered within a condition and the numbering is consistent across conditions.
The search spans $\pm1$~second, which covers half a beat period down to 30~beats per minute, and the maximiser is interior for every participant except two of Bike level~5, whose fitted lag sits on the $+1$~s edge and is therefore censored;
a lag is recoverable only modulo the beat period and the smallest-magnitude maximiser is taken.
The null is the same fit applied to mismatched participant pairs, where no true lag exists.}
\label{tab:supp_waveform_lag} \\
\toprule
Condition & Participant & $r_{\mathrm{wave}}$ at lag 0 & $r_{\mathrm{wave}}$ at fitted lag & Lag (ms) \\
\midrule
\endfirsthead
\multicolumn{5}{c}{\tablename~\thetable\ continued from the previous page.} \\
\toprule
Condition & Participant & $r_{\mathrm{wave}}$ at lag 0 & $r_{\mathrm{wave}}$ at fitted lag & Lag (ms) \\
\midrule
\endhead
\multirow{12}{*}{Static level~1} & 1 & $-0.408$ & $0.663$ & $-480$ \\
 & 2 & $0.840$ & $0.859$ & $+20$ \\
 & 3 & $-0.464$ & $0.696$ & $-460$ \\
 & 4 & $0.670$ & $0.767$ & $-40$ \\
 & 5 & $-0.091$ & $0.505$ & $-260$ \\
 & 6 & $0.599$ & $0.611$ & $-20$ \\
 & 7 & $0.051$ & $0.132$ & $-780$ \\
 & 8 & $0.071$ & $0.131$ & $-540$ \\
 & 9 & $-0.017$ & $0.238$ & $+480$ \\
 & 10 & $0.818$ & $0.835$ & $-700$ \\
 & 11 & $0.203$ & $0.429$ & $+500$ \\
 & 12 & $-0.143$ & $0.326$ & $-420$ \\
 & mean & $0.177$ & $0.516$ & gain $+0.338$, null $+0.070$ \\
\midrule
\multirow{12}{*}{Static level~3} & 1 & $-0.004$ & $0.635$ & $-640$ \\
 & 2 & $0.505$ & $0.660$ & $+700$ \\
 & 3 & $-0.668$ & $0.733$ & $-440$ \\
 & 4 & $0.409$ & $0.643$ & $+580$ \\
 & 5 & $0.585$ & $0.606$ & $-40$ \\
 & 6 & $0.229$ & $0.766$ & $+700$ \\
 & 7 & $0.175$ & $0.495$ & $+540$ \\
 & 8 & $-0.153$ & $0.358$ & $-920$ \\
 & 9 & $0.430$ & $0.598$ & $+840$ \\
 & 10 & $0.729$ & $0.856$ & $+660$ \\
 & 11 & $0.614$ & $0.614$ & $+0$ \\
 & 12 & $0.547$ & $0.548$ & $+20$ \\
 & mean & $0.283$ & $0.626$ & gain $+0.343$, null $+0.052$ \\
\midrule
\multirow{12}{*}{Static level~5} & 1 & $0.233$ & $0.597$ & $-660$ \\
 & 2 & $0.507$ & $0.744$ & $+620$ \\
 & 3 & $-0.130$ & $0.457$ & $-460$ \\
 & 4 & $0.554$ & $0.619$ & $+600$ \\
 & 5 & $0.128$ & $0.496$ & $+620$ \\
 & 6 & $0.287$ & $0.698$ & $+700$ \\
 & 7 & $0.781$ & $0.796$ & $-660$ \\
 & 8 & $0.140$ & $0.307$ & $-700$ \\
 & 9 & $0.544$ & $0.721$ & $+780$ \\
 & 10 & $0.825$ & $0.842$ & $+680$ \\
 & 11 & $0.446$ & $0.541$ & $+580$ \\
 & 12 & $0.012$ & $0.233$ & $+540$ \\
 & mean & $0.361$ & $0.588$ & gain $+0.227$, null $+0.070$ \\
\midrule
\multirow{12}{*}{Speak} & 1 & $0.209$ & $0.525$ & $-620$ \\
 & 2 & $0.246$ & $0.465$ & $+600$ \\
 & 3 & $-0.352$ & $0.435$ & $-500$ \\
 & 4 & $0.144$ & $0.200$ & $+680$ \\
 & 5 & $0.214$ & $0.295$ & $+720$ \\
 & 6 & $0.358$ & $0.402$ & $-60$ \\
 & 7 & $0.337$ & $0.379$ & $-620$ \\
 & 8 & $0.156$ & $0.222$ & $-560$ \\
 & 9 & $0.060$ & $0.102$ & $+680$ \\
 & 10 & $0.452$ & $0.549$ & $+640$ \\
 & 11 & $0.172$ & $0.192$ & $+540$ \\
 & 12 & $-0.078$ & $0.121$ & $-440$ \\
 & mean & $0.160$ & $0.324$ & gain $+0.164$, null $+0.054$ \\
\midrule
\multirow{12}{*}{Rotate} & 1 & $-0.291$ & $0.329$ & $-380$ \\
 & 2 & $0.284$ & $0.356$ & $-820$ \\
 & 3 & $-0.387$ & $0.497$ & $-380$ \\
 & 4 & $0.289$ & $0.420$ & $-720$ \\
 & 5 & $0.310$ & $0.327$ & $-860$ \\
 & 6 & $0.273$ & $0.285$ & $+40$ \\
 & 7 & $0.312$ & $0.368$ & $-680$ \\
 & 8 & $-0.029$ & $0.072$ & $-900$ \\
 & 9 & $0.184$ & $0.234$ & $+720$ \\
 & 10 & $0.458$ & $0.507$ & $-40$ \\
 & 11 & $0.041$ & $0.046$ & $+540$ \\
 & 12 & $0.105$ & $0.160$ & $+960$ \\
 & mean & $0.129$ & $0.300$ & gain $+0.171$, null $+0.083$ \\
\midrule
\multirow{12}{*}{Bike level~1} & 1 & $-0.022$ & $0.272$ & $+960$ \\
 & 2 & $0.433$ & $0.479$ & $+40$ \\
 & 3 & $-0.018$ & $0.082$ & $-520$ \\
 & 4 & $0.351$ & $0.403$ & $+60$ \\
 & 5 & $0.067$ & $0.075$ & $+680$ \\
 & 6 & $0.161$ & $0.268$ & $+640$ \\
 & 7 & $-0.020$ & $0.255$ & $-640$ \\
 & 8 & $0.047$ & $0.099$ & $-900$ \\
 & 9 & $0.079$ & $0.081$ & $-20$ \\
 & 10 & $-0.127$ & $0.254$ & $-400$ \\
 & 11 & $-0.006$ & $0.086$ & $-720$ \\
 & 12 & $0.196$ & $0.463$ & $+140$ \\
 & mean & $0.095$ & $0.235$ & gain $+0.140$, null $+0.046$ \\
\midrule
\multirow{12}{*}{Bike level~3} & 1 & $-0.166$ & $0.177$ & $+280$ \\
 & 2 & $0.437$ & $0.443$ & $-20$ \\
 & 3 & $-0.018$ & $0.166$ & $-580$ \\
 & 4 & $0.073$ & $0.193$ & $+560$ \\
 & 5 & $-0.035$ & $0.107$ & $-240$ \\
 & 6 & $0.086$ & $0.120$ & $-760$ \\
 & 7 & $0.205$ & $0.273$ & $-700$ \\
 & 8 & $-0.005$ & $0.134$ & $-240$ \\
 & 9 & $-0.023$ & $0.095$ & $-900$ \\
 & 10 & $0.436$ & $0.485$ & $+520$ \\
 & 11 & $0.030$ & $0.066$ & $-420$ \\
 & 12 & $0.397$ & $0.528$ & $+80$ \\
 & mean & $0.118$ & $0.232$ & gain $+0.114$, null $+0.040$ \\
\midrule
\multirow{12}{*}{Bike level~5} & 1 & $-0.085$ & $0.151$ & $-400$ \\
 & 2 & $0.398$ & $0.435$ & $+700$ \\
 & 3 & $0.428$ & $0.485$ & $+60$ \\
 & 4 & $0.367$ & $0.399$ & $+40$ \\
 & 5 & $0.101$ & $0.128$ & $-80$ \\
 & 6 & $0.433$ & $0.462$ & $+40$ \\
 & 7 & $0.300$ & $0.394$ & $-700$ \\
 & 8 & $0.045$ & $0.102$ & $+1000$ \\
 & 9 & $0.570$ & $0.586$ & $+20$ \\
 & 10 & $0.099$ & $0.165$ & $+1000$ \\
 & 11 & $0.050$ & $0.187$ & $+440$ \\
 & 12 & $0.025$ & $0.339$ & $+180$ \\
 & mean & $0.228$ & $0.319$ & gain $+0.092$, null $+0.039$ \\
\midrule
\multirow{12}{*}{UBFC-rPPG} & 1 & $0.679$ & $0.688$ & $+33$ \\
 & 2 & $0.872$ & $0.908$ & $-33$ \\
 & 3 & $0.874$ & $0.874$ & $+0$ \\
 & 4 & $0.896$ & $0.896$ & $+0$ \\
 & 5 & $0.497$ & $0.714$ & $+67$ \\
 & 6 & $0.802$ & $0.802$ & $+0$ \\
 & 7 & $0.889$ & $0.889$ & $+0$ \\
 & 8 & $0.482$ & $0.734$ & $+733$ \\
 & 9 & $0.850$ & $0.850$ & $+0$ \\
 & 10 & $0.834$ & $0.834$ & $+0$ \\
 & 11 & $0.909$ & $0.909$ & $+0$ \\
 & 12 & $0.838$ & $0.838$ & $+0$ \\
 & mean & $0.785$ & $0.828$ & gain $+0.043$, null $+0.110$ \\
\bottomrule

\end{longtable}
\endgroup

\paragraph{Why the window was not lengthened.}
Table~\ref{tab:supp_window_length} shows the error falling as the window grows, which invites the question why 5.12~s was kept.
Stability alone cannot answer it, because a longer window always steadies a spectral estimate and the steadiest window is the whole recording, which returns one number for a quantity that moved.
The window has to be judged against a target that needs no window, and the beat-to-beat heart rate of the reference photoplethysmogram, taken from its systolic-peak intervals, is one.
Against that target the error splits in two, as Table~\ref{tab:supp_window_bias_variance} reports.
The estimation part, the spectral estimate minus the mean reference over the same window, falls as more cycles are averaged.
The representation part, that window mean minus the instantaneous reference at the window centre, grows as the heart rate moves inside the window.
Their sum has a minimum, and that minimum is the optimal window.

On Static level~3 the estimation error falls from 2.62 to 0.84~beats per minute between 5.12 and 61.44~s while the representation error rises from 3.42 to 4.27, and the total is least at 7.68~s.
The curve is shallow: 3.34~beats per minute there against 3.60 at 5.12~s, a penalty of 7~\% for the clip the manuscript uses.
We keep 5.12~s because it is what makes the UBFC-rPPG row of Table~\ref{tab:performance_results} comparable at all, its clip being 5.33~s, and 7~\% of a quantity that already carries a standard error of 0.34~beats per minute is not worth losing that comparison for.

Two limits of this decomposition are worth stating.
The cycling minima, 10.24~s on Bike level~3 and 8.96 on Bike level~1, sit on curves that vary by 17 and 13~\% of their own minimum across a twelvefold range of window length, and at 3.7 and 4.8 times the Static level~3 minimum of 3.34~beats per minute, because there the beat-to-beat reference is itself motion-corrupted;
no window is good under cycling, and the limitation is the reference rather than the window.
The instantaneous target also carries respiratory sinus arrhythmia and detection jitter, which inflate the representation term and bias the optimum short, so 7.68~s should be read as a lower bound on the best window rather than a precise value.

\begin{table}[htbp]
\centering
\caption[Choosing the evaluation window]{Root mean square error of the spectral heart rate against the beat-to-beat heart rate of the reference photoplethysmogram, in beats per minute, decomposed by window length.
\emph{Estimation} is the spectral estimate minus the mean reference over the same window and falls with length;
\emph{representation} is that window mean minus the instantaneous reference at the window centre and grows with it;
\emph{total} is the error against the instantaneous reference and is least at the row in bold.
Windows advance one second at a time and every value is the mean over the 12 held-out participants.}
\label{tab:supp_window_bias_variance}
\footnotesize
\setlength{\tabcolsep}{6pt}
\begin{tabular}{llccc}
\toprule
Scenario & Window (s) & Estimation & Representation & Total \\
\midrule
\multirow{12}{*}{Static Level~3} & $5.12$ & $2.62$ & $3.42$ & $3.60$ \\
 & $6.40$ & $2.61$ & $3.33$ & $3.60$ \\
 & $\mathbf{7.68}$ & $\mathbf{1.96}$ & $\mathbf{3.56}$ & $\mathbf{3.34}$ \\
 & $8.96$ & $2.11$ & $3.69$ & $3.38$ \\
 & $10.24$ & $1.78$ & $4.24$ & $4.28$ \\
 & $11.52$ & $2.03$ & $4.24$ & $4.26$ \\
 & $12.80$ & $1.55$ & $4.10$ & $4.30$ \\
 & $15.36$ & $1.46$ & $4.36$ & $4.43$ \\
 & $20.48$ & $1.31$ & $4.51$ & $4.50$ \\
 & $30.72$ & $0.99$ & $4.56$ & $4.68$ \\
 & $40.96$ & $0.89$ & $5.01$ & $5.12$ \\
 & $61.44$ & $0.84$ & $4.27$ & $4.25$ \\
\midrule
\multirow{12}{*}{Bike Level~3} & $5.12$ & $9.48$ & $9.39$ & $12.93$ \\
 & $6.40$ & $9.90$ & $9.72$ & $13.41$ \\
 & $7.68$ & $9.58$ & $10.06$ & $13.27$ \\
 & $8.96$ & $9.86$ & $9.42$ & $12.86$ \\
 & $\mathbf{10.24}$ & $\mathbf{8.81}$ & $\mathbf{9.38}$ & $\mathbf{12.23}$ \\
 & $11.52$ & $8.91$ & $10.39$ & $14.32$ \\
 & $12.80$ & $8.42$ & $10.25$ & $13.19$ \\
 & $15.36$ & $8.15$ & $9.78$ & $12.91$ \\
 & $20.48$ & $7.63$ & $9.97$ & $12.66$ \\
 & $30.72$ & $6.93$ & $10.75$ & $13.20$ \\
 & $40.96$ & $6.75$ & $10.83$ & $12.73$ \\
 & $61.44$ & $6.36$ & $11.59$ & $13.12$ \\
\midrule
\multirow{12}{*}{Bike Level~1} & $5.12$ & $13.38$ & $10.26$ & $16.20$ \\
 & $6.40$ & $13.64$ & $10.66$ & $16.67$ \\
 & $7.68$ & $12.89$ & $10.75$ & $16.63$ \\
 & $\mathbf{8.96}$ & $\mathbf{13.16}$ & $\mathbf{10.45}$ & $\mathbf{16.06}$ \\
 & $10.24$ & $12.93$ & $11.78$ & $17.56$ \\
 & $11.52$ & $12.99$ & $10.88$ & $17.09$ \\
 & $12.80$ & $12.57$ & $11.58$ & $17.49$ \\
 & $15.36$ & $12.49$ & $11.62$ & $17.06$ \\
 & $20.48$ & $12.36$ & $11.92$ & $17.31$ \\
 & $30.72$ & $12.09$ & $12.08$ & $17.67$ \\
 & $40.96$ & $12.34$ & $12.37$ & $18.18$ \\
 & $61.44$ & $11.96$ & $12.41$ & $18.10$ \\
\bottomrule

\end{tabular}
\end{table}

Table~\ref{tab:supp_xai_spearman_full} gives all 252 participant-averaged Spearman estimates from the arrays plotted in Figure~\ref{fig:combined_xai_scatter}.

\begin{landscape}
{\scriptsize
\setlength{\tabcolsep}{2.5pt}
\renewcommand{\arraystretch}{1.06}
\begin{longtable}{>{\raggedright\arraybackslash}p{2.6cm} >{\raggedright\arraybackslash}p{1.15cm} ccccccc}
\caption{Two-sided Spearman summaries for all eight NCKU-rPPG scenarios and UBFC-rPPG, taken within each participant and averaged over participants.
Each cell gives $\rho; p$; coefficients are displayed to two decimal places, and $p$ values to three.
The seven relationships pair each of the three clip-level performance measures, heart-rate error, waveform correlation and signal-to-noise ratio, with each of the two attribution summaries, and add SaCo against skin coverage.
Cells in bold satisfy both $p<0.05$ and $|\rho|\ge0.10$, the condition Table~\ref{tab:xai_direction_counts} counts as directional.
The first sample-size pair in each condition label gives the number of skin-only clips followed by the number of clips in relationships involving SaCo;
the second gives the participants contributing to each.
Those differ because the SaCo run does not cover every held-out participant: it reaches nine of twelve in the five stationary conditions and only three of twelve in the three cycling conditions, so a SaCo relationship there is averaged over three people and should be read accordingly.}
\label{tab:supp_xai_spearman_full} \\
\toprule
Condition & Method & MAE--skin & MAE--SaCo & Pearson--skin & Pearson--SaCo & SNR--skin & SNR--SaCo & SaCo--skin \\
\midrule
\endfirsthead
\multicolumn{9}{c}{\tablename\ \thetable\ (continued)} \\
\toprule
Condition & Method & MAE--skin & MAE--SaCo & Pearson--skin & Pearson--SaCo & SNR--skin & SNR--SaCo & SaCo--skin \\
\midrule
\endhead
\bottomrule
\endfoot
\multirow{4}{*}{\makecell[l]{Static level 1\\($n=289/200$)\\($P=12/9$)}} & raw & $-0.02; .709$ & $\mathbf{-0.19; .013}$ & $\mathbf{0.14; .046}$ & $0.06; .522$ & $0.06; .294$ & $0.13; .114$ & $0.06; .596$ \\*
 & rollout & $-0.07; .295$ & $-0.11; .063$ & $0.03; .567$ & $0.00; .985$ & $0.06; .296$ & $0.03; .658$ & $0.04; .506$ \\*
 & flow & $0.03; .414$ & $-0.03; .829$ & $0.11; .166$ & $0.04; .563$ & $0.08; .072$ & $0.00; .997$ & $0.03; .785$ \\*
 & BI & $0.14; .148$ & $0.14; .080$ & $-0.09; .443$ & $\mathbf{-0.22; .019}$ & $-0.17; .118$ & $-0.01; .922$ & $0.11; .145$ \\
\midrule
\multirow{4}{*}{\makecell[l]{Static level 3\\($n=289/200$)\\($P=12/9$)}} & raw & $\mathbf{-0.14; .020}$ & $0.01; .953$ & $0.13; .051$ & $0.11; .293$ & $-0.06; .390$ & $0.02; .794$ & $0.09; .427$ \\*
 & rollout & $0.04; .508$ & $-0.02; .821$ & $0.02; .750$ & $0.12; .284$ & $0.00; .970$ & $0.01; .874$ & $0.03; .853$ \\*
 & flow & $0.09; .111$ & $-0.00; .944$ & $0.05; .422$ & $0.02; .864$ & $-0.05; .612$ & $0.04; .764$ & $-0.13; .125$ \\*
 & BI & $0.06; .513$ & $-0.19; .344$ & $-0.16; .086$ & $0.02; .809$ & $-0.08; .218$ & $\mathbf{0.19; .044}$ & $-0.13; .264$ \\
\midrule
\multirow{4}{*}{\makecell[l]{Static level 5\\($n=289/200$)\\($P=12/9$)}} & raw & $-0.02; .725$ & $-0.02; .694$ & $0.11; .061$ & $0.00; .976$ & $0.02; .798$ & $0.13; .213$ & $0.04; .464$ \\*
 & rollout & $-0.08; .213$ & $0.06; .492$ & $0.07; .208$ & $-0.02; .834$ & $0.06; .291$ & $0.06; .553$ & $-0.07; .482$ \\*
 & flow & $-0.00; .974$ & $0.05; .434$ & $-0.00; .961$ & $-0.14; .147$ & $-0.07; .247$ & $-0.00; .996$ & $-0.12; .179$ \\*
 & BI & $-0.11; .091$ & $-0.07; .165$ & $0.02; .793$ & $0.14; .156$ & $0.06; .390$ & $\mathbf{0.18; .035}$ & $\mathbf{0.17; .031}$ \\
\midrule
\multirow{4}{*}{\makecell[l]{Speak\\($n=288/200$)\\($P=12/9$)}} & raw & $0.05; .552$ & $-0.11; .450$ & $-0.02; .690$ & $0.05; .535$ & $0.07; .106$ & $0.02; .819$ & $-0.18; .092$ \\*
 & rollout & $0.04; .454$ & $-0.11; .398$ & $\mathbf{0.15; .042}$ & $0.07; .302$ & $-0.03; .599$ & $0.03; .776$ & $-0.01; .845$ \\*
 & flow & $0.02; .789$ & $0.03; .676$ & $\mathbf{0.19; .050}$ & $0.07; .425$ & $-0.04; .638$ & $-0.02; .841$ & $-0.04; .556$ \\*
 & BI & $\mathbf{0.24; {<}10^{-3}}$ & $0.04; .575$ & $\mathbf{-0.19; .038}$ & $-0.02; .832$ & $\mathbf{-0.25; {<}10^{-3}}$ & $-0.13; .329$ & $0.17; .186$ \\
\midrule
\multirow{4}{*}{\makecell[l]{Rotate\\($n=287/200$)\\($P=12/9$)}} & raw & $-0.05; .456$ & $0.12; .142$ & $-0.00; .973$ & $-0.16; .063$ & $0.03; .707$ & $-0.02; .794$ & $0.03; .617$ \\*
 & rollout & $\mathbf{-0.22; .023}$ & $0.06; .457$ & $0.16; .082$ & $-0.10; .204$ & $0.13; .143$ & $0.04; .625$ & $-0.15; .132$ \\*
 & flow & $-0.05; .498$ & $-0.05; .517$ & $0.00; .974$ & $-0.02; .731$ & $0.03; .777$ & $0.05; .444$ & $0.09; .339$ \\*
 & BI & $0.07; .421$ & $0.08; .285$ & $-0.09; .250$ & $\mathbf{-0.22; .005}$ & $0.02; .781$ & $\mathbf{-0.18; .041}$ & $\mathbf{0.22; .020}$ \\
\midrule
\multirow{4}{*}{\makecell[l]{Bike level 1\\($n=853/200$)\\($P=12/3$)}} & raw & $0.01; .910$ & $0.02; .671$ & $-0.01; .824$ & $-0.04; .789$ & $0.02; .670$ & $0.10; .324$ & $-0.01; .899$ \\*
 & rollout & $-0.00; .941$ & $0.03; .602$ & $-0.04; .235$ & $-0.02; .894$ & $0.01; .913$ & $0.08; .355$ & $-0.00; .939$ \\*
 & flow & $-0.05; .160$ & $0.05; .731$ & $-0.00; .938$ & $-0.07; .717$ & $0.02; .513$ & $-0.02; .906$ & $0.03; .598$ \\*
 & BI & $\mathbf{0.24; {<}10^{-3}}$ & $0.04; .788$ & $-0.10; .027$ & $-0.07; .451$ & $\mathbf{-0.12; .011}$ & $0.05; .715$ & $0.29; .206$ \\
\midrule
\multirow{4}{*}{\makecell[l]{Bike level 3\\($n=853/200$)\\($P=12/3$)}} & raw & $-0.01; .762$ & $-0.03; .857$ & $0.00; .993$ & $0.08; .562$ & $0.08; .082$ & $0.08; .279$ & $-0.06; .703$ \\*
 & rollout & $-0.07; .163$ & $-0.01; .931$ & $0.06; .113$ & $0.09; .423$ & $0.07; .134$ & $0.11; .118$ & $-0.07; .064$ \\*
 & flow & $-0.04; .599$ & $\mathbf{0.13; .023}$ & $0.04; .261$ & $-0.05; .797$ & $0.02; .715$ & $-0.06; .394$ & $0.11; .586$ \\*
 & BI & $0.06; .253$ & $-0.00; .936$ & $-0.05; .207$ & $-0.05; .591$ & $-0.01; .736$ & $-0.03; .475$ & $-0.05; .300$ \\
\midrule
\multirow{4}{*}{\makecell[l]{Bike level 5\\($n=853/200$)\\($P=12/3$)}} & raw & $-0.01; .727$ & $-0.05; .015$ & $0.02; .601$ & $0.02; .887$ & $-0.03; .424$ & $0.03; .405$ & $0.03; .597$ \\*
 & rollout & $\mathbf{0.12; .012}$ & $-0.05; .102$ & $-0.03; .617$ & $0.07; .439$ & $-0.10; .029$ & $0.01; .790$ & $0.15; .555$ \\*
 & flow & $-0.02; .651$ & $-0.06; .592$ & $0.02; .701$ & $\mathbf{0.13; .021}$ & $-0.01; .788$ & $0.05; .512$ & $0.09; .445$ \\*
 & BI & $\mathbf{0.11; .009}$ & $-0.02; .603$ & $0.02; .570$ & $-0.02; .851$ & $-0.09; .101$ & $-0.04; .143$ & $0.17; .417$ \\
\midrule
\multirow{4}{*}{\makecell[l]{UBFC-rPPG\\($n=141/141$)\\($P=12/12$)}} & raw & $-0.05; .714$ & $-0.01; .853$ & $-0.12; .372$ & $0.01; .914$ & $-0.08; .471$ & $-0.03; .821$ & $-0.08; .656$ \\*
 & rollout & $-0.09; .405$ & $-0.05; .490$ & $\mathbf{0.32; .047}$ & $0.06; .495$ & $0.15; .113$ & $0.01; .931$ & $0.01; .869$ \\*
 & flow & $-0.07; .326$ & $0.02; .848$ & $\mathbf{0.21; .041}$ & $0.06; .533$ & $-0.01; .932$ & $-0.03; .796$ & $0.14; .066$ \\*
 & BI & $-0.20; .073$ & $-0.02; .800$ & $0.25; .062$ & $0.05; .659$ & $0.13; .145$ & $0.04; .773$ & $\mathbf{0.28; .007}$ \\

\end{longtable}
}
\end{landscape}

Table~\ref{tab:supp_xai_spearman_full} is mostly empty of resolved relationships, and that is its result.
Twenty-five of the 252 coefficients are directional, between one and five in each of the nine conditions, and the largest of them reaches $\rho=+0.32$.
No condition resolves more than five, no relationship resolves in more than six of its 36 method--condition pairs, and Beyond Intuition supplies 14 of the 25, more than the three attention-only methods together, which give 3, 4 and 4.
Within a participant, then, neither skin coverage nor SaCo predicts how well the model does on a given clip in any condition, which supports the conclusion drawn from the scenario-level comparison in Section~\ref{sec:scenario_reliability_results} on evidence of a different kind.

\clearpage
\begin{figure}[p]
\centering
\begin{subfigure}[t]{0.94\textwidth}
\centering
\includegraphics[width=\linewidth]{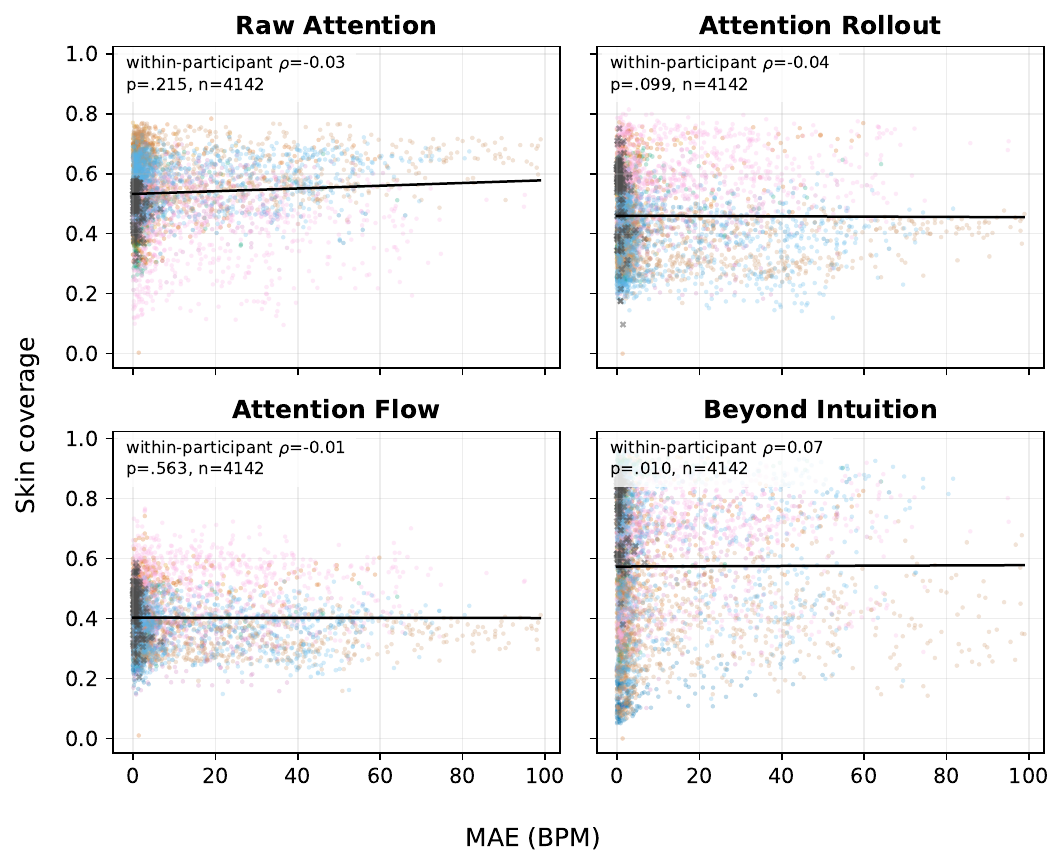}
\caption{MAE--skin coverage.}
\end{subfigure}
\caption{Combined descriptive scatter plots for raw attention, attention rollout, attention flow, and Beyond Intuition across all eight NCKU-rPPG scenarios and UBFC-rPPG.
NCKU-rPPG clips use circular markers and UBFC-rPPG clips use crosses; no observations are sampled, and all panels use linear axes that retain the complete data ranges.
Each method panel contains one ordinary least-squares line fitted to all displayed clips, which is a visual aid only.
Each panel is annotated with the same estimator the tables report, the Spearman correlation taken inside a participant and averaged through the Fisher transform, with a one-sample $t$ test over participants.
The panel pools all nine conditions, treating each participant of each condition as its own group, so its coefficient summarises the whole display rather than reproducing any single row of Table~\ref{tab:static3_ubfc_spearman} or \xaisuppspearmanreference;
restricted to one condition the identical estimator reproduces that table's entry exactly.
Colours denote Static level~1 (blue, \texttt{\#0173B2}), Static level~3 (orange, \texttt{\#DE8F05}), Static level~5 (green, \texttt{\#029E73}), Speak (red-orange, \texttt{\#D55E00}), Rotate (purple, \texttt{\#CC78BC}), Bike level~1 (brown, \texttt{\#CA9161}), Bike level~3 (pink, \texttt{\#FBAFE4}), Bike level~5 (light blue, \texttt{\#56B4E9}), and UBFC-rPPG (dark grey, \texttt{\#4D4D4D}).
Table~\ref{tab:static3_ubfc_spearman} reports the Static level~3 and UBFC-rPPG Spearman summaries, and \xaisuppspearmanreference\ reports all nine groups.}
\label{fig:combined_xai_scatter}
\end{figure}
\clearpage
\begin{figure}[p]\ContinuedFloat
\centering
\begin{subfigure}[t]{0.94\textwidth}
\centering
\includegraphics[width=\linewidth]{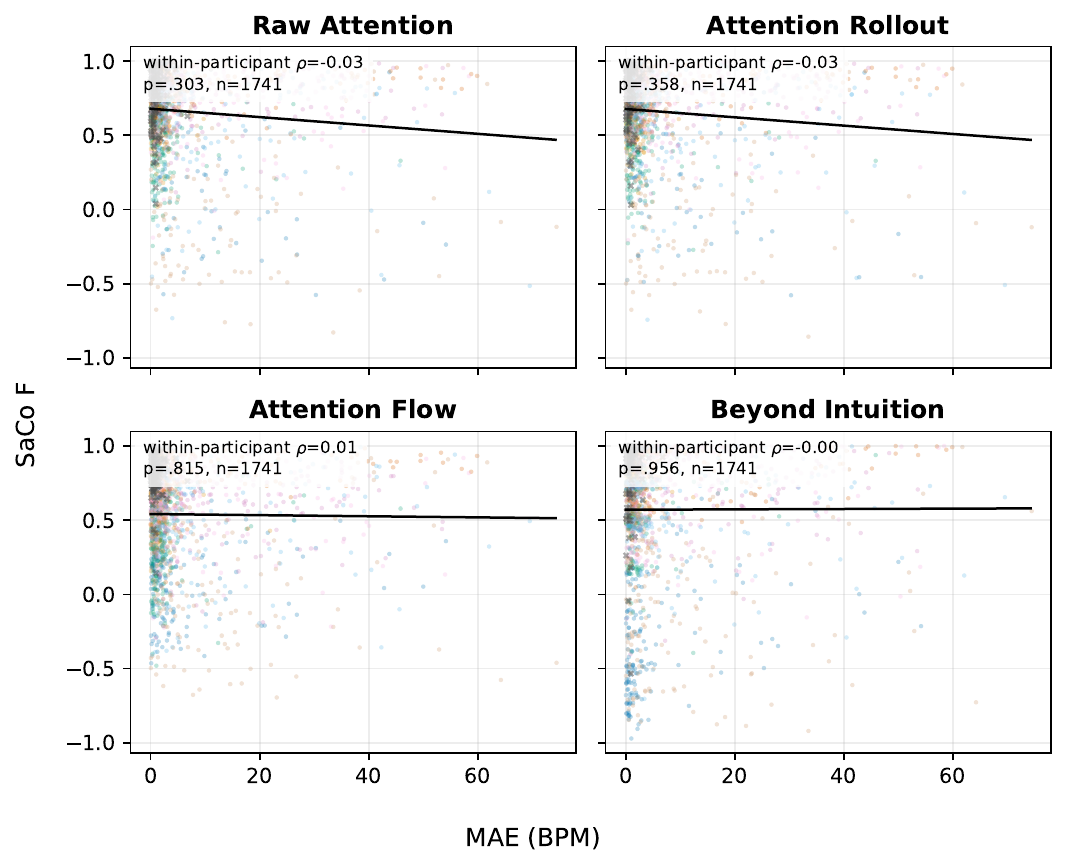}
\caption{MAE--SaCo.}
\end{subfigure}
\caption[]{Combined NCKU-rPPG and UBFC-rPPG descriptive scatter plots (continued).}
\end{figure}
\clearpage
\begin{figure}[p]\ContinuedFloat
\centering
\begin{subfigure}[t]{0.94\textwidth}
\centering
\includegraphics[width=\linewidth]{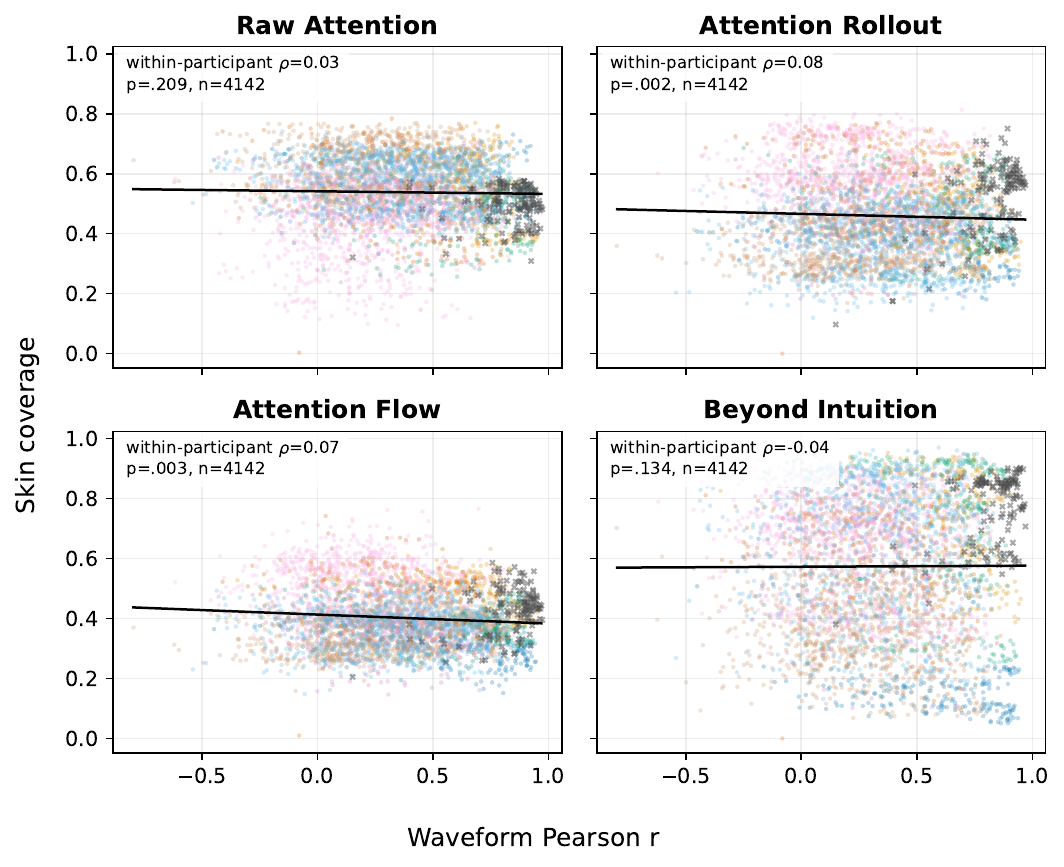}
\caption{Waveform Pearson--skin coverage.}
\end{subfigure}
\caption[]{Combined NCKU-rPPG and UBFC-rPPG descriptive scatter plots (continued).}
\end{figure}
\clearpage
\begin{figure}[p]\ContinuedFloat
\centering
\begin{subfigure}[t]{0.94\textwidth}
\centering
\includegraphics[width=\linewidth]{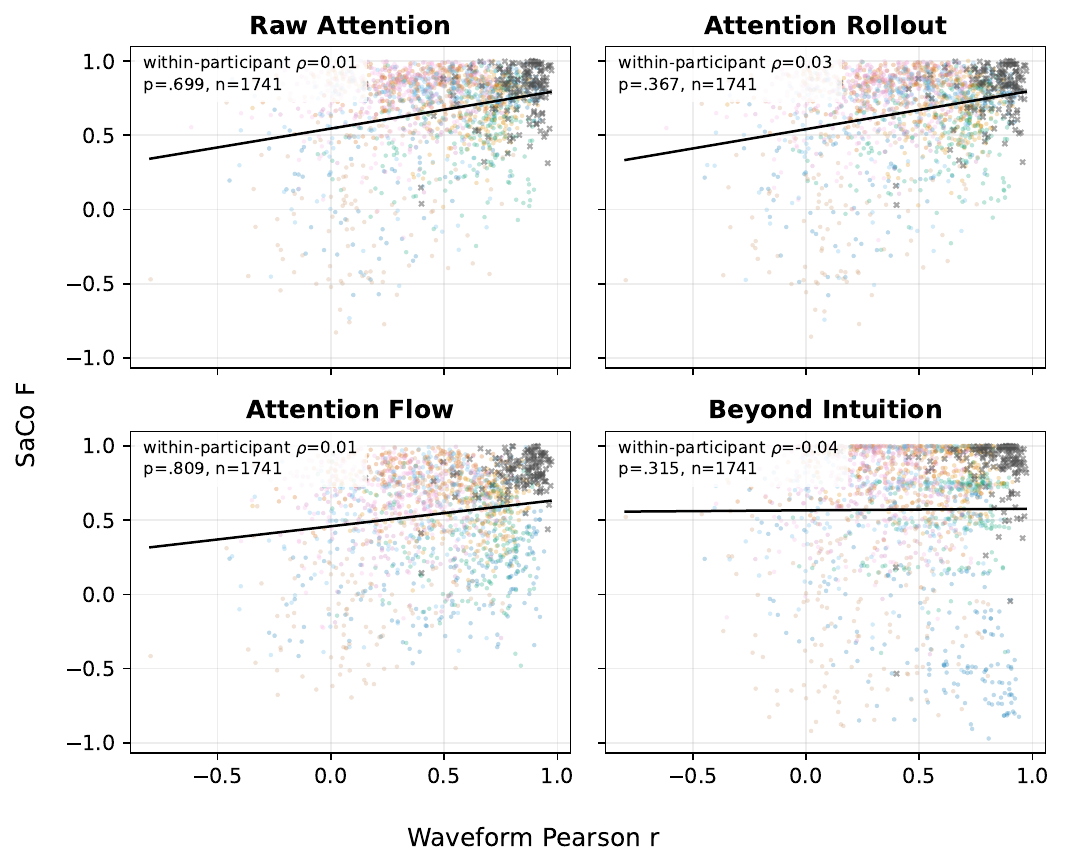}
\caption{Waveform Pearson--SaCo.}
\end{subfigure}
\caption[]{Combined NCKU-rPPG and UBFC-rPPG descriptive scatter plots (continued).}
\end{figure}
\clearpage
\begin{figure}[p]\ContinuedFloat
\centering
\begin{subfigure}[t]{0.94\textwidth}
\centering
\includegraphics[width=\linewidth]{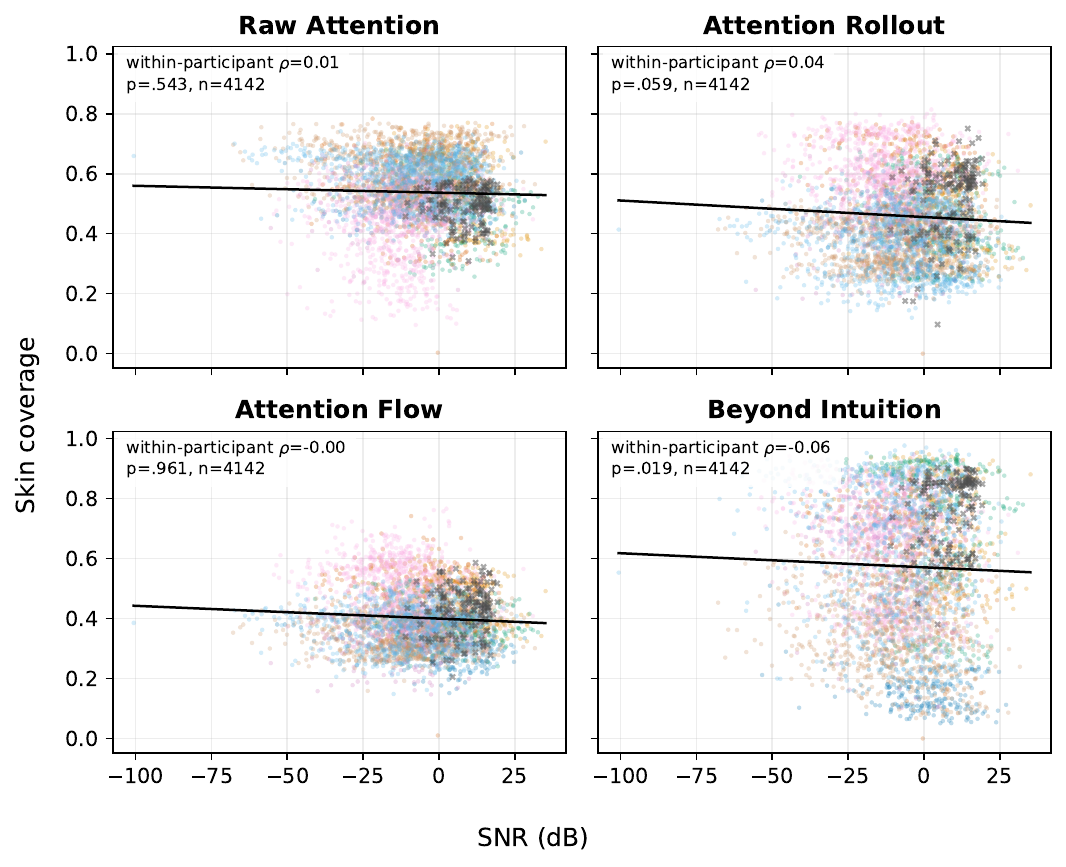}
\caption{SNR--skin coverage.}
\end{subfigure}
\caption[]{Combined NCKU-rPPG and UBFC-rPPG scatter plots (continued).}
\end{figure}
\clearpage
\begin{figure}[p]\ContinuedFloat
\centering
\begin{subfigure}[t]{0.94\textwidth}
\centering
\includegraphics[width=\linewidth]{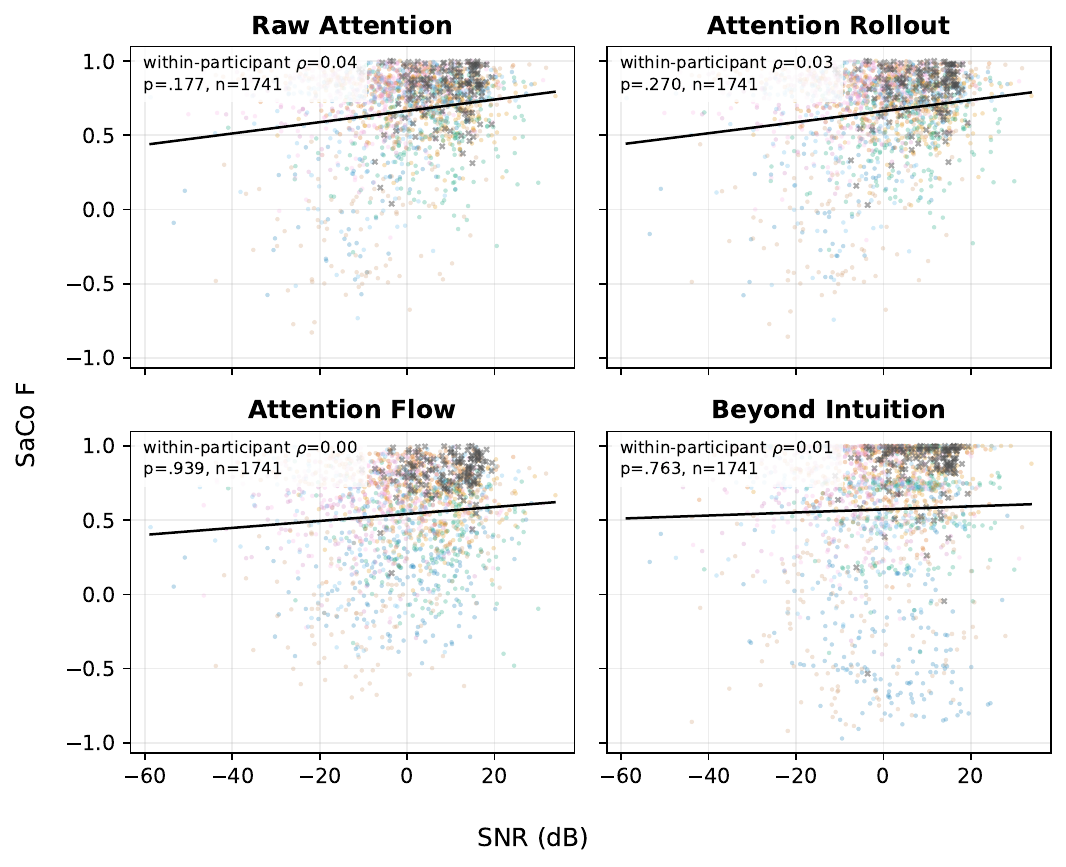}
\caption{SNR--SaCo.}
\end{subfigure}
\caption[]{Combined NCKU-rPPG and UBFC-rPPG scatter plots (continued).}
\end{figure}
\clearpage
\begin{figure}[p]\ContinuedFloat
\centering
\begin{subfigure}[t]{0.94\textwidth}
\centering
\includegraphics[width=\linewidth]{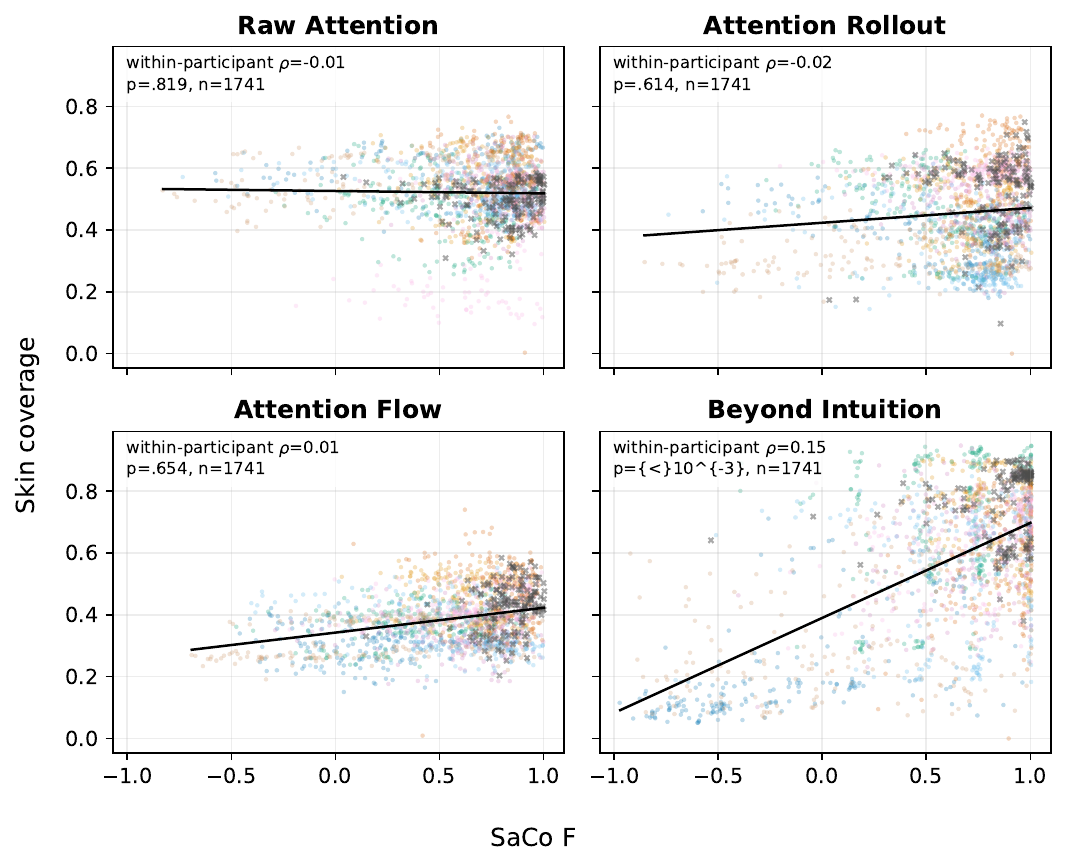}
\caption{SaCo--skin coverage.}
\end{subfigure}
\caption[]{Combined NCKU-rPPG and UBFC-rPPG scatter plots (continued).}
\end{figure}
\clearpage

Table~\ref{tab:supp_xai_cross_method} gives the complete cross-method skin-coverage summary.
Each row reports the pooled pre-attention and refined-attention means for one scenario and attribution method, together with their difference.

{\footnotesize
\newcommand{\ourdatasetxaicrossmethodrows}{%
Static level~1 & Raw Attention & $0.345\pm0.022$ & $0.350\pm0.012$ & $0.004\pm0.019$ \\
Static level~1 & Attention Rollout & $0.374\pm0.027$ & $0.359\pm0.025$ & $-0.015\pm0.011$ \\
Static level~1 & Attention Flow & $0.343\pm0.027$ & $0.331\pm0.018$ & $-0.012\pm0.011$ \\
\midrule
Static level~3 & Raw Attention & $0.398\pm0.031$ & $0.416\pm0.033$ & $0.018\pm0.018$ \\
Static level~3 & Attention Rollout & $0.465\pm0.039$ & $0.465\pm0.038$ & $0.0004\pm0.0068$ \\
Static level~3 & Attention Flow & $0.474\pm0.038$ & $0.470\pm0.029$ & $-0.005\pm0.014$ \\
\midrule
Static level~5 & Raw Attention & $0.359\pm0.030$ & $0.500\pm0.020$ & $0.141\pm0.019$ \\
Static level~5 & Attention Rollout & $0.487\pm0.032$ & $0.461\pm0.032$ & $-0.0256\pm0.0080$ \\
Static level~5 & Attention Flow & $0.478\pm0.027$ & $0.431\pm0.022$ & $-0.047\pm0.010$ \\
\midrule
Speak & Raw Attention & $0.397\pm0.015$ & $0.483\pm0.033$ & $0.086\pm0.026$ \\
Speak & Attention Rollout & $0.576\pm0.035$ & $0.557\pm0.033$ & $-0.019\pm0.019$ \\
Speak & Attention Flow & $0.587\pm0.033$ & $0.521\pm0.029$ & $-0.066\pm0.015$ \\
\midrule
Rotate & Raw Attention & $0.402\pm0.018$ & $0.469\pm0.013$ & $0.067\pm0.011$ \\
Rotate & Attention Rollout & $0.473\pm0.024$ & $0.391\pm0.016$ & $-0.0818\pm0.0094$ \\
Rotate & Attention Flow & $0.410\pm0.023$ & $0.348\pm0.012$ & $-0.062\pm0.013$ \\
\midrule
Bike level~1 & Raw Attention & $0.391\pm0.016$ & $0.496\pm0.023$ & $0.105\pm0.015$ \\
Bike level~1 & Attention Rollout & $0.443\pm0.030$ & $0.324\pm0.021$ & $-0.118\pm0.019$ \\
Bike level~1 & Attention Flow & $0.358\pm0.026$ & $0.295\pm0.020$ & $-0.062\pm0.014$ \\
\midrule
Bike level~3 & Raw Attention & $0.375\pm0.015$ & $0.411\pm0.023$ & $0.036\pm0.012$ \\
Bike level~3 & Attention Rollout & $0.622\pm0.024$ & $0.619\pm0.027$ & $-0.003\pm0.019$ \\
Bike level~3 & Attention Flow & $0.602\pm0.028$ & $0.563\pm0.028$ & $-0.039\pm0.017$ \\
\midrule
Bike level~5 & Raw Attention & $0.409\pm0.023$ & $0.529\pm0.021$ & $0.1195\pm0.0087$ \\
Bike level~5 & Attention Rollout & $0.467\pm0.027$ & $0.339\pm0.027$ & $-0.128\pm0.013$ \\
Bike level~5 & Attention Flow & $0.421\pm0.029$ & $0.351\pm0.021$ & $-0.070\pm0.012$ \\
}

\setlength{\tabcolsep}{3.5pt}
\renewcommand{\arraystretch}{1.08}
\begin{longtable}{>{\raggedright\arraybackslash}p{2.6cm} >{\raggedright\arraybackslash}p{3.1cm} rrr}
\caption{Mean skin coverage pooled over all held-out clips and TPT$_1$--TPT$_3$ for the three attention-only attribution methods, as estimate $\pm$ its standard error across the 12 test participants of the scenario.
The difference is between the two means of the same clips and therefore carries its own, smaller standard error.}
\label{tab:supp_xai_cross_method} \\
\toprule
Scenario & Method & Pre & Refined & Difference \\
\midrule
\endfirsthead
\multicolumn{5}{c}{\tablename\ \thetable\ (continued)} \\
\toprule
Scenario & Method & Pre & Refined & Difference \\
\midrule
\endhead
\bottomrule
\endfoot
\ourdatasetxaicrossmethodrows
\end{longtable}
}

Read across all nine conditions, Table~\ref{tab:supp_xai_cross_method} qualifies the contrast drawn in Section~\ref{sec:static3_xai_results}.
Refinement raises the skin coverage of raw attention in every one of the eight NCKU-rPPG scenarios, from $+0.004$ on Static level~1 to $+0.141$ on Static level~5, and lowers that of attention flow in every one, from $-0.005$ to $-0.070$;
attention rollout falls in seven of the eight, and the three directions are the same on UBFC-rPPG, where the changes are $+0.224$, $-0.117$ and $-0.167$.
What separates the two evaluations is therefore magnitude and not direction: every one of the three changes is smaller than twice its standard error only in Static levels~1 and~3, and in Bike level~5 they reach about 14 and 10 times it for raw attention and attention rollout, so the refinement step tracks how difficult the condition is rather than which dataset it belongs to.

Table~\ref{tab:supp_xai_tpt_skin} gives the TPT-level mean and standard deviation values.
Each row covers one scenario, attribution method, and attention type, with one column per Temporal Periodic Transformer level.

{\scriptsize
\newcommand{\ourdatasetxaitptrows}{%
Static level~1 & Raw Attention & Pre-Attention & $0.385 \pm 0.076$ & $0.294 \pm 0.089$ & $0.356 \pm 0.095$ \\
Static level~1 & Raw Attention & Refined Attention & $0.338 \pm 0.067$ & $0.163 \pm 0.045$ & $0.549 \pm 0.083$ \\
Static level~1 & Attention Rollout & Pre-Attention & $0.314 \pm 0.086$ & $0.296 \pm 0.091$ & $0.51 \pm 0.12$ \\
Static level~1 & Attention Rollout & Refined Attention & $0.348 \pm 0.094$ & $0.363 \pm 0.086$ & $0.367 \pm 0.081$ \\
Static level~1 & Attention Flow & Pre-Attention & $0.34 \pm 0.11$ & $0.336 \pm 0.083$ & $0.352 \pm 0.083$ \\
Static level~1 & Attention Flow & Refined Attention & $0.345 \pm 0.088$ & $0.330 \pm 0.051$ & $0.317 \pm 0.057$ \\
\midrule
Static level~3 & Raw Attention & Pre-Attention & $0.37 \pm 0.12$ & $0.368 \pm 0.095$ & $0.46 \pm 0.12$ \\
Static level~3 & Raw Attention & Refined Attention & $0.40 \pm 0.15$ & $0.29 \pm 0.10$ & $0.57 \pm 0.12$ \\
Static level~3 & Attention Rollout & Pre-Attention & $0.39 \pm 0.14$ & $0.43 \pm 0.14$ & $0.58 \pm 0.13$ \\
Static level~3 & Attention Rollout & Refined Attention & $0.44 \pm 0.14$ & $0.47 \pm 0.13$ & $0.49 \pm 0.12$ \\
Static level~3 & Attention Flow & Pre-Attention & $0.46 \pm 0.14$ & $0.46 \pm 0.13$ & $0.50 \pm 0.13$ \\
Static level~3 & Attention Flow & Refined Attention & $0.48 \pm 0.13$ & $0.464 \pm 0.097$ & $0.463 \pm 0.071$ \\
\midrule
Static level~5 & Raw Attention & Pre-Attention & $0.37 \pm 0.14$ & $0.31 \pm 0.10$ & $0.398 \pm 0.073$ \\
Static level~5 & Raw Attention & Refined Attention & $0.55 \pm 0.10$ & $0.46 \pm 0.10$ & $0.485 \pm 0.099$ \\
Static level~5 & Attention Rollout & Pre-Attention & $0.41 \pm 0.10$ & $0.48 \pm 0.11$ & $0.57 \pm 0.13$ \\
Static level~5 & Attention Rollout & Refined Attention & $0.446 \pm 0.095$ & $0.47 \pm 0.11$ & $0.47 \pm 0.12$ \\
Static level~5 & Attention Flow & Pre-Attention & $0.487 \pm 0.084$ & $0.470 \pm 0.098$ & $0.48 \pm 0.10$ \\
Static level~5 & Attention Flow & Refined Attention & $0.493 \pm 0.086$ & $0.415 \pm 0.084$ & $0.384 \pm 0.061$ \\
\midrule
Speak & Raw Attention & Pre-Attention & $0.474 \pm 0.088$ & $0.364 \pm 0.064$ & $0.354 \pm 0.067$ \\
Speak & Raw Attention & Refined Attention & $0.40 \pm 0.13$ & $0.50 \pm 0.12$ & $0.55 \pm 0.13$ \\
Speak & Attention Rollout & Pre-Attention & $0.56 \pm 0.14$ & $0.58 \pm 0.13$ & $0.59 \pm 0.12$ \\
Speak & Attention Rollout & Refined Attention & $0.54 \pm 0.12$ & $0.56 \pm 0.12$ & $0.56 \pm 0.12$ \\
Speak & Attention Flow & Pre-Attention & $0.61 \pm 0.13$ & $0.59 \pm 0.12$ & $0.57 \pm 0.11$ \\
Speak & Attention Flow & Refined Attention & $0.57 \pm 0.12$ & $0.52 \pm 0.11$ & $0.468 \pm 0.090$ \\
\midrule
Rotate & Raw Attention & Pre-Attention & $0.440 \pm 0.078$ & $0.40 \pm 0.15$ & $0.366 \pm 0.070$ \\
Rotate & Raw Attention & Refined Attention & $0.320 \pm 0.073$ & $0.57 \pm 0.11$ & $0.518 \pm 0.091$ \\
Rotate & Attention Rollout & Pre-Attention & $0.35 \pm 0.10$ & $0.46 \pm 0.13$ & $0.60 \pm 0.12$ \\
Rotate & Attention Rollout & Refined Attention & $0.321 \pm 0.071$ & $0.42 \pm 0.10$ & $0.44 \pm 0.10$ \\
Rotate & Attention Flow & Pre-Attention & $0.326 \pm 0.099$ & $0.41 \pm 0.11$ & $0.49 \pm 0.11$ \\
Rotate & Attention Flow & Refined Attention & $0.336 \pm 0.072$ & $0.373 \pm 0.079$ & $0.337 \pm 0.070$ \\
\midrule
Bike level~1 & Raw Attention & Pre-Attention & $0.346 \pm 0.069$ & $0.467 \pm 0.089$ & $0.360 \pm 0.049$ \\
Bike level~1 & Raw Attention & Refined Attention & $0.37 \pm 0.13$ & $0.51 \pm 0.11$ & $0.609 \pm 0.083$ \\
Bike level~1 & Attention Rollout & Pre-Attention & $0.29 \pm 0.14$ & $0.45 \pm 0.11$ & $0.593 \pm 0.095$ \\
Bike level~1 & Attention Rollout & Refined Attention & $0.265 \pm 0.082$ & $0.347 \pm 0.075$ & $0.361 \pm 0.074$ \\
Bike level~1 & Attention Flow & Pre-Attention & $0.24 \pm 0.11$ & $0.364 \pm 0.096$ & $0.468 \pm 0.093$ \\
Bike level~1 & Attention Flow & Refined Attention & $0.225 \pm 0.097$ & $0.316 \pm 0.079$ & $0.345 \pm 0.061$ \\
\midrule
Bike level~3 & Raw Attention & Pre-Attention & $0.496 \pm 0.074$ & $0.297 \pm 0.046$ & $0.331 \pm 0.065$ \\
Bike level~3 & Raw Attention & Refined Attention & $0.41 \pm 0.14$ & $0.39 \pm 0.11$ & $0.44 \pm 0.12$ \\
Bike level~3 & Attention Rollout & Pre-Attention & $0.66 \pm 0.10$ & $0.588 \pm 0.063$ & $0.62 \pm 0.11$ \\
Bike level~3 & Attention Rollout & Refined Attention & $0.63 \pm 0.11$ & $0.615 \pm 0.092$ & $0.616 \pm 0.090$ \\
Bike level~3 & Attention Flow & Pre-Attention & $0.63 \pm 0.13$ & $0.590 \pm 0.088$ & $0.582 \pm 0.079$ \\
Bike level~3 & Attention Flow & Refined Attention & $0.60 \pm 0.13$ & $0.588 \pm 0.091$ & $0.501 \pm 0.092$ \\
\midrule
Bike level~5 & Raw Attention & Pre-Attention & $0.345 \pm 0.061$ & $0.417 \pm 0.087$ & $0.47 \pm 0.12$ \\
Bike level~5 & Raw Attention & Refined Attention & $0.487 \pm 0.089$ & $0.514 \pm 0.088$ & $0.586 \pm 0.074$ \\
Bike level~5 & Attention Rollout & Pre-Attention & $0.31 \pm 0.13$ & $0.48 \pm 0.11$ & $0.611 \pm 0.097$ \\
Bike level~5 & Attention Rollout & Refined Attention & $0.28 \pm 0.11$ & $0.36 \pm 0.11$ & $0.38 \pm 0.11$ \\
Bike level~5 & Attention Flow & Pre-Attention & $0.32 \pm 0.14$ & $0.42 \pm 0.11$ & $0.52 \pm 0.11$ \\
Bike level~5 & Attention Flow & Refined Attention & $0.32 \pm 0.10$ & $0.360 \pm 0.076$ & $0.373 \pm 0.072$ \\
}

\setlength{\tabcolsep}{2.5pt}
\renewcommand{\arraystretch}{1.06}
\begin{longtable}{>{\raggedright\arraybackslash}p{2.0cm} >{\raggedright\arraybackslash}p{2.5cm} >{\raggedright\arraybackslash}p{2.3cm} ccc}
\caption{Clip-level skin coverage as mean $\pm$ standard deviation by scenario, attribution method, attention type, and Temporal Periodic Transformer (TPT) level.}
\label{tab:supp_xai_tpt_skin} \\
\toprule
Scenario & Method & Type & TPT$_1$ & TPT$_2$ & TPT$_3$ \\
\midrule
\endfirsthead
\multicolumn{6}{c}{\tablename\ \thetable\ (continued)} \\
\toprule
Scenario & Method & Type & TPT$_1$ & TPT$_2$ & TPT$_3$ \\
\midrule
\endhead
\bottomrule
\endfoot
\ourdatasetxaitptrows
\end{longtable}
}

Table~\ref{tab:supp_xai_tpt_skin} shows that the two families of method acquire their depth dependence at opposite stages.
Pre-attention skin coverage rises with Temporal Periodic Transformer level for attention rollout in seven of the eight scenarios and for attention flow in five, by as much as 0.30 from TPT$_1$ to TPT$_3$ on Bike level~1, and refinement then compresses that gradient to at most 0.12.
Raw attention behaves the other way: its pre-attention coverage carries no consistent depth trend, rising in four scenarios and falling in four, and a rising one appears after refinement in seven of the eight, reaching $+0.24$ from TPT$_1$ to TPT$_3$ on Bike level~1 against $+0.01$ before it.
Pooling the three levels, as Table~\ref{tab:static3_xai_coverage} and Table~\ref{tab:supp_xai_cross_method} do, therefore averages over a gradient that runs in opposite directions for raw attention and for the two propagation-based methods.

\newcommand{\facexxaisummary}[5]{%
\begin{figure}[p]
\centering
#3
\caption{#1 attribution visualisation.
The published representative face appears only in the top panels labelled \emph{Face}, \emph{Skin Mask}, and \emph{Skin Region}.
The attribution maps below those panels, the average face, and the displayed summary statistics use the complete #1 test set rather than that representative alone.}
\label{fig:supp_xai_#2_visualization}
\end{figure}
\begin{figure}[p]
\centering
\begin{subfigure}[t]{0.49\textwidth}
\centering
#4
\caption{SaCo.}
\end{subfigure}
\hfill
\begin{subfigure}[t]{0.49\textwidth}
\centering
#5
\caption{Skin coverage.}
\end{subfigure}
\caption{#1 clip-level distribution summaries for raw attention, attention rollout, attention flow, and Beyond Intuition.
Each SaCo distribution contains clip-level faithfulness values, and each skin-coverage distribution contains clip-level values averaged across TPT$_1$--TPT$_3$ before plotting.}
\label{fig:supp_xai_#2_distributions}
\end{figure}
}

\newcommand{\facexscattergroup}[7]{%
\begin{landscape}
\begin{figure}[p]
\centering
\begin{subfigure}[t]{0.295\linewidth}
\centering
#3
\caption{MAE--skin coverage.}
\end{subfigure}
\hfill
\begin{subfigure}[t]{0.295\linewidth}
\centering
#4
\caption{MAE--SaCo.}
\end{subfigure}
\hfill
\begin{subfigure}[t]{0.295\linewidth}
\centering
#5
\caption{Waveform Pearson--skin coverage.}
\end{subfigure}
\par\medskip
\hspace*{\fill}
\begin{subfigure}[t]{0.295\linewidth}
\centering
#6
\caption{Waveform Pearson--SaCo.}
\end{subfigure}
\hfill
\begin{subfigure}[t]{0.295\linewidth}
\centering
#7
\caption{SaCo--skin coverage.}
\end{subfigure}
\hspace*{\fill}
\caption{#1 descriptive scatter plots for raw attention, attention rollout, attention flow, and Beyond Intuition.
Each point is a clip, and colours distinguish held-out sessions.
The $\rho$ and $p$ values printed within the panels are descriptive Spearman summaries; the nominal $p$ values are not interpreted as significance tests of independent observations.}
\label{fig:supp_xai_#2_scatter}
\end{figure}
\clearpage
\end{landscape}
}

\clearpage
Static level~1 is the condition where the attributions are least consistent with the rest and refinement changes them least.
Figure~\ref{fig:supp_xai_static1_visualization} shows the attribution maps before and after refinement, which move the pooled skin coverage of raw attention by $+0.004\pm0.019$, of attention rollout by $-0.015\pm0.011$ and of attention flow by $-0.012\pm0.011$.
All three changes are smaller than twice their standard error, so this and Static level~3 are the only two NCKU-rPPG conditions in which refinement is not resolved for any attention-only method.
Figure~\ref{fig:supp_xai_static1_distributions} nonetheless separates the four methods sharply: Beyond Intuition is the only one with a median SaCo below zero, $-0.178$, and its median refined skin coverage of $0.180$ is the lowest of any condition, against 0.34--0.38 for the three attention-only methods.
Raw attention gives the heart-rate error against SaCo at $\rho=-0.19$ ($p=.013$) and the waveform correlation against skin coverage at $\rho=+0.14$ ($p=.046$), both in the direction a reliability indicator would take, while Beyond Intuition gives the waveform correlation against SaCo at $\rho=-0.22$ ($p=.019$), which runs opposite to it.

\facexxaisummary{Static level~1}{static1}%
{\includegraphics[width=0.72\textwidth]{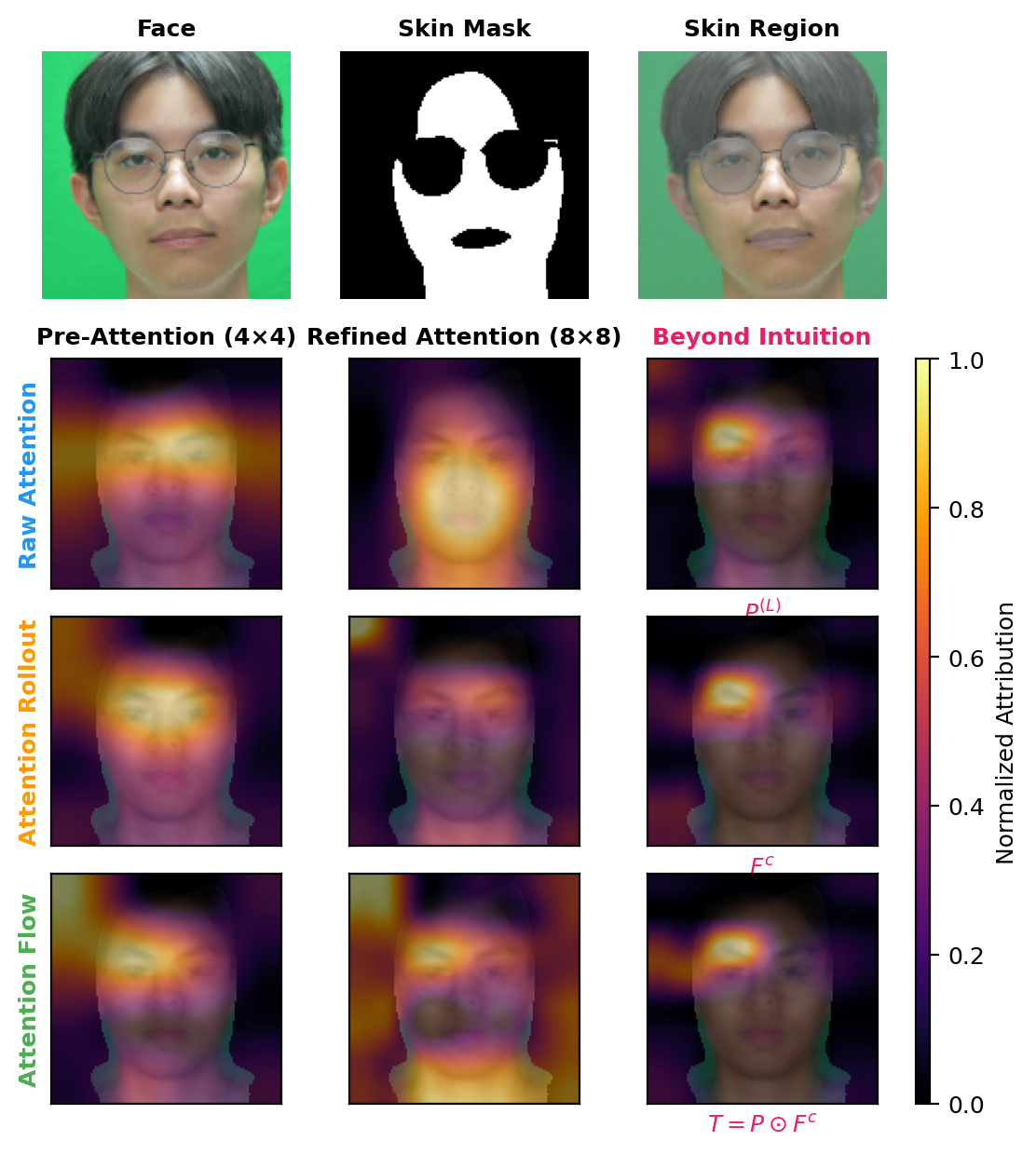}}%
{\includegraphics[width=\textwidth]{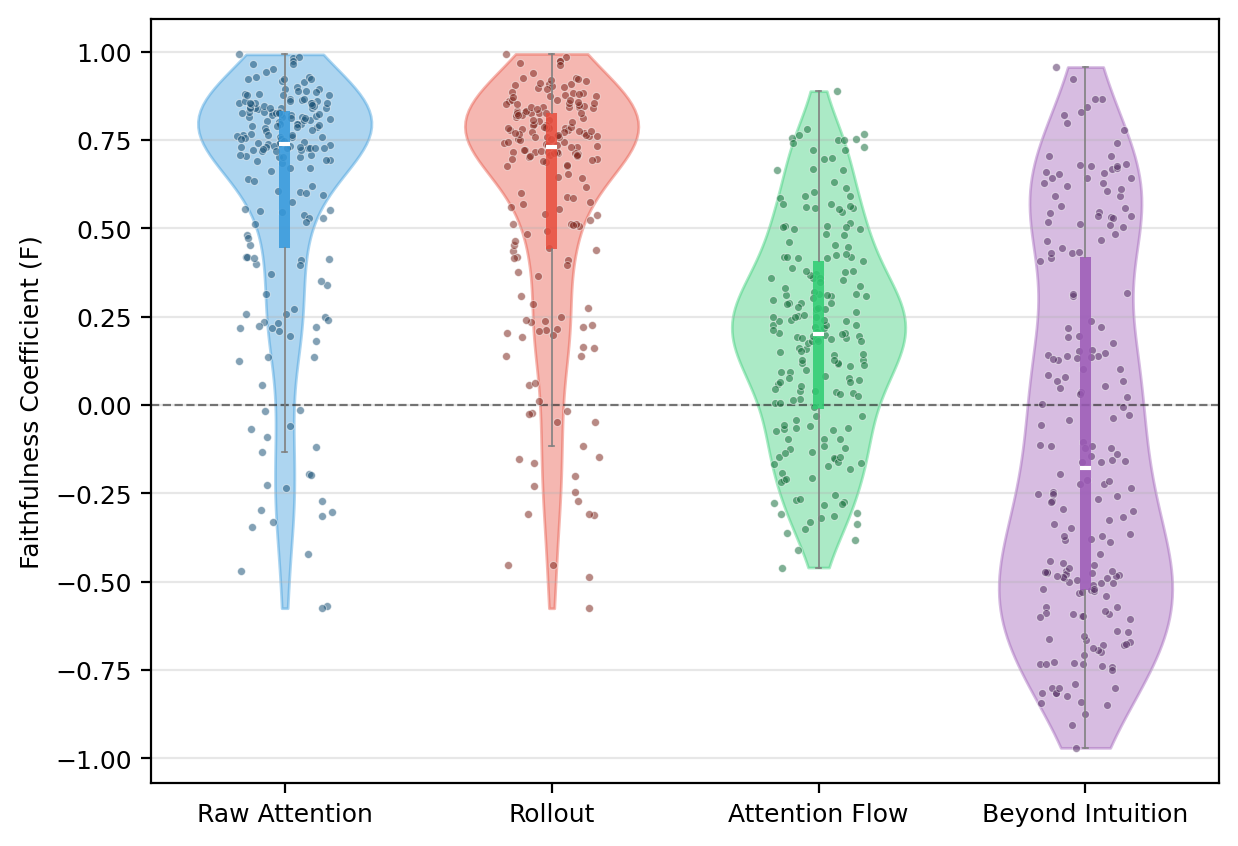}}%
{\includegraphics[width=\textwidth]{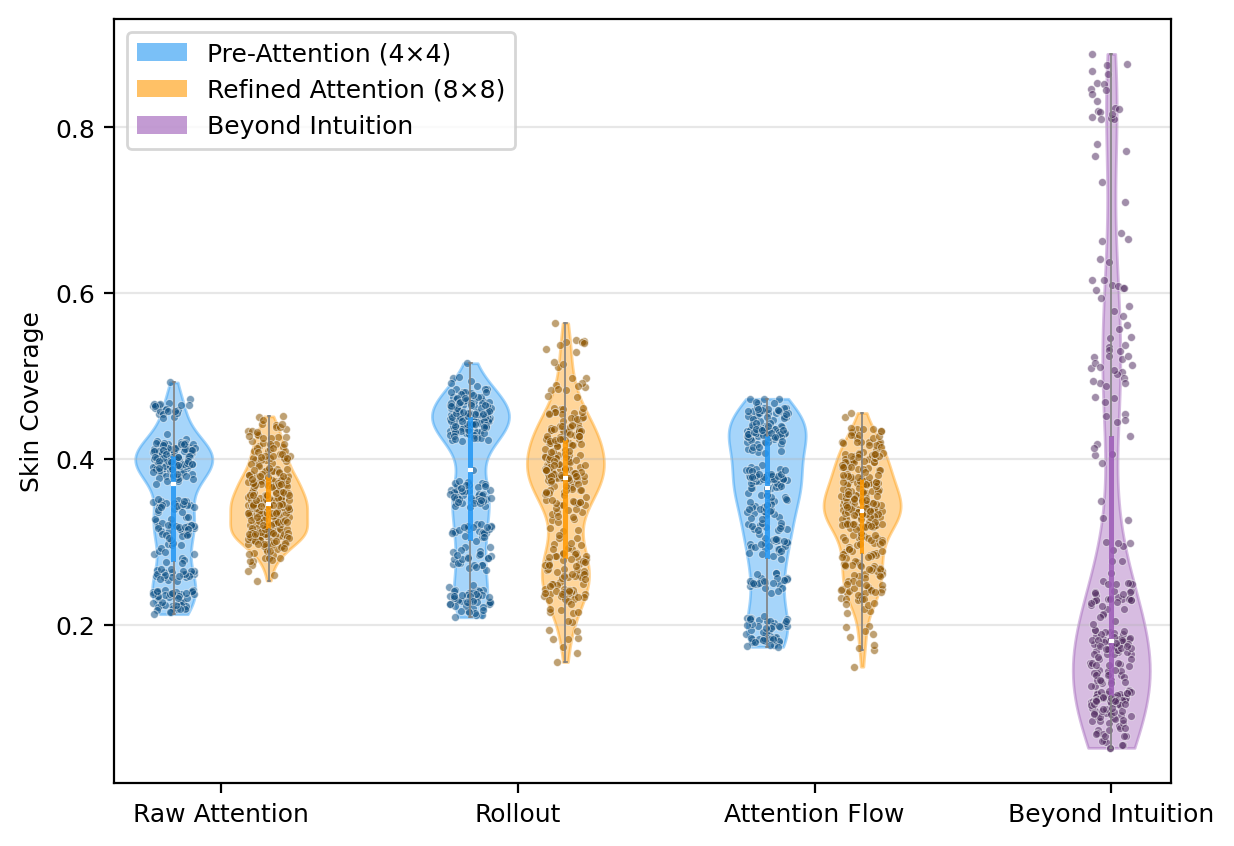}}
\FloatBarrier
\clearpage

Static level~5 is where refinement changes the attention-only methods most among the static conditions.
Figure~\ref{fig:supp_xai_static5_visualization} shows the maps that produce that change: raw attention gains $0.141\pm0.019$ of skin coverage between the two stages, more than seven times its standard error, while attention rollout loses $0.026\pm0.008$ and attention flow $0.047\pm0.010$.
Figure~\ref{fig:supp_xai_static5_distributions} places the median SaCo of the four methods between 0.416 for attention flow and 0.708 for Beyond Intuition, and their median refined skin coverage between 0.435 and 0.779 in the same order.

\facexxaisummary{Static level~5}{static5}%
{\includegraphics[width=0.72\textwidth]{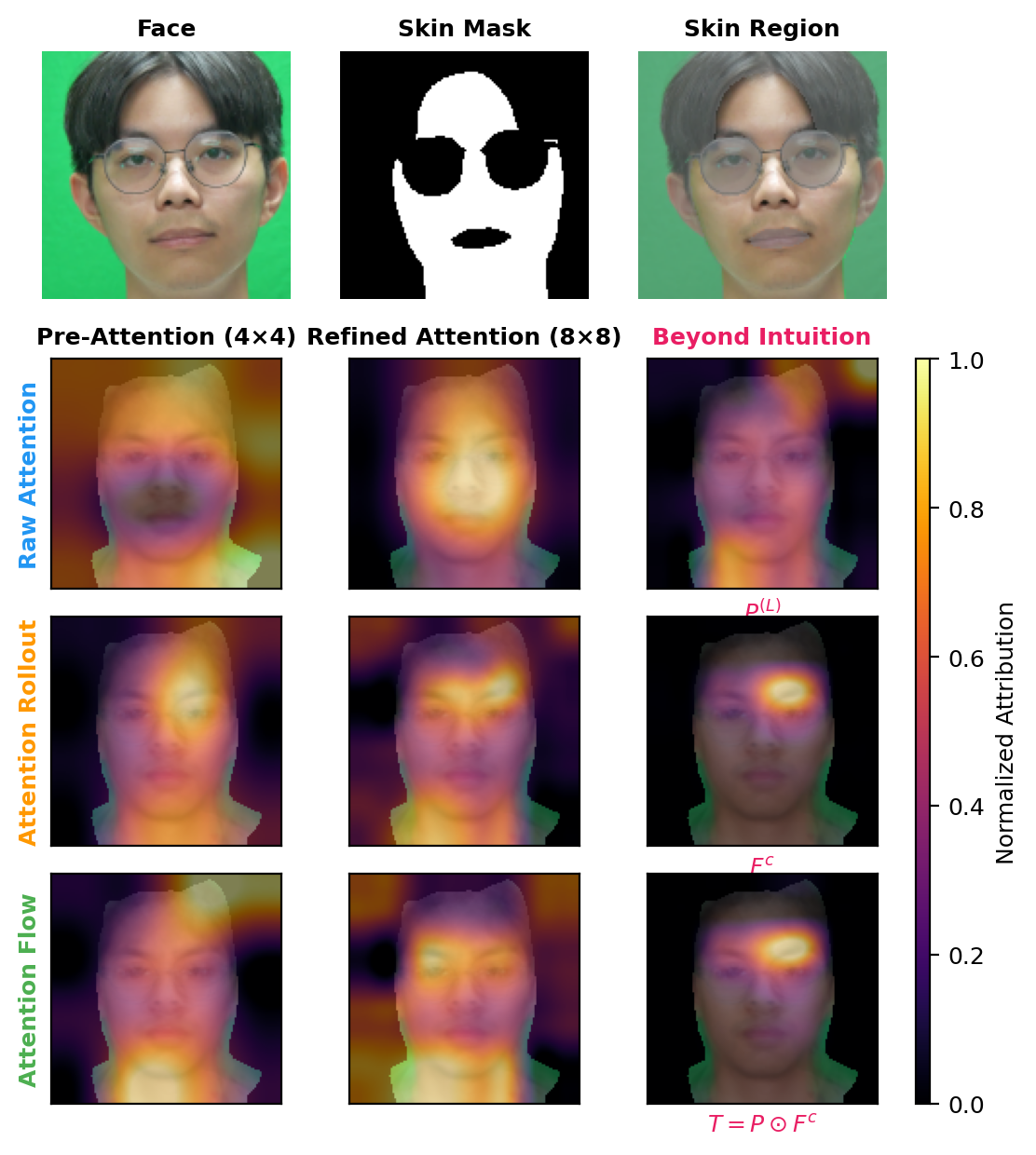}}%
{\includegraphics[width=\textwidth]{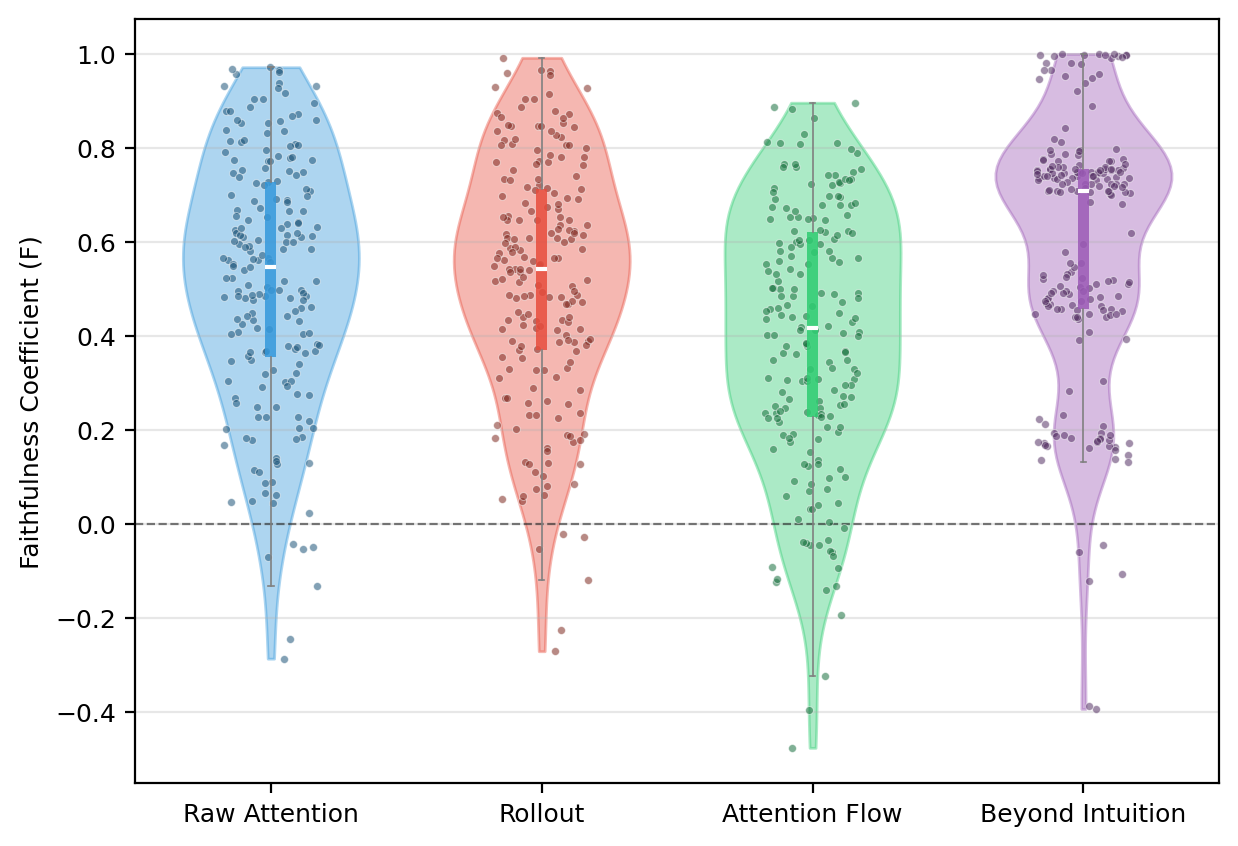}}%
{\includegraphics[width=\textwidth]{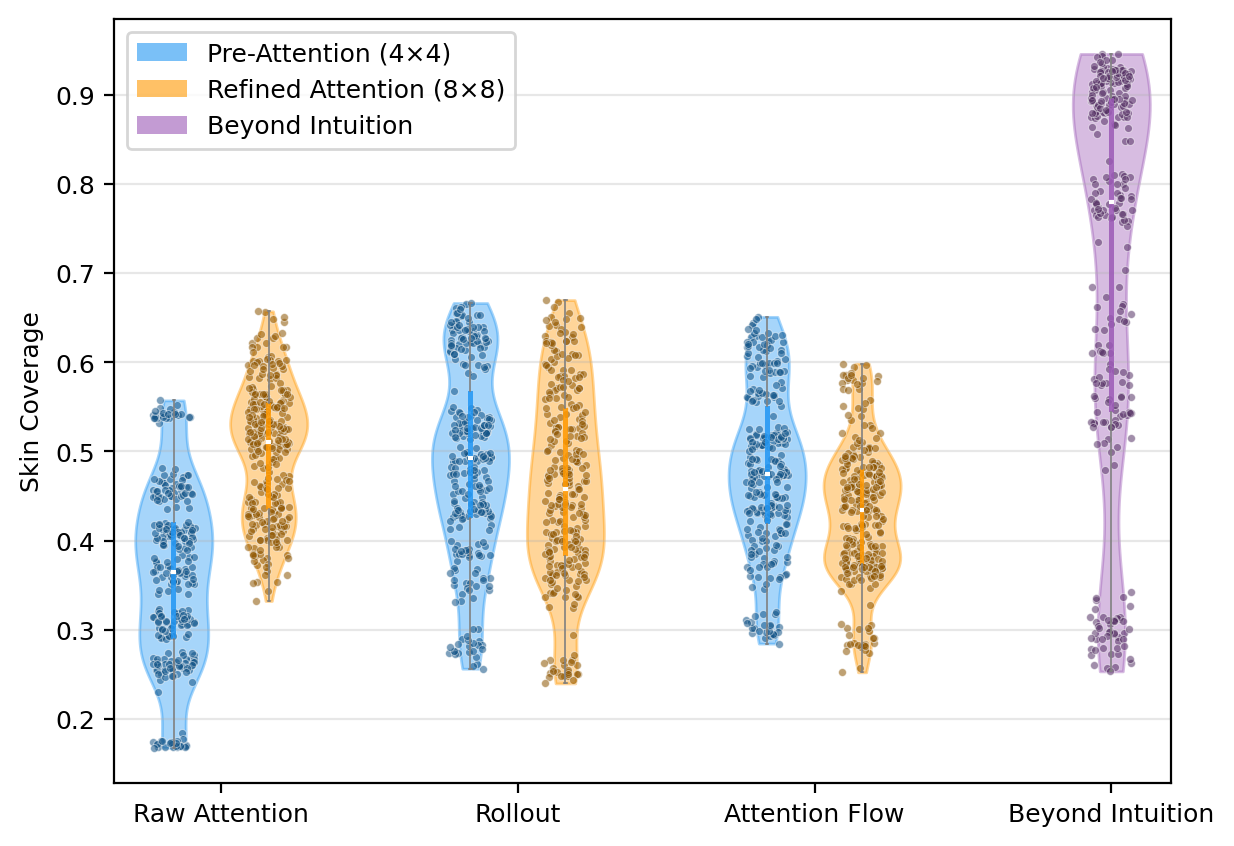}}
\FloatBarrier
\clearpage

Figure~\ref{fig:supp_xai_speak_visualization} shows the attribution maps for speaking, where refinement raises the skin coverage of raw attention by $0.086\pm0.026$ and lowers that of attention flow by $0.066\pm0.015$, while its effect on attention rollout, $-0.019\pm0.019$, is not resolved.
Figure~\ref{fig:supp_xai_speak_distributions} gives the highest median SaCo of any NCKU-rPPG condition, 0.824--0.863 across the four methods, even though the heart-rate error of $8.9\pm2.6$~beats per minute is the fifth worst of the eight in Table~\ref{tab:performance_results}.
That pairing is the clearest single illustration of the conclusion in Section~\ref{sec:scenario_reliability_results}, because a faithfulness score near its maximum accompanies a model that is not working well.

\facexxaisummary{Speak}{speak}%
{\includegraphics[width=0.72\textwidth]{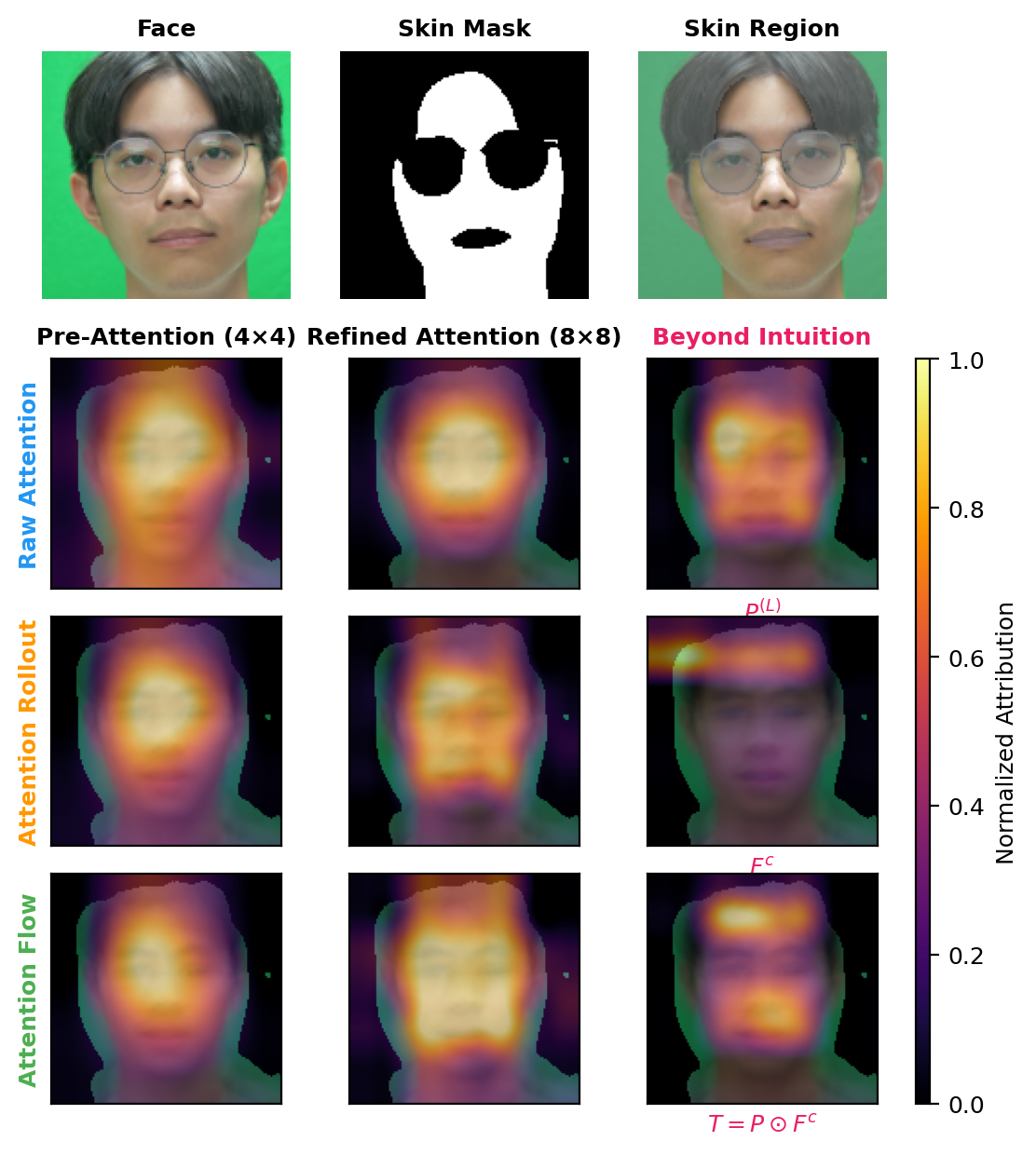}}%
{\includegraphics[width=\textwidth]{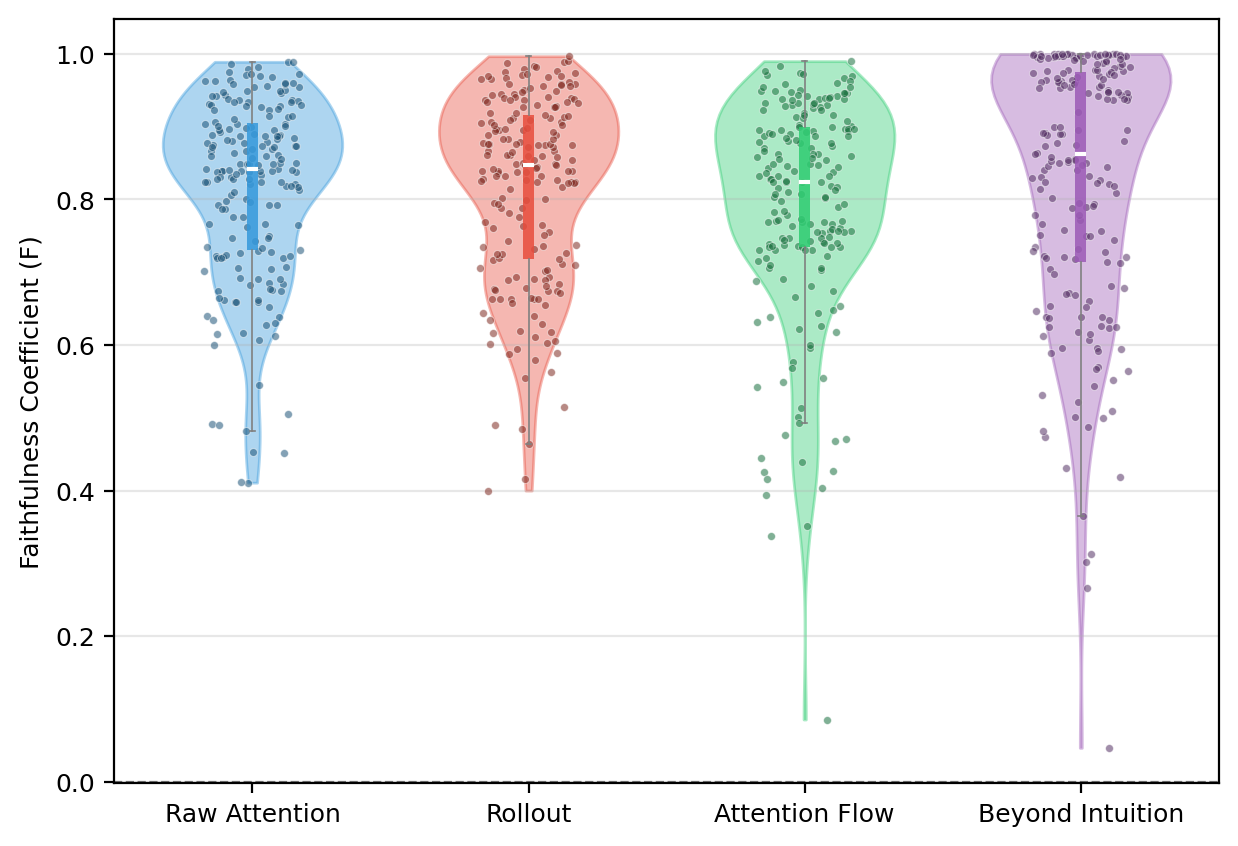}}%
{\includegraphics[width=\textwidth]{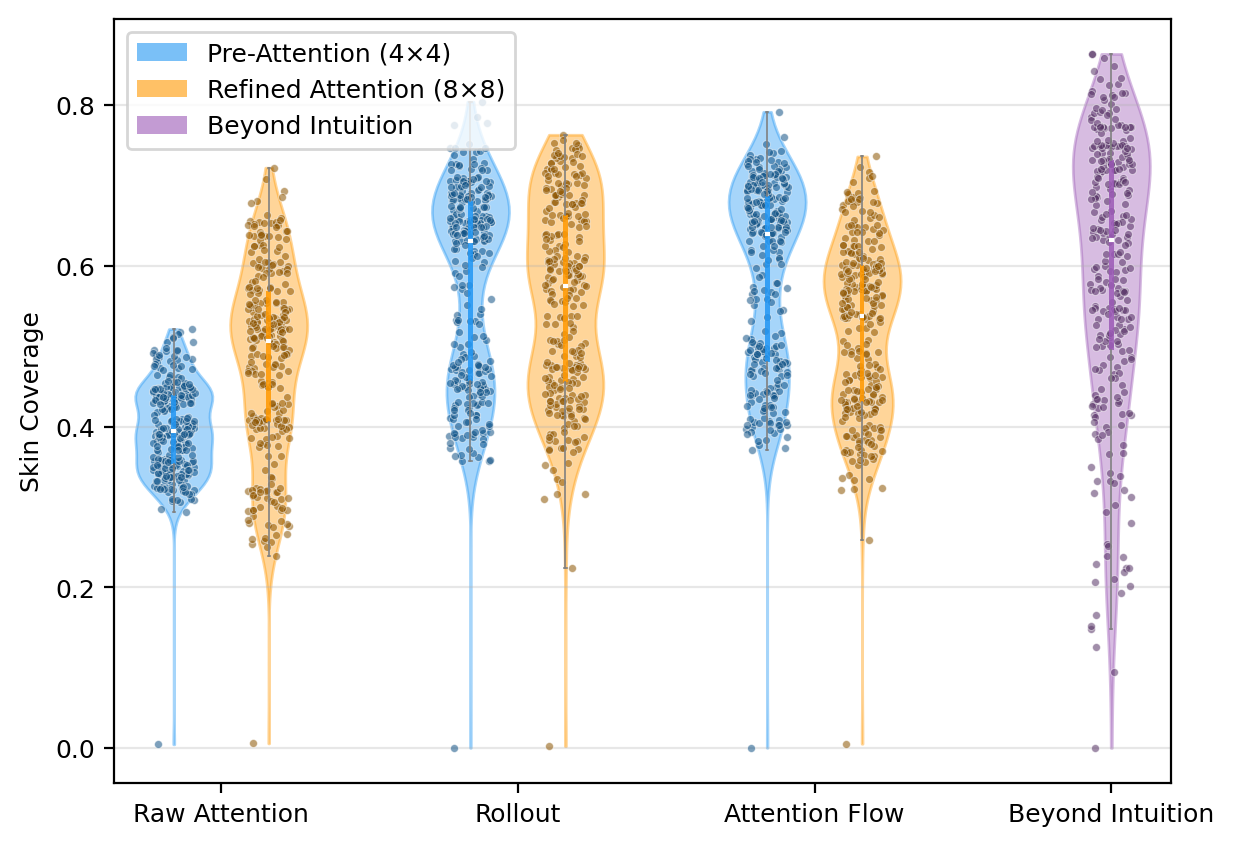}}
\FloatBarrier
\clearpage

Figure~\ref{fig:supp_xai_rotate_visualization} shows attribution maps that remain concentrated on the face despite the changing pose, with refinement raising the skin coverage of raw attention by $0.067\pm0.011$ and lowering that of attention rollout by $0.082\pm0.009$ and attention flow by $0.062\pm0.013$.
Figure~\ref{fig:supp_xai_rotate_distributions} gives median SaCo values of 0.652--0.819, close to those of the static conditions, while Table~\ref{tab:performance_results} places rotation second worst of the eight on SNR and fourth worst on heart-rate error.

\facexxaisummary{Rotate}{rotate}%
{\includegraphics[width=0.72\textwidth]{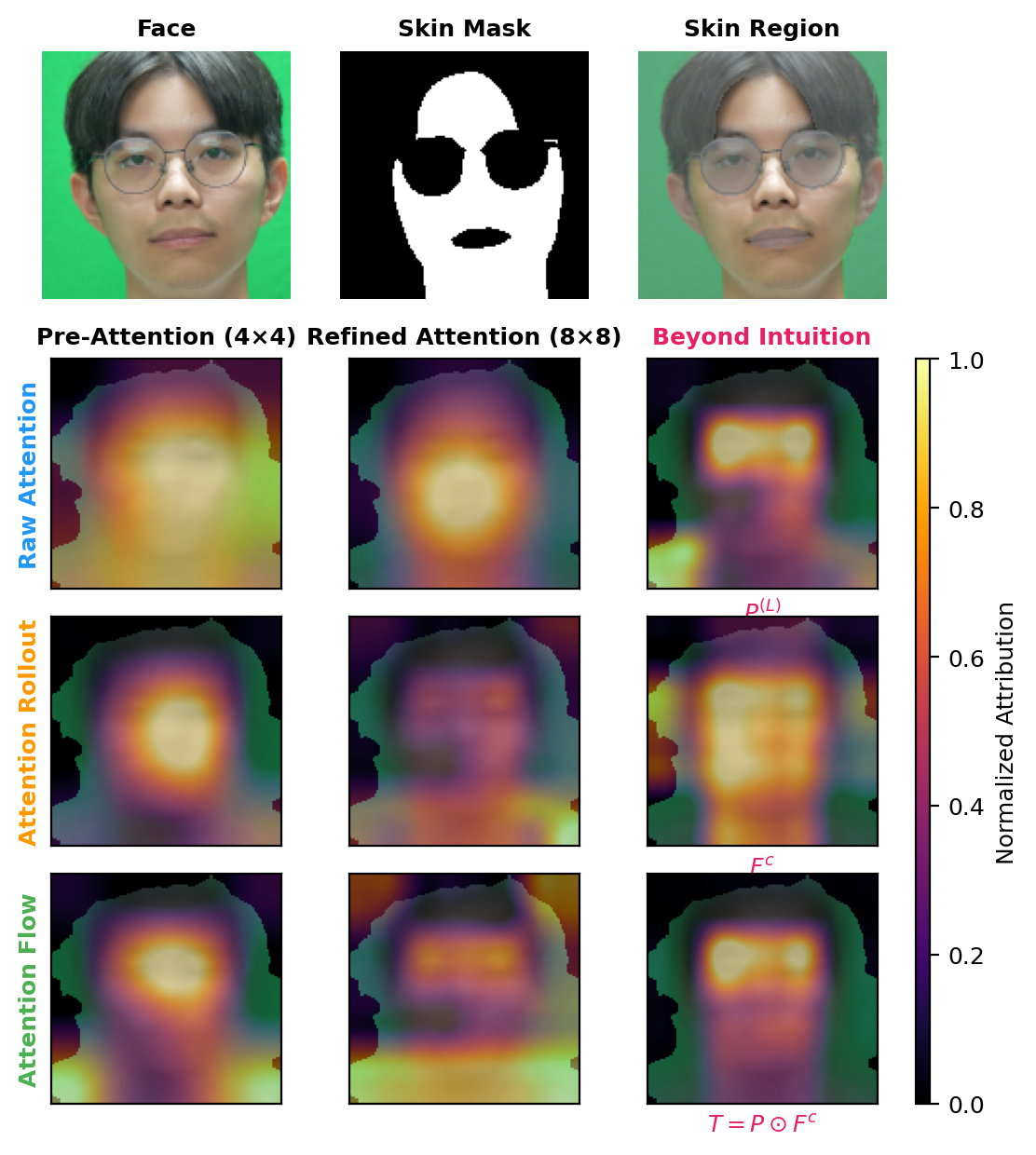}}%
{\includegraphics[width=\textwidth]{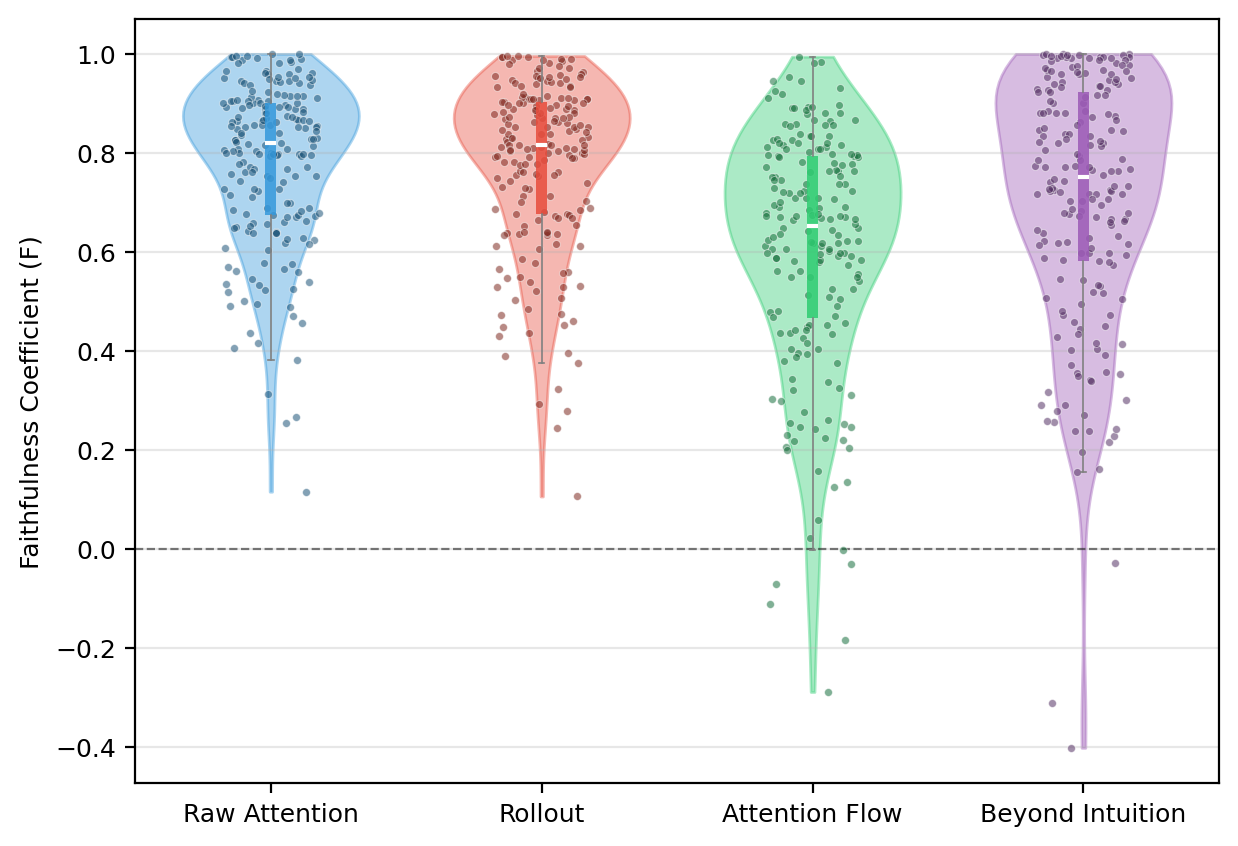}}%
{\includegraphics[width=\textwidth]{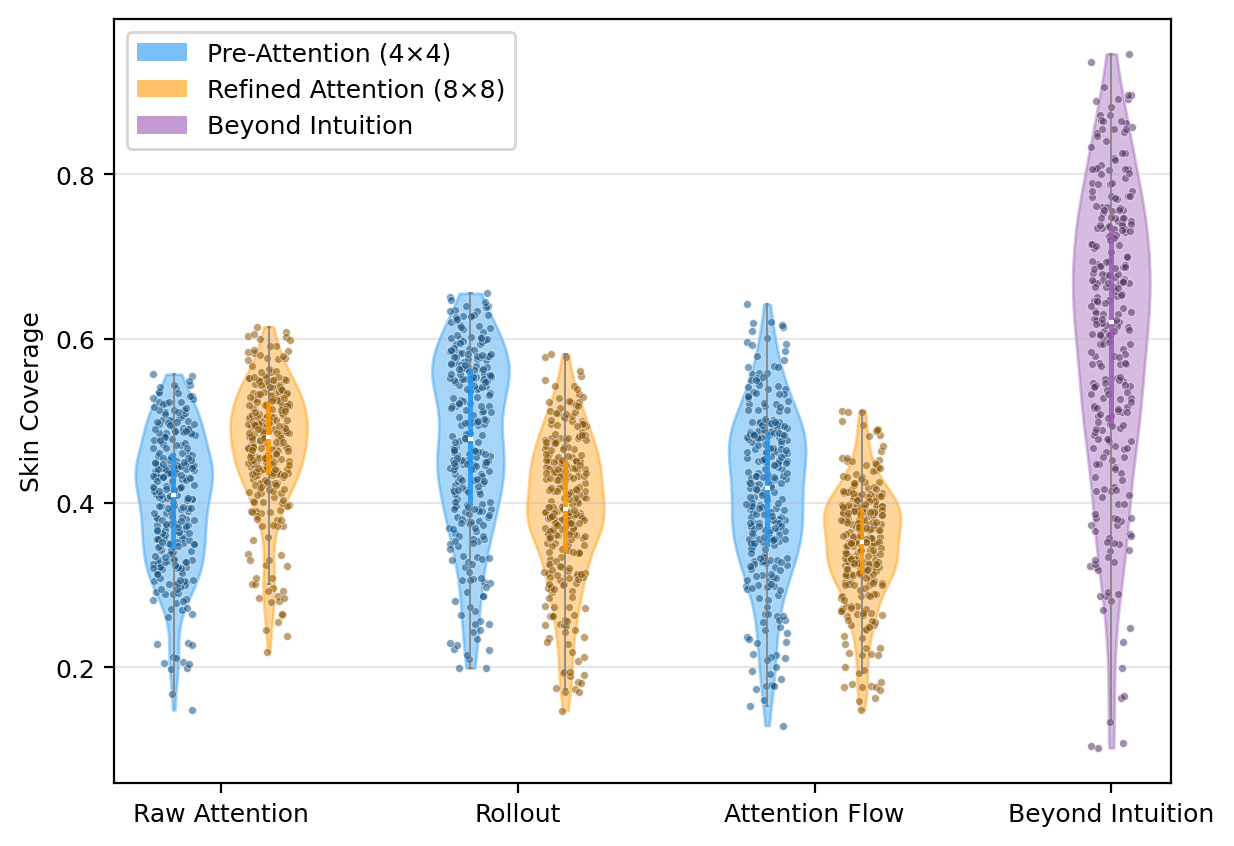}}
\FloatBarrier
\clearpage

Bike level~1 has the largest heart-rate error of the eight conditions, $18.6\pm5.5$~beats per minute.
Figure~\ref{fig:supp_xai_bike1_visualization} shows refinement pulling the attention-only methods apart, raising the skin coverage of raw attention by $0.105\pm0.015$ and lowering that of attention rollout by $0.118\pm0.019$ and attention flow by $0.062\pm0.014$.
Figure~\ref{fig:supp_xai_bike1_distributions} gives the lowest median SaCo of the three cycling conditions, 0.201 for Beyond Intuition and 0.345 for attention flow, the first being the lowest of any condition except Static level~1.

\facexxaisummary{Bike level~1}{bike1}%
{\includegraphics[width=0.72\textwidth]{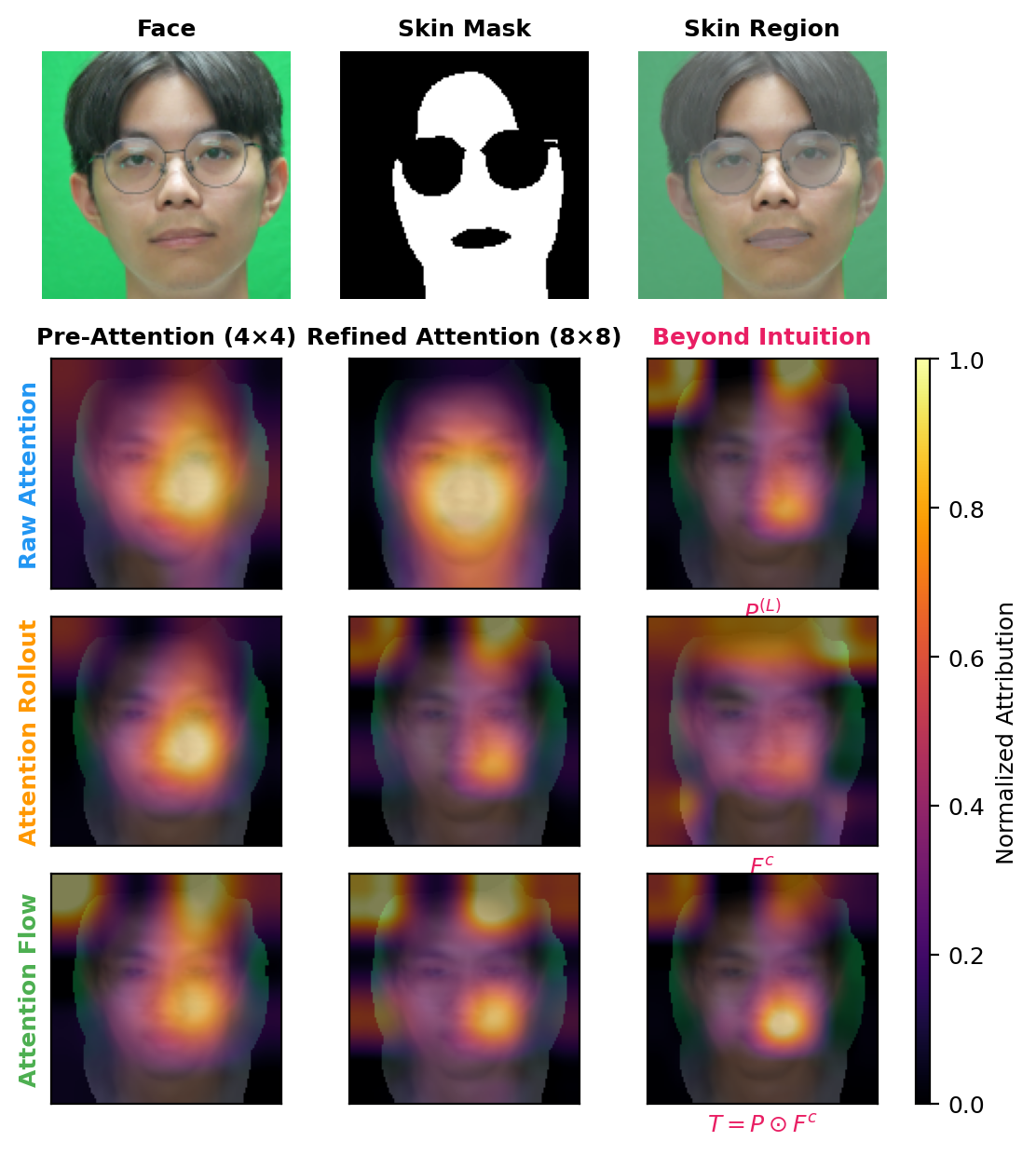}}%
{\includegraphics[width=\textwidth]{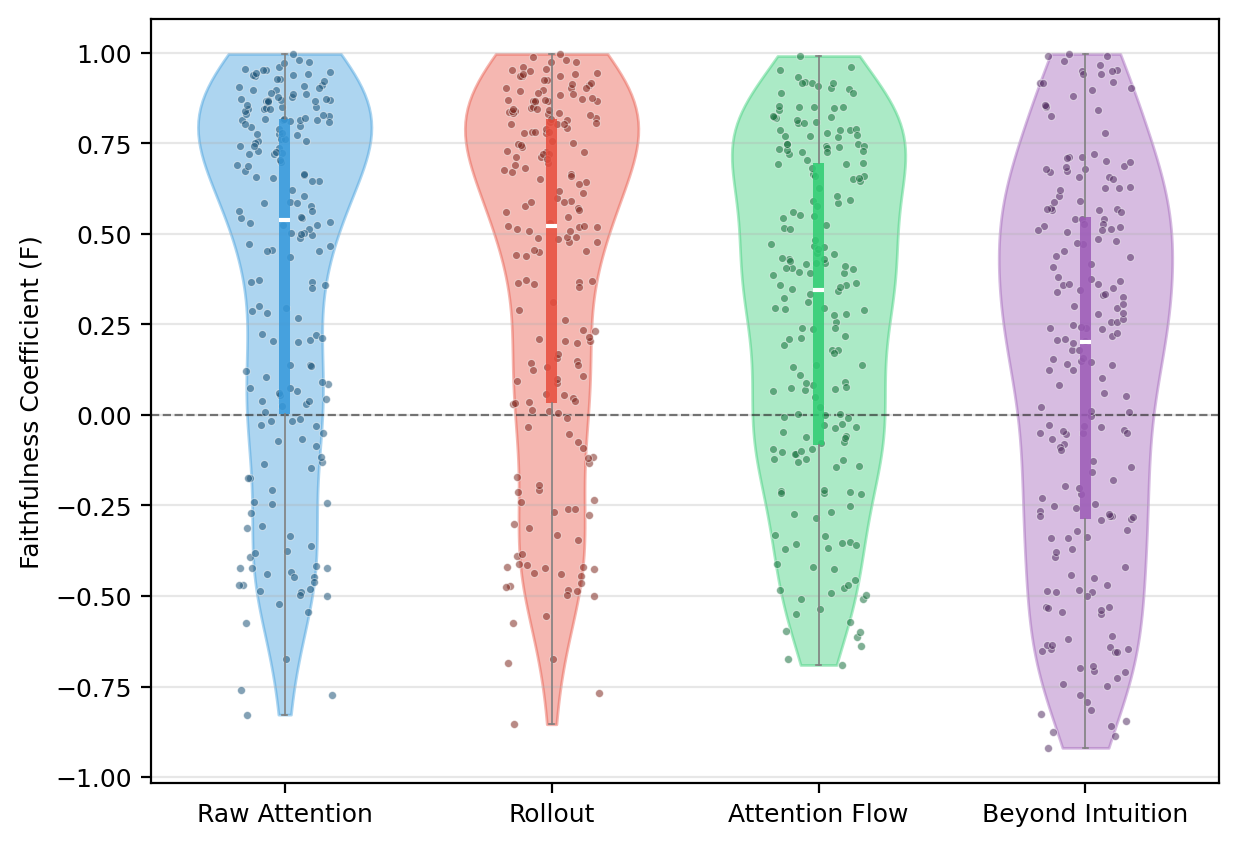}}%
{\includegraphics[width=\textwidth]{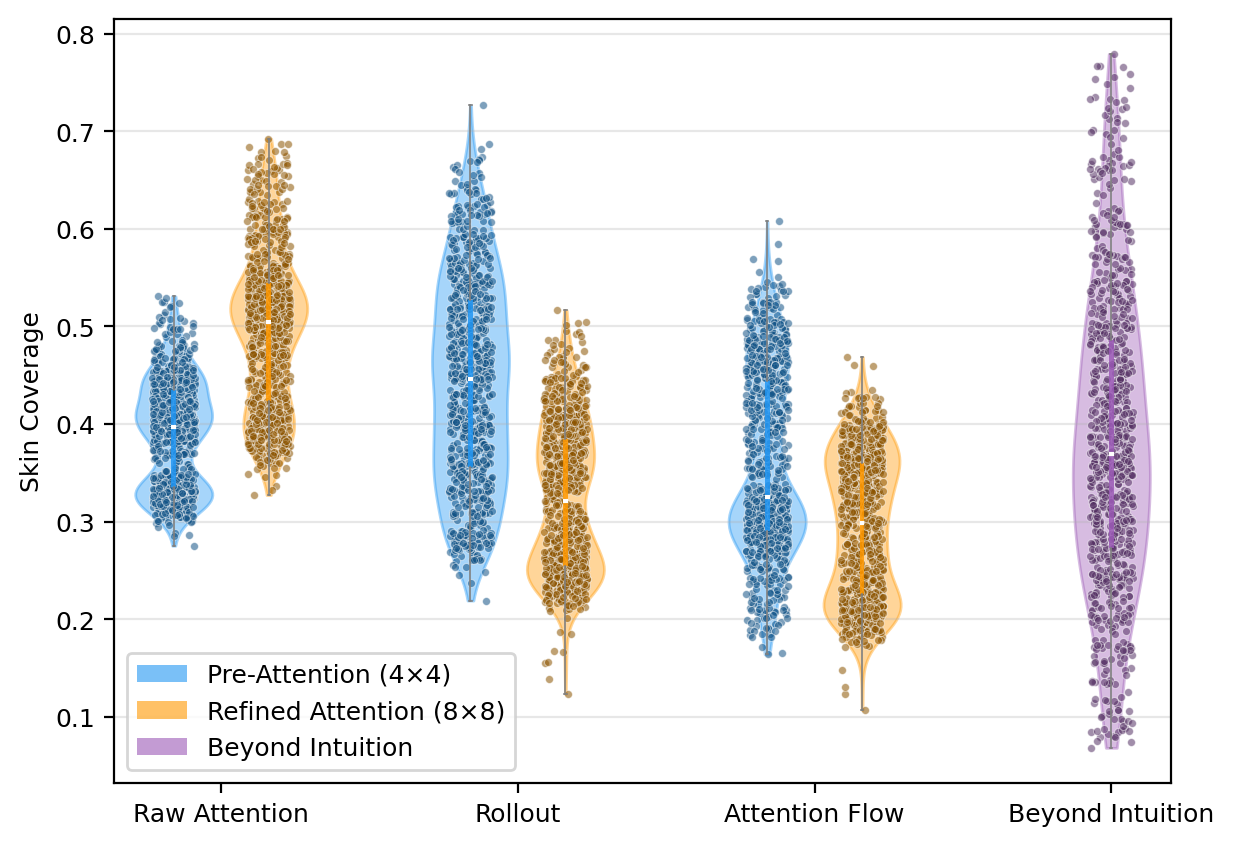}}
\FloatBarrier
\clearpage

Figure~\ref{fig:supp_xai_bike3_visualization} shows the maps of the condition in which refinement is nearly inert for attention rollout, at $-0.003\pm0.019$, while it moves raw attention by $+0.036\pm0.012$ and attention flow by $-0.039\pm0.017$.
Figure~\ref{fig:supp_xai_bike3_distributions} pairs a high median SaCo, 0.665--0.830 across the four methods, with the worst SNR of the eight conditions, $-11.9\pm2.0$~dB in Table~\ref{tab:performance_results}.

\facexxaisummary{Bike level~3}{bike3}%
{\includegraphics[width=0.72\textwidth]{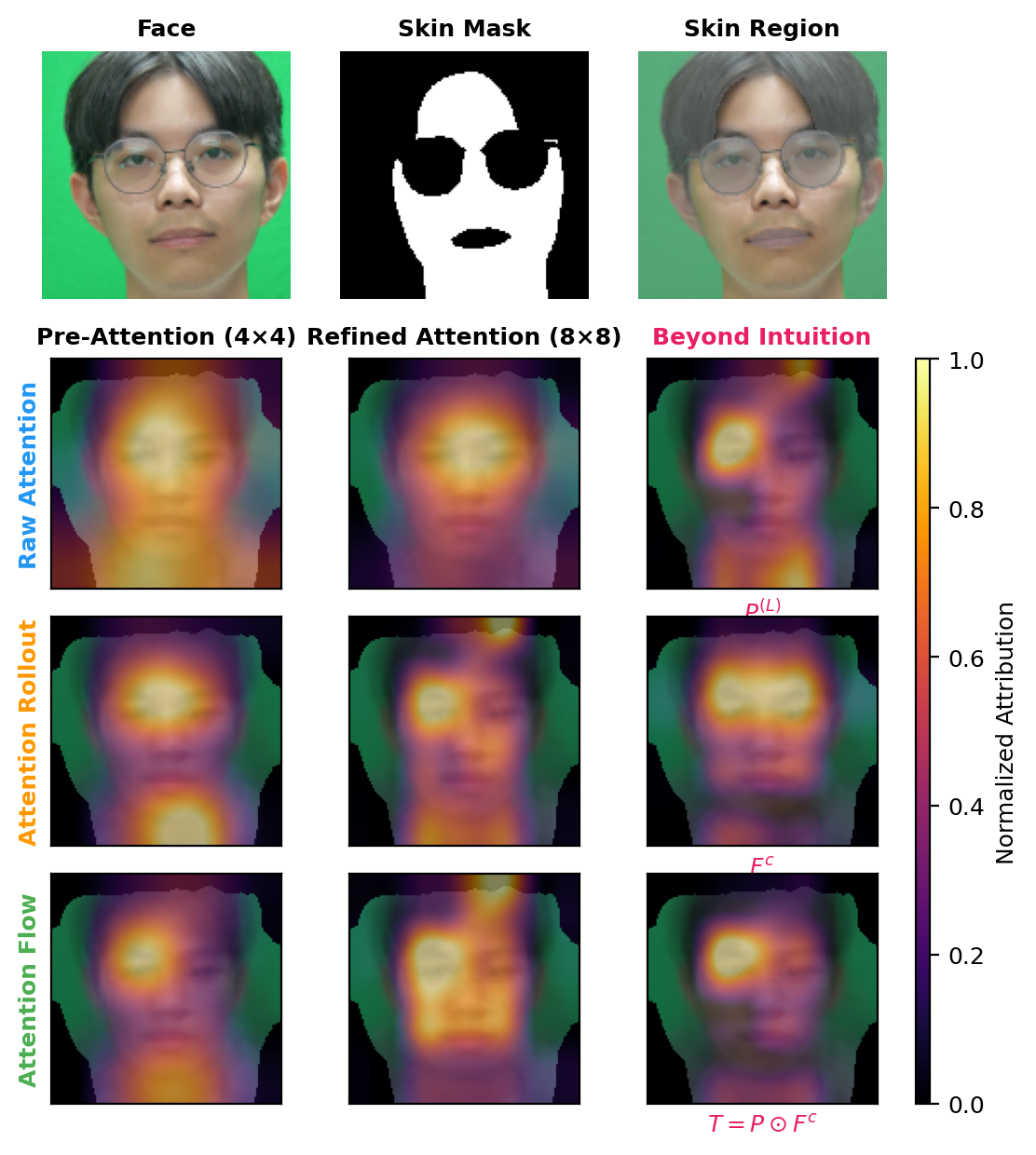}}%
{\includegraphics[width=\textwidth]{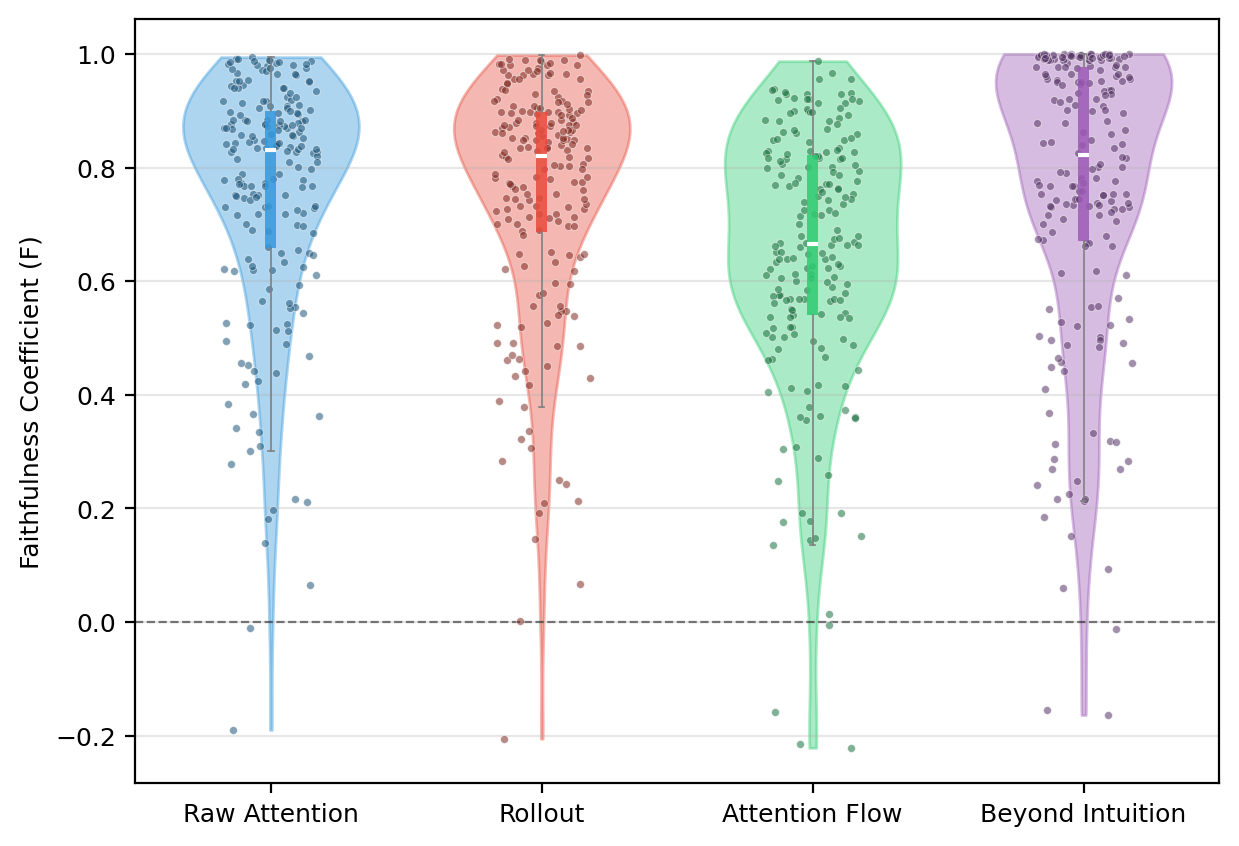}}%
{\includegraphics[width=\textwidth]{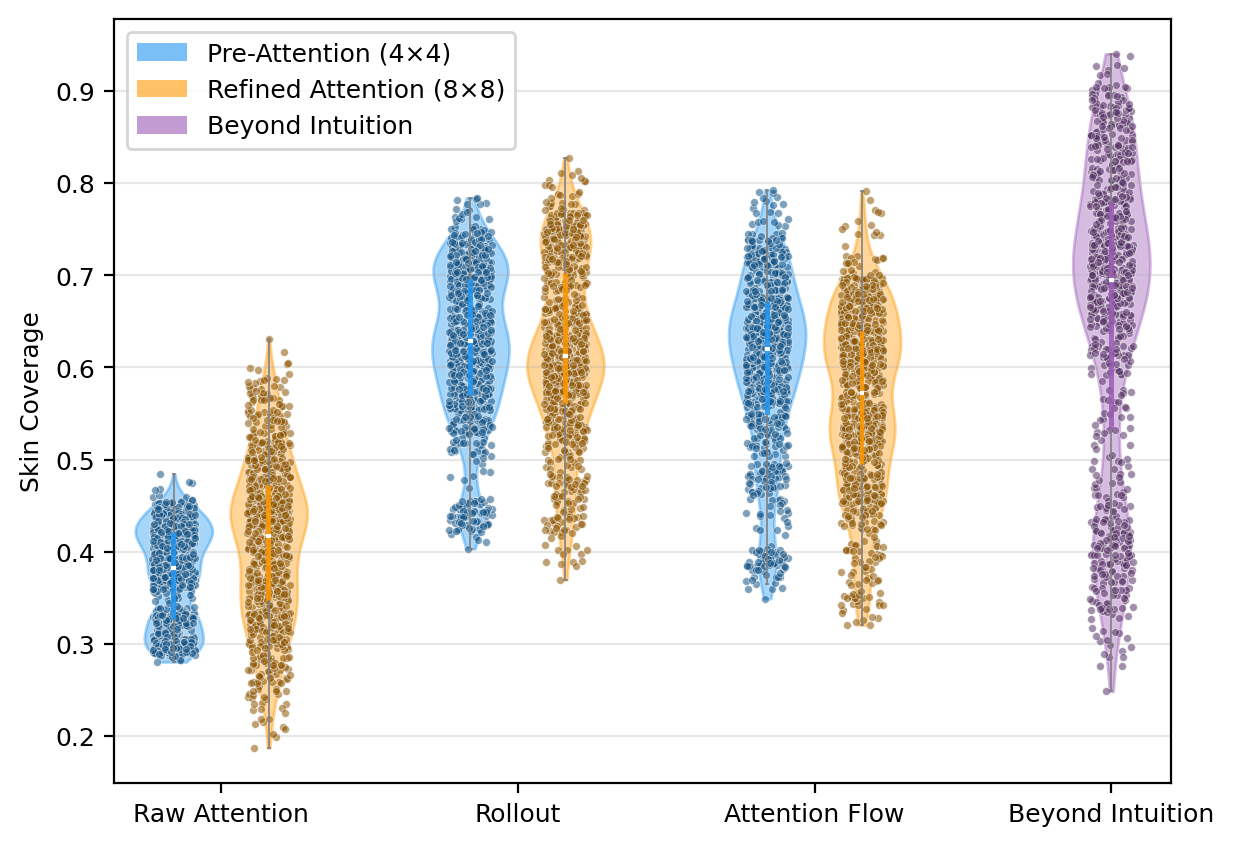}}
\FloatBarrier
\clearpage

Figure~\ref{fig:supp_xai_bike5_visualization} shows the condition in which refinement moves skin coverage furthest, by $+0.119\pm0.009$ for raw attention and $-0.128\pm0.013$ for attention rollout, about 14 and 10 times their standard errors, which is visible as a change of map between the two stages.
Figure~\ref{fig:supp_xai_bike5_distributions} gives median SaCo values of 0.680--0.767 for a condition whose heart-rate error is $15.1\pm3.5$~beats per minute.

\facexxaisummary{Bike level~5}{bike5}%
{\includegraphics[width=0.72\textwidth]{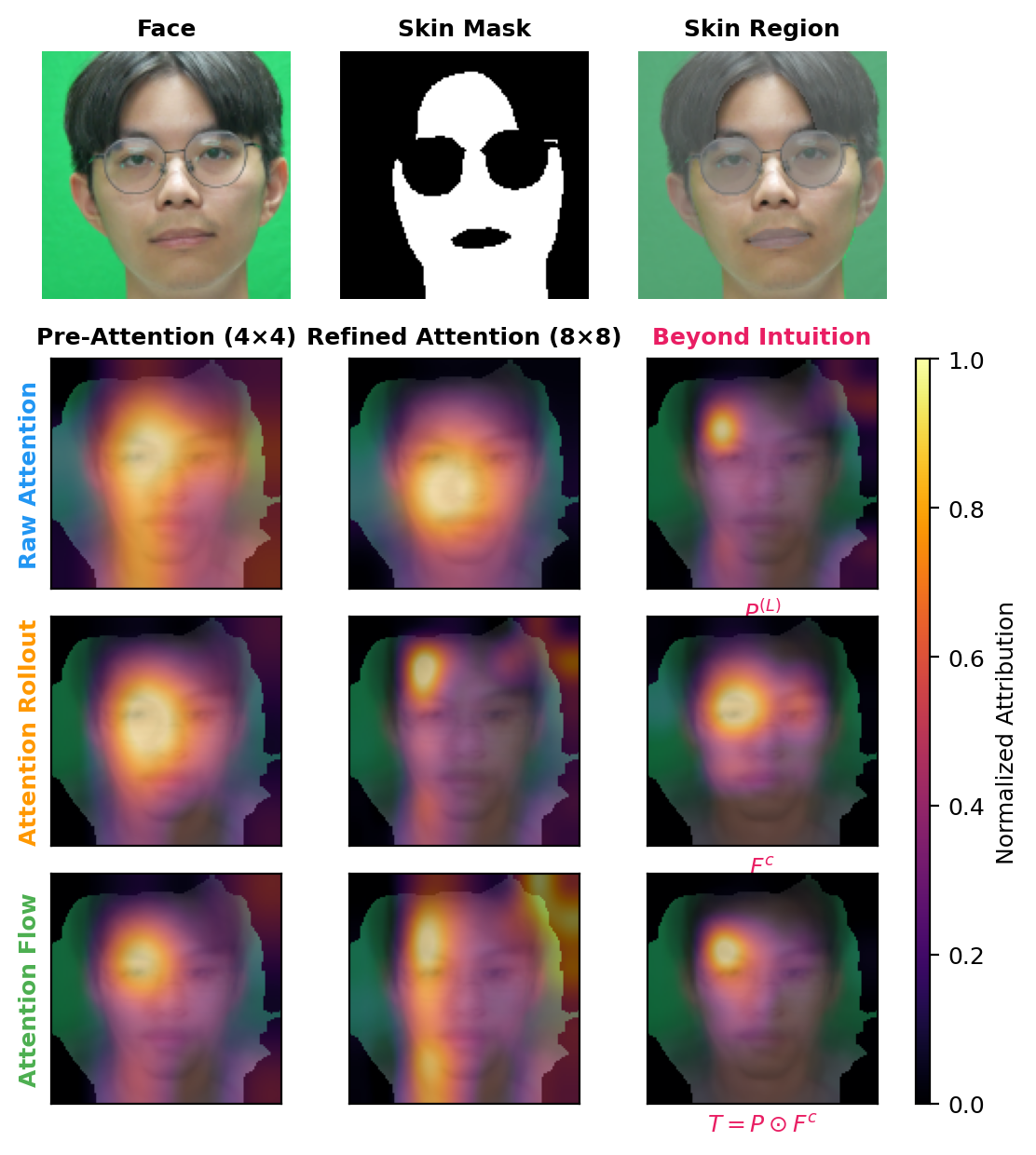}}%
{\includegraphics[width=\textwidth]{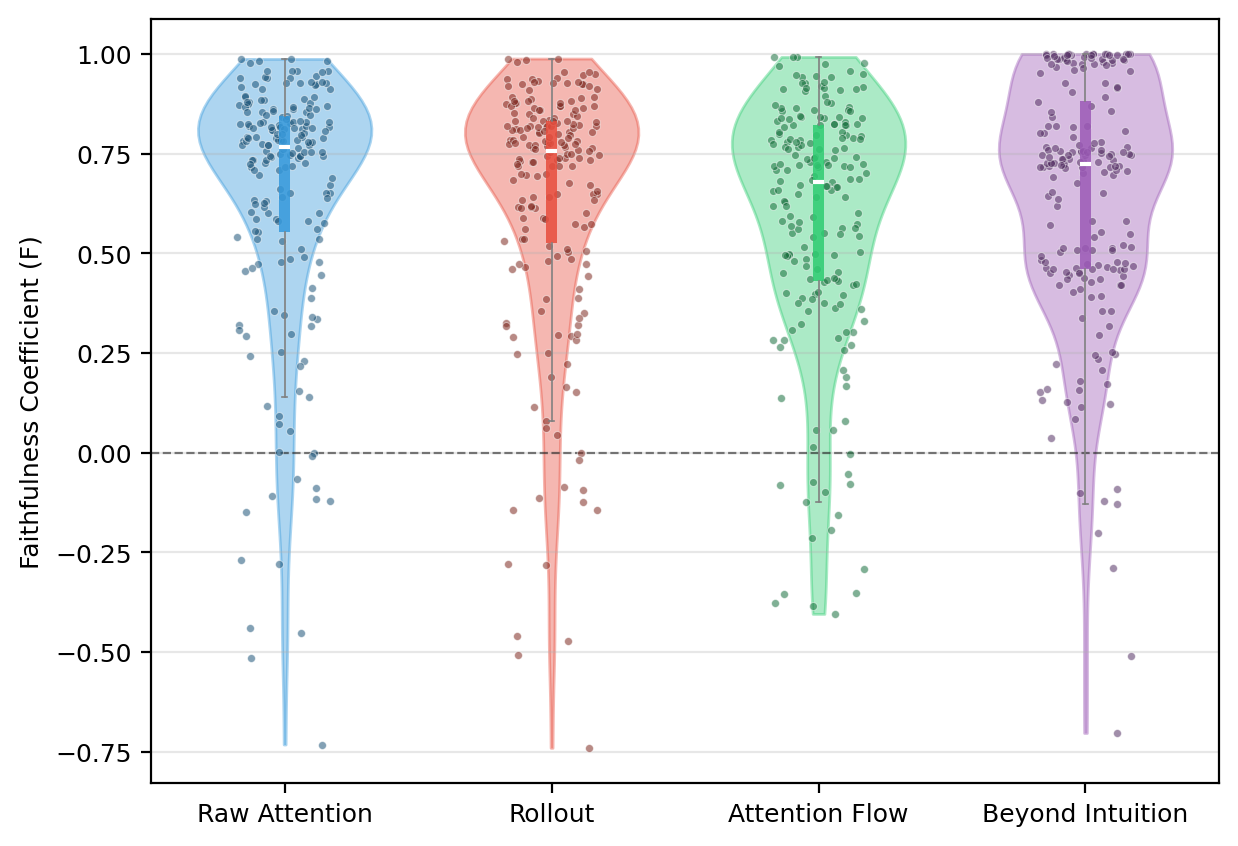}}%
{\includegraphics[width=\textwidth]{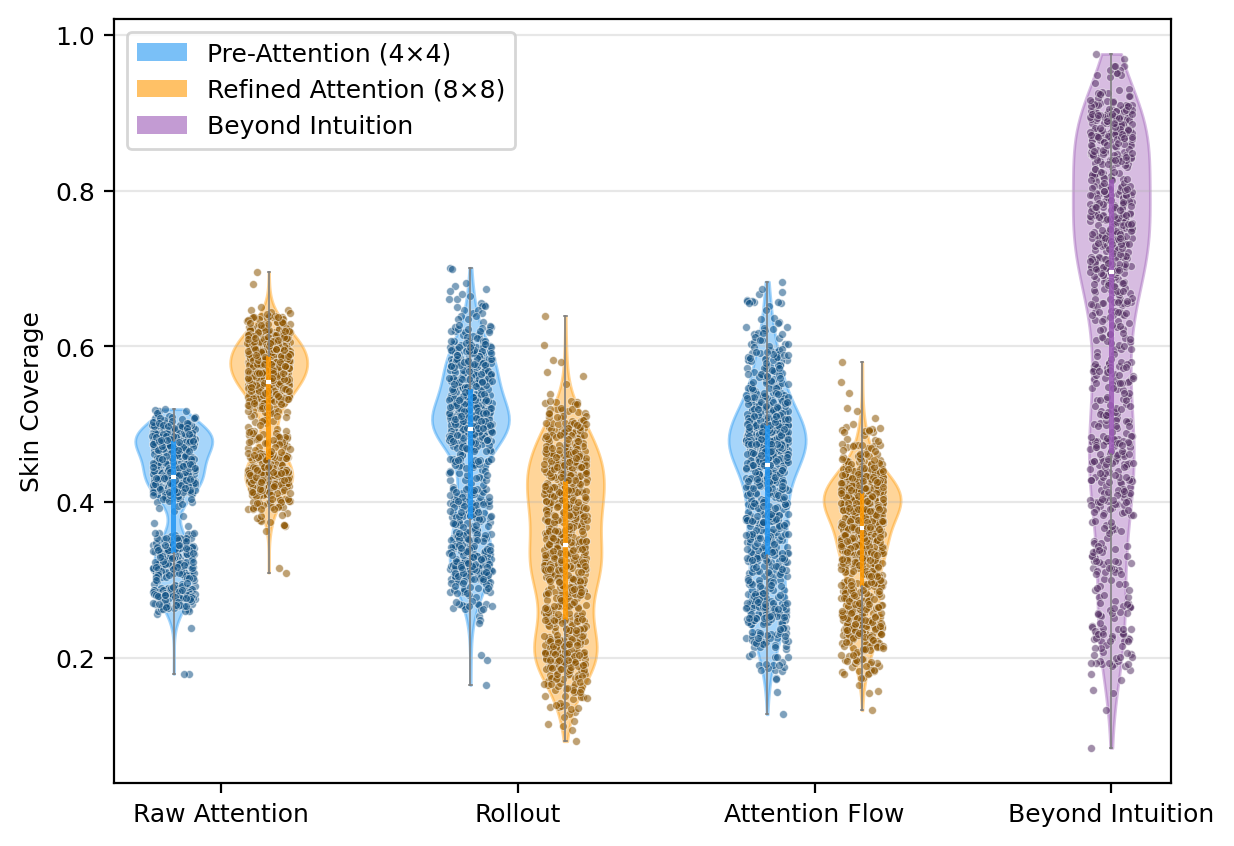}}
\FloatBarrier
\clearpage

\suppsection{sec:supp_hr_settings}{S7}{Heart-rate estimation settings in the remote photoplethysmography literature since 2020}

The evaluation in Section~\ref{sec:performance_evaluation} fixes a 5.12~s window, a Welch segment capped at 256 samples, and a 45--150~beats-per-minute search band, and Table~\ref{tab:supp_window_length} shows how strongly the reported error depends on the first of those.
A reader comparing our numbers with published ones therefore needs to know what the published ones used.
We read every article in our literature collection that was published from 2020 onwards and that applies a remote photoplethysmography method, and recorded four things from each: the video frame rate, the length of the window over which one heart rate is computed, how far consecutive windows overlap, and the spectral settings and the estimator that turn the waveform into a heart rate.
Table~\ref{tab:supp_hr_settings} gives one row for each of the 58 articles, and a last row for the present work.
Where an article does not state a setting we write that it does not, rather than filling in the value its toolbox would have used.

Four things follow from the table.
The window is the least standardised of the four settings.
It runs from 5~s to 60~s among the articles that state one, with a further article computing a single heart rate over the whole video, and the two commonest choices, a 30~s clip and a 10~s clip, differ by a factor of three;
Table~\ref{tab:supp_window_length} measures that factor in our own data: moving from 10.24 to 30.72~s changes the mean absolute error by 0.23 to 1.63~beats per minute across the eight conditions, which is of the same order as the difference between many of the methods being compared.
The 5.12~s we use is at the short end of the range but is not alone there: four articles evaluate on 160-frame clips, that is 5.33~s at 30~frames per second.

Second, consecutive windows do not overlap in most of the articles, so a longer window buys segment averaging at the cost of fewer independent estimates, exactly as in our sweep.
Ten articles overlap their evaluation windows, by between one frame and 97~\% of the window, and three more overlap only their training clips.

Third, almost every article estimates the heart rate as the largest spectral peak inside a physiological band, most often 0.75--2.5~Hz, and yet only nine of the 58 state the transform length or the window function that produces that spectrum.
Since the spectral resolution of an $N$-point transform at 30~frames per second is $1800/N$~beats per minute, a 160-frame clip transformed without zero-padding resolves no better than 11~beats per minute, which is of the same order as the errors being reported.
The articles that do state these settings differ widely: a 1024-point transform \citep{song2021PulseGANPPGIEEEJBiomedHealInform}, zero-padding to 2048 points \citep{huang2021FramPPGIEEEJBiomedHealInform}, a 256-sample Hanning window zero-padded to 3.3~k \citep{wang2024NormalizationPhotoplethysmographyArXiv}, and zero-padding chosen to give a resolution of 0.33~beats per minute \citep{speth2023NonContrastiveUnsupervisedCVPR}.
This is the setting a reader is least able to recover from a published method section, and it is the one we state explicitly in Section~\ref{sec:performance_evaluation}.

Fourth, the frame rate is the one setting on which the field has converged.
Fifty of the 58 articles process video at 30~Hz, either as recorded or after resampling to it, and the remaining eight work at 20, 25, 28, 50 or 84~Hz, or state no rate at all.
Our recordings are at 50~Hz, so a window of a given number of frames is shorter in seconds here than in most of this literature, which is why we give both throughout.

Taken together, the table shows that the settings a heart rate depends on most are the settings least often reported.
Seventeen of the 58 articles state no window over which one heart rate is computed, 23 state no overlap between consecutive windows, and four do not identify the spectral method at all.
Forty-three read the heart rate from the largest peak of a spectrum, eight derive it from detected beats or regress it directly from the video, three estimate no heart rate at all because they classify atrial fibrillation from the waveform, and four do not say which they do.
A published mean absolute error is therefore not comparable with ours, or with another study's, without the four settings this table records, and for close to a third of the literature they cannot be recovered from the article.
That is the reason we state ours in Section~\ref{sec:performance_evaluation} and vary the one that matters most in Table~\ref{tab:supp_window_length}.

\begin{landscape}
{\scriptsize
\setlength{\tabcolsep}{2.5pt}
\renewcommand{\arraystretch}{1.06}
\begin{longtable}{>{\raggedright\arraybackslash}p{3.2cm} >{\raggedright\arraybackslash}p{2.1cm} >{\raggedright\arraybackslash}p{4.0cm} >{\raggedright\arraybackslash}p{3.2cm} >{\raggedright\arraybackslash}p{4.9cm} >{\raggedright\arraybackslash}p{4.3cm}}
\caption[Heart-rate estimation settings since 2020]{Frame rate, evaluation window, window overlap, spectral settings, and heart-rate estimator of every article in our literature collection published from 2020 onwards that applies a remote photoplethysmography method, ordered by year and then by first author.
``Window'' is the span of signal from which one heart rate is computed, which is not always the model's input clip;
where the two differ, both are given.
``Not stated'' means the article does not report the setting, and we did not substitute the default of the toolbox it used.
Frame rates written as $a\rightarrow b$ mean the recording rate $a$ was resampled to $b$ before processing.}
\label{tab:supp_hr_settings} \\
\toprule
Study & Frame rate (Hz) & Window & Overlap & Spectral settings & Heart-rate estimator \\
\midrule
\endfirsthead
\multicolumn{6}{c}{\tablename\ \thetable\ (continued)} \\
\toprule
Study & Frame rate (Hz) & Window & Overlap & Spectral settings & Heart-rate estimator \\
\midrule
\endhead
\bottomrule
\endfoot
Meta-rPPG \citep{lee2020MetarPPGTransductiveLectNotesComputSci} & 30; $61\rightarrow30$ & 60-frame (2~s) model input; heart rate reported over a 30~s clip (frames 306--2135) & Not stated & Power spectral density; settings not stated & Ordinal regression to the waveform, then the spectral peak \\
DeeprPPG \citep{liu2020PhotoplethysmographySpatiotemporalIEEE} & 30; 20; 61 & 30~s clip (frames 306--2135); region-of-interest clips of length $T$ processed with stride $T/2$ & 50~\% between region-of-interest clips & Welch power spectral density with a Hann window & Spectral peak \\
MTTS-CAN \citep{liu2020MultitaskVitalsAdvNeuralInfProcessSyst} & $120\rightarrow30$; 25 & 30~s; 10-frame temporal window inside the model & None & Second-order Butterworth, 0.75--2.5~Hz; transform not stated & Spectral peak \\
HeartTrack \citep{perepelkina2020HeartTrackConvolutionalIEEE} & 25; 50 & 200 frames (8~s) & Not stated & None & Direct convolutional regression of the median heart rate \\
Siamese-rPPG \citep{tsou2020SiameserPPGPhotoplethysmographyACM} & 30; 20 & 600 frames, that is 30~s at 30~Hz and 20~s at 20~Hz & None & Fast Fourier transform; settings not stated & Spectral peak \\
Multi-task generation \citep{tsou2020MultiTaskPhotoplethysmographyAsianVision} & 30; 20 & 600 frames, following Siamese-rPPG & None & Fast Fourier transform; settings not stated & Spectral peak \\
\citet{wang2020NoncontactMeasurementNCKU} & 50 & 30~s & 25~s, that is a 5~s shift & Overlap-add with a Hann window of 80 frames for CHROM and POS and 320 frames for ICA & Spectral peak; peak intervals rejected as unreliable at low signal-to-noise ratio \\
AutoHR \citep{yu2020AutoHRBaselineIEEESignalProcessLett} & $61\rightarrow30$ & 30~s clip split into three 10~s clips & None & Power spectral density; settings not stated & Spectral peak, averaged over the three clips \\
\citet{dasari2021BiasesPhotoplethysmographyNPJDigitMed} & 25 & 10~s & Not stated & Not stated & CHROM, POS, ICA and HR-CNN; estimator not stated \\
ETA-rPPGNet \citep{hu2021ETArPPGNetDomainIEEETransInstrumMeas} & 30; 20; 25 & 10~s, with 4, 6 and 8~s also reported & None & Power spectral density; settings not stated & Spectral peak \\
\citet{huang2021FramPPGIEEEJBiomedHealInform} & 30 & 512 samples (17.06~s) & 497 samples (16.56~s), a hop of 15 samples (0.5~s); zero for one Bland--Altman figure & Zero-padded to 2048 points; third-order Butterworth, 30--200~beats per minute & Spectral peak tracking \\
Dual-GAN \citep{lu2021DualGANBVPCVPR} & 30 & 256 frames (8.53~s) & Not stated & Power spectral density; settings not stated & Spectral peak \\
PulseGAN \citep{song2021PulseGANPPGIEEEJBiomedHealInform} & 30; 25--30; 61 & 10~s sliding window & Not stated & 1024-point fast Fourier transform in the spectral loss & 60 divided by the mean inter-beat interval of the detected peaks \\
\citet{yang2022PPGIllumIEEETransHumMachSyst} & 30; about 20 & 128-frame and 64-frame model clips sampled from a 30~s video; heart-rate window not stated & Not stated & Fast Fourier transform; settings not stated & Spectral peak \\
AND-rPPG \citep{birla2022ANDrPPGDenoisingrPPGComputBiolMed} & 30; 20 & Fixed-size clips; length not stated & None & Fast Fourier transform whose length follows the clip frame rate & Spectral peak \\
\citet{lin2022MeasurementPhotoplethysmographyICSSE} & 60; 30 & 10~s & 5~s, a 5~s stride, with clips summed over the overlap & Hamming window; eleventh-order Chebyshev~II band-pass at the heart rate $\pm0.15$~Hz & CHROM spectral peak \\
CDCA-rPPGNet \citep{liu2022MeasurementConvolutionSensors} & 30 & Not stated & Not stated & Power spectral density after a Butterworth filter, 0.7--2.5~Hz & Spectral peak \\
PhysFormer \citep{yu2022PhysFormerTransformerCVF} & $61\rightarrow30$; $60\rightarrow30$ & 30~s video split into three 10~s clips & None & Power spectral density, also the training target & Spectral peak, averaged over the three clips \\
Contrast-Phys \citep{sun2022PhysSpatiotemporalECCV} & 30 & 30~s & None & Power spectral density; settings not stated & Spectral peak \\
\citet{sun2022ContactlessFibrillationSciRep} & 84 & 30~s segments, voted over the whole recording & None & Not applicable & Not estimated; the network classifies atrial fibrillation from the waveform \\
\citet{wu2022MappingPhotoplethysmographyIEEETransInstrumMeas} & 30 & Not stated & Not stated & Magnitude spectrum over the 0.5--3~Hz band, its sorted bin indices used as features & CHROM spectral peak plus a learned error-compensation term \\
EfficientPhys \citep{liu2023EfficientPhysCardiacCVF} & 30 & The whole video, one heart rate per video & None & Band-pass 0.75--2.5~Hz (45--150~beats per minute) & Peak detection and fast Fourier transform \\
rPPG-Toolbox \citep{liu2023RPPGToolboxToolboxNeurIPS} & 30; 25 & Non-overlapping chunks of $N$ frames, 180 in the worked example & None & Second-order Butterworth, 0.75--2.5~Hz (45--150~beats per minute) & Fast Fourier transform peak or peak detection \\
LSTC-rPPG \citep{lee2023LSTCrPPGPhotoplethysmographyCVPR} & 30 & 160 frames (5.33~s) & 130 frames in training, a 30-frame shift; none in testing & Power spectral density in the frequency loss & Spectral peak \\
PFE and TFA \citep{li2023PPGArbitrProcAAAIConfArtifIntell} & $\rightarrow30$ & 160 frames (5.33~s) & None & Not stated & Not stated \\
\citet{li2023ContactlessLeveragingIEEE} & $\rightarrow30$ & 10~s & None & Power spectral density; settings not stated & Spectral peak, one average pulse rate per clip \\
\citet{nguyen2023NonContactDeterioratedETFA} & $28\text{--}35\rightarrow20$ & Not stated & Not stated & Fast Fourier transform over 0.75--4~Hz for POS; Butterworth 0.75--2.5~Hz for the learned model & Spectral peak \\
\citet{speth2023NonContrastiveUnsupervisedCVPR} & 30 & 120-frame (4~s) training clip; 10~s evaluation window & Sliding; step not stated & Fast Fourier transform, input zero-padded to a resolution of 0.33~beats per minute & Highest spectral peak in 0.66--3~Hz (40--180~beats per minute) \\
\citet{chiu2023PPGInfraredIEEETransInstrumMeas} & 30 RGB; 60 near-infrared; 15 at night & 1024 samples for the Res-MF estimator, that is 34.1~s at 30~Hz & One frame & Fast Fourier transform in the training loss & Res-MF regression network, benchmarked against the spectral peak \\
\citet{wu2023PPGContactlessIEEETransInstrumMeas} & 30 & 16~s in the fatigue analysis & None, the step equals the window & Fast Fourier transform loss during training & Ordinal-regression network output, smoothed by a conditional moving average \\
\citet{wu2023FibrillationPPGIEEEJBiomedHealInform} & 84; $84\rightarrow28$ & 30~s, with 60~s also reported & None & Not applicable & Not estimated; peak localisation feeds atrial fibrillation classification \\
PhysFormer++ \citep{yu2023PhysFormerSlowFastArXiv} & $61\rightarrow30$; $60\rightarrow30$ & 30~s video split into three 10~s clips & None & Power spectral density; settings not stated & Spectral peak averaged over the three clips; MATLAB \texttt{findpeaks} for the inter-beat intervals \\
Dual-path TokenLearner \citep{qian2024TokenLearnerPPGIEEETransComputSocSyst} & 30; 25--30 & 300 frames (10~s) per spatial-temporal map & 9.5~s, one segment every 0.5~s & Fast Fourier transform; settings not stated & Peak detection and fast Fourier transform per segment, averaged to one heart rate per video \\
DiffPhys \citep{chen2024DiffPhysPhotoplethysmographyBioengineering} & 30 & Not stated & Not stated & Not used for the heart rate & $60N$ divided by the sum of the detected inter-beat intervals \\
\citet{castellanoontiveros2024ConstrPPGCommunMed} & 25; 30 & 10~s & None & Welch power spectral density; segment settings not stated & Highest spectral peak \\
CIN-rPPG \citep{li2024InteractEstimIEEETransCircuitsSystVideoTechnol} & 30 & 180 frames in training; 300 frames (10~s) in testing & 15 frames in training; none in testing & Welch power spectral density via the fast Fourier transform, after a second-order Butterworth filter & Spectral peak \\
Cluster-Phys \citep{qia2024PhysClusteringACM} & 30 & Segment length not stated; one heart rate per video by averaging segments & Not stated & Power spectral density after a first-order Butterworth filter, 0.75--2.5~Hz (45--150~beats per minute) & Spectral peak \\
Temporal normalisation \citep{wang2024NormalizationPhotoplethysmographyArXiv} & 30; 20 & Not stated & Not stated & Welch with a 256-sample Hanning window, zero-padded to 3.3~k & Spectral peak \\
STGNet \citep{xiong2024STGNetPPGBiomedSignalProcessControl} & 30 & Not stated & Not stated & Not stated & The network maps the waveform to beats per minute; estimator not stated \\
\citet{wu2024AtrialFibrillationIEEEJBiomedHealInform} & 84; 30 & 30~s segment; a sliding window of length $w$ over the inter-beat-interval series & Not stated & Frequency-domain heart-rate variability indices & Mean inter-beat interval from peak detection \\
CbPPGGAN \citep{yang2024CbPPGGANPPGIEEEJBiomedHealInform} & 30 & 128-frame (4.27~s) model clips, merged before the heart rate is computed & 118 frames in training, a 10-frame step; none in testing & Fast Fourier transform; settings not stated & Most prominent frequency of the merged waveform; SciPy peak detection for the inter-beat intervals \\
\citet{zou2024PhysiologicalSpatioFrontPhysiol} & 30; 20 & Not stated & Not stated & Power spectrum after a Butterworth filter & Highest spectral peak \\
pyMMER \citep{liu2024DetectingPhotoplethysmographyArXiv} & Not stated & 6~s & Overlapping; step not stated & Fourier power spectrum; settings not stated & Most significant frequency \\
Lifelight \citep{vanputten2024PulseExplainableDiscovApplSci} & Not stated & Not stated & Not stated & Butterworth band-pass, 0.125--4~Hz; no spectral heart-rate step & Fiducial points counted on the second derivative of the waveform \\
\citet{chiu2024ContactlessApneaIEEETransInstrumMeas} & 30 & Not stated & Not stated & Spectral peak of the waveform, used to reject implausible R--R intervals & Spectral peak, with peak detection for the intervals \\
\citet{cen2025GeneralizablePhotoplethysmographyArXiv} & 30 & Not stated & Not stated & Welch power spectral density after a second-order Butterworth filter & Frequency of maximum spectral power \\
Periodic-MAE \citep{choi2025PeriodicMAErPPGArXiv} & 30; 25 & 160 frames (5.33~s) & Not stated & Power spectral density via the fast Fourier transform & Spectral peak \\
CodePhys \citep{chu2025CodePhysCodebookArXiv} & $\rightarrow30$ & 30~s video split into three 10~s clips & None & Power spectral density; settings not stated & Spectral peak averaged over the three clips \\
DD-rPPGNet \citep{huang2025DDrPPGNetrPPGIEEETransInfForensicsSecur} & 30 & 30~s & None & Power spectral density; settings not stated & Spectral peak \\
TS-CAN+ \citep{li2025TSCANNonContactIEEETransConsumElectron} & 30 & Not stated & Not stated & Not stated & The network outputs the heart rate; estimator not stated \\
PhysKANNet \citep{liu2025PhysKANNetKANbasedBiomedSignalProcessControl} & 30; 20; 35 & Not stated; the sampling interval is half the temporal dimension of the spatial-temporal block & Half the block length between samples & Power spectral density after a second-order Butterworth filter & Spectral peak \\
\citet{tseng2025AtrialFibrillationIEEEJBiomedHealInform} & 84; $84\rightarrow30$ & 20, 30 and 60~s compared & None & Not applicable & Not estimated; a one-stage network classifies atrial fibrillation \\
\citet{wang2025NonEndtoEndEstimationNCKU} & 50; 30 & 500-frame model clip; 5~s short-time Fourier transform window for the heart rate & None between clips; 4~s between transform windows, a 1~s step & Short-time Fourier transform over 0.65--4~Hz (39--240~beats per minute); Hamming window in preprocessing & Dominant short-time Fourier transform frequency \\
CAP-rPPG \citep{zhang2025PyramidPhysiologicalSciRep} & 30; 60 & 180 frames (6~s) in training; 900 frames (30~s) in testing & None & Fast Fourier transform; a power spectral density loss over a defined band & Spectral peak \\
RhythmFormer \citep{zou2024RhythmFormerrPPGArXiv} & 30; 20 & 160 frames (5.33~s) & None & Welch power spectral density after a second-order Butterworth filter, 0.75--2.5~Hz & Spectral peak \\
\citet{chang2026PhotoplethysmographySystolicJCardiol} & 30 & Not stated & Not stated & Frequency-domain heart-rate variability indices, reported as sensitive to the observation window & Mean heart rate from a proprietary software development kit; algorithm not disclosed \\
FreqPhys \citep{qian2026FreqPhysPhotoplethysmographyArXiv} & 30 & Not stated & Not stated & Discrete Fourier transform with an ideal band-pass over 0.66--3.0~Hz & Dominant spectral peak, multiplied by 60 \\
KDPhys \citep{sahoo2026KDPhysDistillationArXiv} & 30; 20 & 10~s & Not stated & Fast Fourier transform on Hanning-windowed signals & Peak of the periodogram \\
This work & 50 & 5.12~s (256 frames), swept to 61.44~s in Table~\ref{tab:supp_window_length} & None & Welch power spectral density, segment capped at 256 samples, 0.75--2.5~Hz (45--150~beats per minute) & Spectral peak \\
\end{longtable}
}
\end{landscape}
\FloatBarrier
\clearpage

\end{document}